\documentclass[journal]{IEEEtran}
\usepackage[utf8]{inputenc}
\usepackage{amsmath}
\usepackage{soul}
\usepackage{array}
\usepackage{amsmath,amsthm,amssymb,epsf}
\usepackage{bm} 
\usepackage{graphicx,psfrag}
\usepackage{epstopdf}
\usepackage{verbatim}
\usepackage{multirow}
\usepackage{color}
\usepackage{setspace}
\usepackage{enumitem}
\usepackage{subfigure}
\usepackage{stackengine}
\usepackage{amssymb}
\usepackage{tabularx}
\usepackage{booktabs}
\usepackage{dirtytalk}
\usepackage{algorithm}
\usepackage{MnSymbol}
\usepackage{mathdots}
\usepackage{amsmath}
{
      \theoremstyle{plain}
      \newtheorem{assumption}{Assumption}
  }
\usepackage{csquotes}
\usepackage{algorithm}
\usepackage{algorithmic}

\newtheorem{theorem}{Theorem}
\newtheorem{definition}{Definition}

\newtheorem{lemma}{Lemma}
\newtheorem{corollary}{Corollary}

\newtheorem{remark}{Remark}

\usepackage{xcolor}

\usepackage{soul}
\usepackage{cite}
\usepackage{relsize}
\usepackage{stfloats}
\usepackage{url}          

\begin{document}

\title{\LARGE From Shadow to Light: Toward Safe and Efficient Policy Learning Across MPC, DeePC, RL, and LLM Agents}

\author{Amin Vahidi-Moghaddam, Sayed Pedram Haeri Boroujeni, Iman Jebellat, Ehsan Jebellat, Niloufar Mehrabi, and Zhaojian Li
\thanks{Zhaojian Li is the corresponding author.}
\thanks{Amin Vahidi-Moghaddam and Zhaojian Li are with the Department of Mechanical Engineering, Michigan State University, East Lansing, MI 48824, USA (e-mail: vahidimo@msu.edu, lizhaoj1@egr.msu.edu).}
\thanks{Sayed Pedram Haeri Boroujeni and Niloufar Mehrabi are with the School of Computing, Clemson University, Clemson, SC 29632, USA (e-mail: shaerib@g.clemson.edu, nmehrab@g.clemson.edu).}
\thanks{Iman Jebellat is with the Department of Mechanical and Industrial Engineering, University of Toronto, Toronto, ON M5S 2E4, Canada (email: iman.jebellat@utoronto.ca)}
\thanks{Ehsan Jebellat is with the Department of Mechanical Engineering, Colorado School of Mines, Golden, CO 80401, USA (e-mail: jebellat@mines.edu).}
} 

\maketitle

\begin{abstract}
One of the primary challenges in modern control applications, particularly in robot/vehicle motion control, is achieving accurate, fast, and safe movement. To address this challenge, optimal control policies have been developed to enforce safety while ensuring optimal behavior. Typically, a basic first-principles model of the real system is available, which motivates the use of model-based optimal controllers. Model predictive control (MPC) is a state-of-the-art approach that optimizes control performance while explicitly handling safety constraints. However, obtaining an accurate first-principles model for complex systems remains a major challenge, which is addressed using different data-driven optimal policies. Machine learning (ML)-based MPC leverages learned models to reduce reliance on hand-crafted models. Similarly, reinforcement learning (RL) has shown promise in learning near-optimal control policies directly from interaction data. Moreover, as a recently emerging data-driven optimal policy, data-enabled predictive control (DeePC) eliminates the need for model learning by directly learning a safe, optimal control policy from raw input/output (I/O) data. On the other hand, large language model (LLM) agents have recently emerged as a promising paradigm for robot/vehicle motion control, facilitating the translation of high-level natural language instructions into structured formulations of optimal control policies. However, these data-driven optimal policies suffer from slow response times and high computational demands, particularly for complex and nonlinear systems. They also often require large amounts of memory to store data-driven representations, further straining limited hardware resources. This limits their practicality for real-world applications with fast dynamics, limited on-board computing power, or constrained memory. To mitigate this issue, various techniques—including reduced-order modeling, optimal policy learning using function approximators, and convex approximations—have been proposed to reduce the computational complexity of the data-driven optimal policies. In this paper, we discuss eight approaches to improve efficiency and demonstrate their performance on real-world applications, including robotic arms, soft robots, and vehicle motion control.
\end{abstract}


\IEEEpeerreviewmaketitle

\section{Introduction}
\begin{figure}[!ht]
     \centering
     \includegraphics[width=8.8cm, height=7cm]{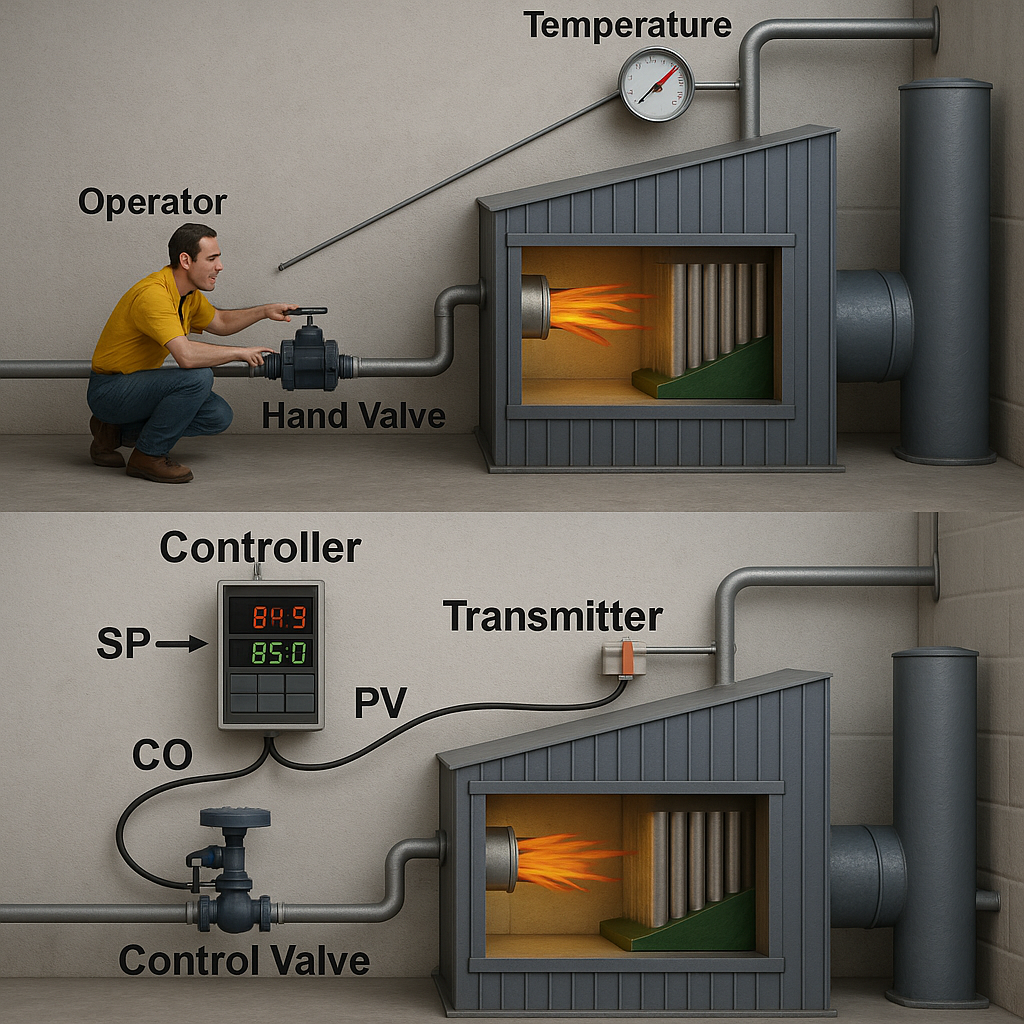}
     \caption{Comparison of manual control and automated control (PV as process variable, SP as setpoint, and CO as control input).}
     \label{automated control}
\end{figure}

The story of automatic control began when humans first sought to make their lives easier. The field of automatic control has played a crucial role in the advancement of engineering and industrial automation, enabling precise and stable regulation of dynamic systems. As shown in Fig. \ref{automated control}, early control methods relied heavily on manual adjustments, which introduced significant challenges such as slow response times, human errors, and inconsistencies in maintaining optimal operating conditions. Operators were required to continuously monitor real-time data and fine-tune control input, making it difficult to manage complex and rapidly changing processes efficiently. As industrial systems grew in scale and complexity, the need for automated control mechanisms became more evident. This necessity led to the development of classical controllers, such as proportional–integral–derivative (PID) control \cite{minorsky1922directional}, which emerged as a reliable and widely used solution to maintain system stability and performance. Over time, control theory has evolved beyond classical controllers, incorporating advanced techniques such as robust \cite{zams1981feedback}, adaptive \cite{whitaker1958design, aastrom1973self, astrom1994adaptive}, optimal \cite{bellman1954theory, boltyanskiy1961theory, kalman1960contributions, pontryagin2018mathematical}, predictive \cite{testud1978model}, fuzzy \cite{zadeh1965fuzzy, zadeh1973outline, mamdani1975experiment}, and intelligent control \cite{werbos1974beyond,  kumpati1990identification} to meet the demands of modern applications.

\begin{figure}[!ht]
     \centering
     \includegraphics[width=8.8cm, height=7cm]{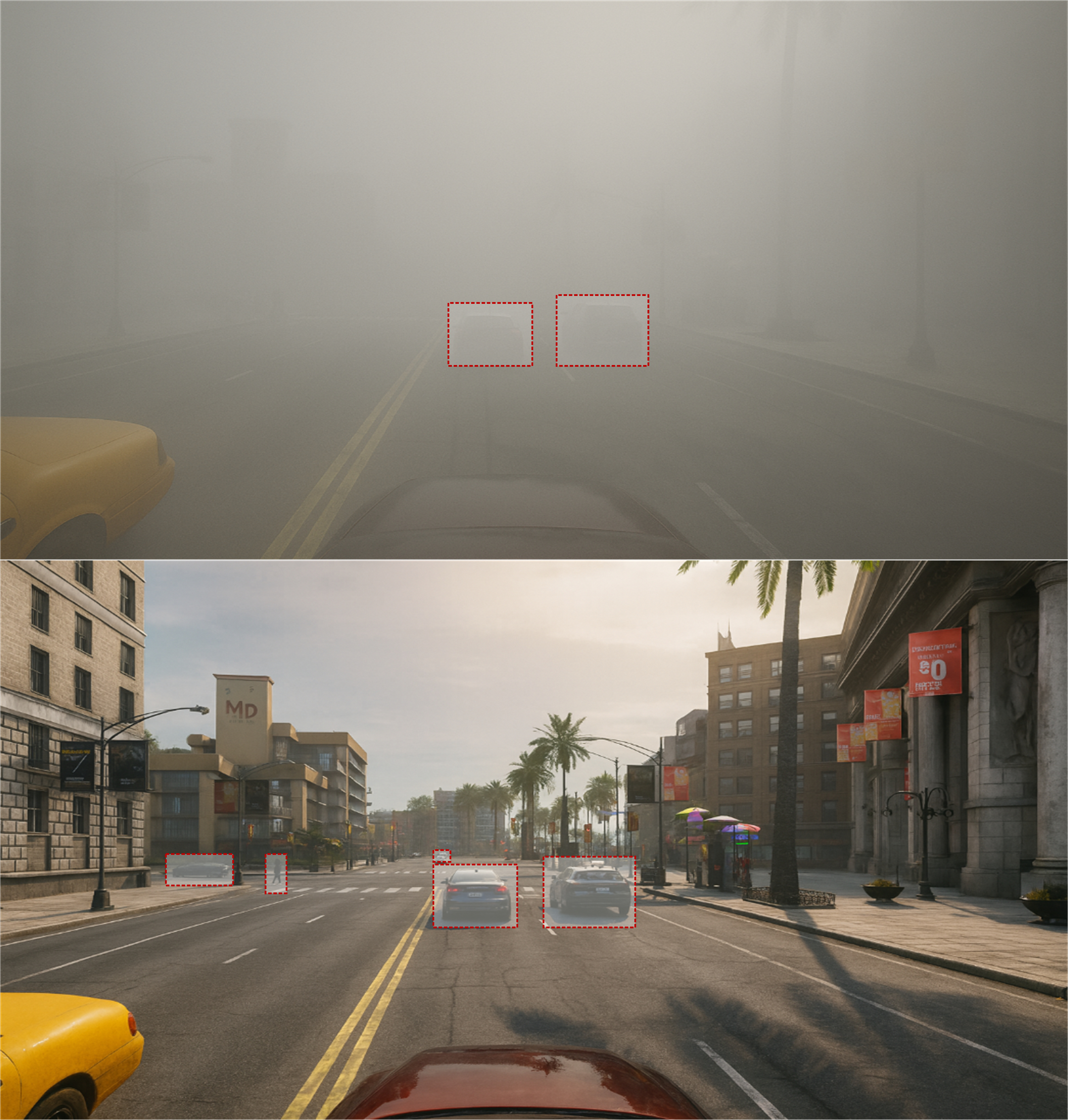}
     \caption{Illustration of PID and MPC as driving in two different weather conditions.}
     \label{PIDvsMPC}
\end{figure}

One of the primary challenges in modern control applications, particularly in robot/vehicle motion control, is achieving accurate, fast, and safe movement. Although PID control provides high-precision tracking and fast response, it is a model-free controller and does not inherently account for optimal decisions and system safety. The dynamic nature of robots/vehicles, combined with their interactions with uncertain and dynamic environments, introduces safety constraints that can significantly impact control performance. Failure to address these constraints can lead to instability, performance degradation, and safety hazards. To mitigate these risks, optimal control has been developed to ensure both safety and optimal performance. Model-based optimal control, particularly model predictive control (MPC) \cite{testud1978model}, has gained significant attention due to their ability to explicitly handle constraints while optimizing control actions over a prediction horizon. MPC uses a predictive model to forecast the consequences of its decisions/actions in advance, and by leveraging the optimization science, explore and analyze different decisions to choose the best one that makes a safe, optimal behavior for the system. As shown in Fig. \ref{PIDvsMPC}, we can illustrate these two control frameworks, i.e., PID and MPC, as driving in two different weather conditions. Imagine yourself driving in bright, sunny weather that allows you, as the controller of this system, to predict the consequences of your decisions in advance since you have clear information about the environment. However, if you choose to act like PID, you actually reject the advantage of the sunny weather in choosing your actions. It seems like you put yourself in a situation where you are driving in thick fog. You only see the car directly in front of you (your current measurement), so you accelerate/brake based on your limited information. You do not know what is beyond your limited view, which leads to over- or under-reactions (non-optimal decisions) and also hazardous reactions (unsafe decisions). However, if you act like MPC, you use your information from your clear day, which enables you to see far ahead (curves, hills, or slow traffic) and plan your throttle/brake sequence accordingly making safe, optimal actions.

However, one of the key limitations of MPC is its reliance on accurate system models, which can be difficult to obtain for complex, highly nonlinear systems such as soft robots or autonomous vehicles in unstructured environments. To overcome this challenge, data-driven optimal policies, including deep learning (DL)-based MPC \cite{krishnamoorthy2021adaptive}, Gaussian process regression (GPR)-based MPC \cite{arcari2020meta}, reinforcement learning (RL) \cite{sutton2018reinforcement}, and MPC-based RL \cite{zanon2020safe}, have emerged as powerful alternatives, leveraging real-time data to improve model accuracy and adaptability. The data-driven optimal policies are particularly beneficial in scenarios where developing a comprehensive first-principles model is not conceivable (e.g., in human-in-the-loop applications), or is not practical for control design due to the complexity of the models (e.g., in fluid dynamics systems), or is not cheap (e.g., in robotics). In contrast to the traditional MPC, which is based on first-principles models, the data-driven optimal policies involve development of control policy derived directly from input/output (I/O) data from actual systems \cite{hou2013model,breschi2022design}. RL methods are powerful for learning flexible policies in unstructured environments, but in practice they often suffer from poor sample efficiency and limited safety guarantees, making them difficult to deploy directly on physical systems \cite{jebellat4559765reinforcement, jebellat2021training, jebellat2023trajectory}. ML-based MPC approaches (e.g., DL-based MPC and GPR-based MPC) are more practical when moderate amounts of data are available, since they retain the optimization-based structure of MPC while leveraging learned models to improve predictive accuracy, thereby reducing reliance on large-scale exploration and preserving interpretability and safety. Nonetheless, ML-based MPC normally requires continual adaptation to cope with distribution shifts and evolving system dynamics, which can limit robustness and increase implementation complexity. MPC-based RL methods combine the strengths of both paradigms, integrating the safety handling of MPC with the adaptability of RL, which allows policies to adjust more flexibly while still benefiting from optimization-based safeguards. Consequently, they provide a more deployment-ready solution that balances exploration costs, data efficiency, and real-world safety—offering a sample-efficient and safer alternative for applications where direct trial-and-error learning may be infeasible.

There are two distinct approaches within the data-driven optimal policy. The first is indirect data-driven optimal policy, where a model is initially learned offline using collected I/O data from the real system, followed by the optimal policy synthesis based on this model \cite{hewing2020learning, vahidi2022data}. The second approach is direct data-driven optimal policy, which bypasses model learning process, deriving a data-driven optimal policy straight from the I/O data \cite{hou2013model, berberich2020data}. 
The main reason for a shift from indirect to direct design is that the two-step procedure of the indirect design may be suboptimal \cite{markovsky2023data}. The indirect model-based approach splits the
overall problem into two sequential sub-problems: 1) model
fitting, and 2) model-based optimal policy. The model learning sub-problem minimizes a (maximum-likelihood) data fitting cost function over the model parameters, using the data and the prior knowledge about the true system, but not the optimal policy cost function in MPC (or return in RL). On the other hand, the model-based optimal policy sub-problem minimizes the cost function in MPC (or maximizes the return in RL), using the learned model, but not the data and the prior knowledge about the true system. This two-step procedure is in general suboptimal because, except for special cases, there is no separation principle for model learning and model-based optimal policy. Therefore, an end-to-end direct method may outperform indirect methods. Additional arguments in favor of the direct approaches are incompatibility of model learning and model-based optimal policy approaches and the need of model structure selection in the learning process. To conclude, it is more efficient to learn optimal policies directly from the data rather than through model formulation, exemplified by the classical PID control.

Recently, a key result from behavioral systems theory \cite{willems1986time, willems1997introduction}, known as Fundamental Lemma \cite{willems2005note}, has gained renewed attention in the direct data-driven optimal policy. Instead of relying on model learning (parametric representation), this result allows for direct learning of linear time-invariant (LTI) system behavior (non-parametric representation) by capturing the subspace of the I/O trajectories through the column span (range space or image) of a data Hankel matrix. Compared to ML-based approaches, the Fundamental Lemma requires less data, offers improved computational efficiency, and is more amenable to rigorous stability and robustness analysis \cite{de2019formulas}. One notable application of this principle is data-enabled predictive control (DeePC), a recently developed direct data-driven optimal policy method that eliminates the need for explicit model learning by directly deriving an optimal and safe control policy from raw I/O data. Under perfect (noiseless and uncorrupted) I/O data, DeePC can accurately predict the future behaviors of the LTI systems using the Fundamental Lemma. In this ideal case, DeePC exhibits closed-loop behavior equivalent to MPC with an explicit model and perfect state estimation \cite{zanon2021constrained, luo2020adaptive, coulson2019data}. However, in real-world applications, noise and/or nonlinearity make this ideal case inaccessible, leading to inaccuracies in system's behavior prediction and suboptimal control performance. Additionally, since the Fundamental Lemma is formulated for the LTI systems, it does not directly extend to nonlinear dynamics. To address these challenges, DeePC is enhanced with robustification techniques (i.e., regularization terms, auxiliary slack variables, and low-rank approximations) to improve performance under noisy conditions and system nonlinearities \cite{dorfler2022bridging, breschi2023data}. Moreover, quadratic regularization has been shown to be crucial for stability guarantees in data-driven control frameworks \cite{berberich2020data}.

Similar to PID control, DeePC removes the dependency on a parametric system model; however, it extends control capabilities by explicitly incorporating safety constraints and optimal decision-making, two critical advantages over traditional PID controllers. By leveraging predictions of future system behavior and optimizing control inputs over a finite horizon, both MPC and DeePC ensure constraint satisfaction and enhanced stability in dynamic and uncertain environments. Nevertheless, this added layer of safety and optimization comes at the expense of slower response times and large amounts of memory, as both methods require solving complex optimization problems at each control step. The high computational burden of real-time optimization makes MPC and DeePC less suitable for applications demanding ultra-fast response times. This limitation is particularly evident in industries such as autonomous robotics, vehicular control, power electronics, and high-frequency actuation systems, where rapid control updates are critical. Consequently, despite their advanced capabilities, MPC and DeePC are often avoided in favor of simpler, more responsive controllers like PID in such high-speed applications. Compared to MPC, DeePC generally incurs a higher computational cost due to its reliance on high-dimensional optimization variables, resulting from the use of the collected I/O data rather than explicit model-based dynamics. 

This paper reviews and compares eight proposed approaches that aim to improve the efficiency of the data-driven optimal policies while maintaining their performance benefits. Although the primary focus is on DeePC, many of these methods are also applicable to ML-based MPC, RL, and LLM agents. Specifically, subspace system identification \cite{di1997subspace,favoreel1999spc, huang2019data} uses least-squares to derive a low-dimensional autoregressive model with exogenous inputs (ARX) that captures the system dynamics. Reduced-order modeling \cite{ravindran2000reduced, ashtiani2024data, ashtiani2022scalable} simplifies system representation, thereby decreasing the optimization problem’s dimensionality. Optimal policy learning through function approximators, including DL \cite{kumar2018deep, thananjeyan2019extending} and GPR \cite{rasmussen2003gaussian, ashtiani2024reconstructing, liu2018gaussian}, enables faster decision-making by replacing computationally expensive optimization with pre-trained optimal policies. Convex approximations, e.g., perturbation analysis \cite{ghaemi2009integrated} and neighboring extremal \cite{ghaemi2008neighboring}, and reformulations of the underlying optimization problem, e.g., range space (image), null space (kernel), and discrete Fourier transform (DFT)-based factorization of the data matrix, have been explored to improve efficiency. These advancements collectively aim to make the data-driven optimal policy a more viable option for real-time applications, bridging the gap between data-driven and model-based optimal control policy approaches in complex dynamical systems. In Section IV, we discuss all eight proposed approaches in detail. Using these efficient approaches, DeePC has been successfully applied in real-world applications, including quadcopters \cite{elokda2021data}, robotic arms \cite{vahidi2025data}, and soft robots \cite{wang2024mechanical}. By analyzing these efficient frameworks, this survey highlights key advancements in data-driven optimal policy and provides insights into improving their efficiency and practicality for real-world implementations.


\textbf{Notations}. We adopt the following notations across the paper. $\mathbb{R}^n$ and $\mathbb{R}^{n\times m}$ represent the set of $n$-dimensional real vectors and the set of $n\times m$-dimensional real matrices, respectively. $\mathbb{R}^{n \times m}[z]$ denotes the set of all polynomial matrices in the indeterminate $z$, i.e., matrices of the form $\mathcal{R}(z) = \sum_{i=0}^{l} \mathcal{R}_i z^i, \hspace{1 mm} \mathcal{R}_i \in \mathbb{R}^{n \times m}$. $x^{\top}$ and $A^{\top}$ stand for the transpose of the vector $x$ and the matrix $A$, respectively. $\rm{rank}(A)$ represents the rank of $A$. $\sigma_{\rm{min}} (A)$ and $\sigma_{\rm{max}} (A)$ stand for the minimum singular value and the maximum singular value of $A$, respectively. $\sigma_{\rm{r}} (A)$ represents the corresponding singular values to $\rm{rank}(A)$, which is the minimum non-zero singular value of matrix $A$. $\mathrm{im}(M)$ and $\ker(M)$ denote image and kernel of $A$, respectively. When a basis of $\ker(A)$ is to be computed, we write $N = \mathrm{null}(A)$, which returns a matrix $N$ such that $AN=0$. 
Given a signal $w$, $w_k$ represents the signal $w$ at time step $k$, and $w_{1:T}$ denotes the restriction of $w_k$ to the interval $[1, T]$, namely $w_{1:T} = [w^{\top}_1, w^{\top}_2, \cdots, w^{\top}_T]^{\top}$. For a symmetric positive-definite matrix $P=P^\top\succ0$, we define the weighted norm of the vector $x$ as:
\[ \|x\|_P \;=\;\sqrt{x^\top P\,x}. \] 
In contrast, $\|x\|_i$, $i\in\{1,2,\infty\}$, denote the standard $\ell_1$, $\ell_2$ (Euclidean), and $\ell_\infty$ norms.

\section{Data-Driven System Representation}
Physics aims to describe, classify, and predict natural phenomena as a statical/dynamical model, which requires deep knowledge about the considered phenomenon \cite{wigner1990unreasonable}. This model ideally has the same behavior as the real-life phenomenon; however, achieving such an exact model from physical laws (first-principles modeling) is either impossible or very hard for complex phenomena. Another way to represent a natural phenomenon as a model is to achieve knowledge from observed data (machine learning \cite{sarker2021machine}) or a combination of physical laws and observed data (physics-informed machine learning \cite{meng2025physics}). These three methods describe a phenomenon using a parametric model, such as state-space representations \cite{aoki2013state} and transfer functions \cite{bode1945network}. However, a phenomenon can also be described using a non-parametric model provided by impulse response techniques and frequency response techniques. As we will see in the following parts, under certain conditions, the raw observed data can also be viewed as a non-parametric model representation. One can consider a natural phenomenon as a system. Traditionally, a system is viewed as a signal processor. A signal processor accepts an input signal and produces an output signal. The rational for the separation of the signals into input and output is causality: the input causes the output. The input/output (I/O) relationship is mathematically formalized by a function mapping the input to the output, which leads to the ubiquitous notion of a system as an I/O map. The I/O map is visualized as a box with an incoming arrow for the input and an outgoing arrow for the output. One of the factors that undoubtedly contributed very much to the great success of frequency response techniques for system behavior learning was the fact that the design methods were accompanied by a very powerful technique, i.e. frequency analysis. This technique made it possible to determine the transfer functions accurately, which is precisely what is needed to apply the synthesis methods based on logarithmic diagrams. However, the frequency response and the transfer function only represent the part of the system that is completely controllable and completely observable from I/O measurements. This is why machine learning (ML) outperforms frequency-response techniques and is at the forefront of predictive model development. 

The most simple architecture of ML models is a linear/nonlinear I/O mapping function, which works well for static systems. However, the I/O map is deficient in modeling many physical phenomena, where described as dynamical systems, because it ignores the initial conditions. Considering an autonomous vehicle example, steering angle command and acceleration/deceleration (throttle/brake) command are the inputs of the system, and pose (position and orientation) is the output of the system. In order to model the autonomous vehicle as a signal processor, the pose of the vehicle depends not only on the steering angle command and the acceleration/deceleration command but also on the initial pose, that is the initial condition. Thus, modeling the autonomous vehicle as a mapping function from the steering angle and acceleration/deceleration commands to the pose of the vehicle ignores the initial condition, or equivalently, assumes zero initial pose. This illustrates the important role of the initial condition in predictive models; however, there is another issue here. Are we able to accurately predict the pose of the vehicle using the steering angle and acceleration/deceleration commands and the initial pose? The answer is NO since we are missing another important feature in our predictive model which is the initial velocity of the vehicle. The whole condition of the system is described by state (of the system) which includes both pose and velocity in the autonomous vehicle case. To achieve a perfect predictive model, we need to consider both the input and the state of the system. The state is a set of variables often arranged in the form of a vector that fully describes the system. Formalizing mathematically the effect of the state (or condition) of the system led to the state-space approach in the 1960s, which led to major results such as Kalman filter \cite{kalman1960new} and subspace identification \cite{ljung1998signal}. Consequently, the output is a function of the input and the initial state of the system. But what should we do if we do not have access to the initial state of the system?

\begin{figure}[!ht]
     \centering
     \includegraphics[width=8.8cm, height=3.2cm]{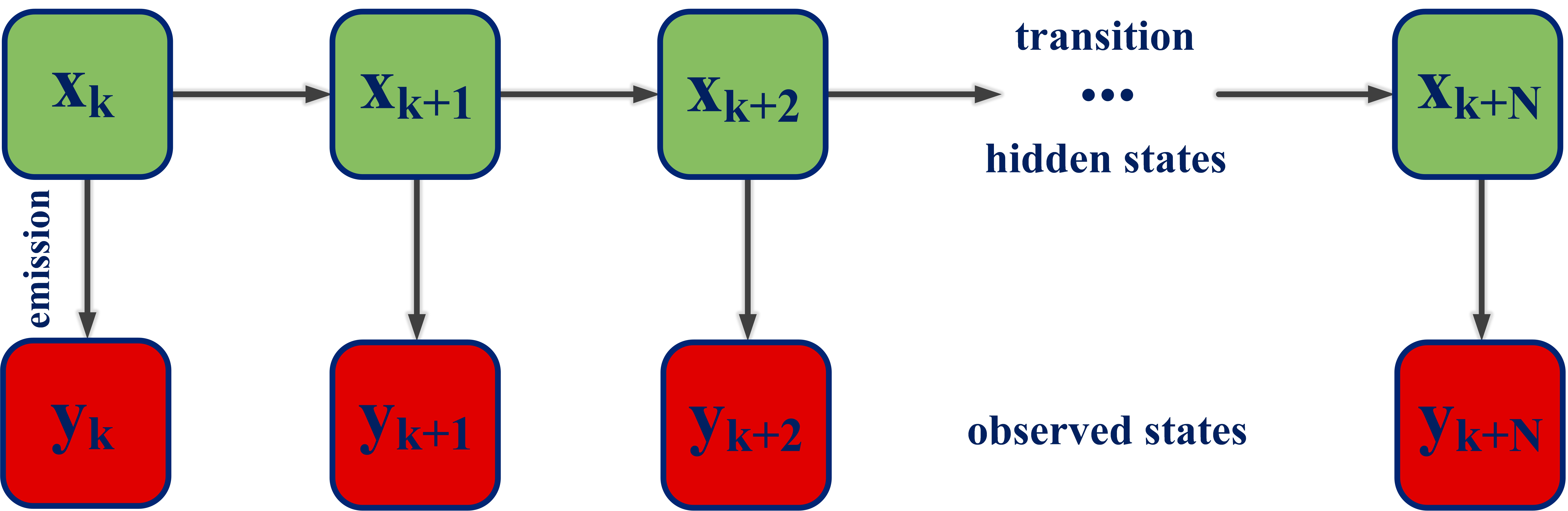}
     \caption{Markov Process: Importance of current state.}
     \label{MP}
\end{figure}

\begin{figure}[!ht]
     \centering
     \includegraphics[width=8.8cm, height=3.4cm]{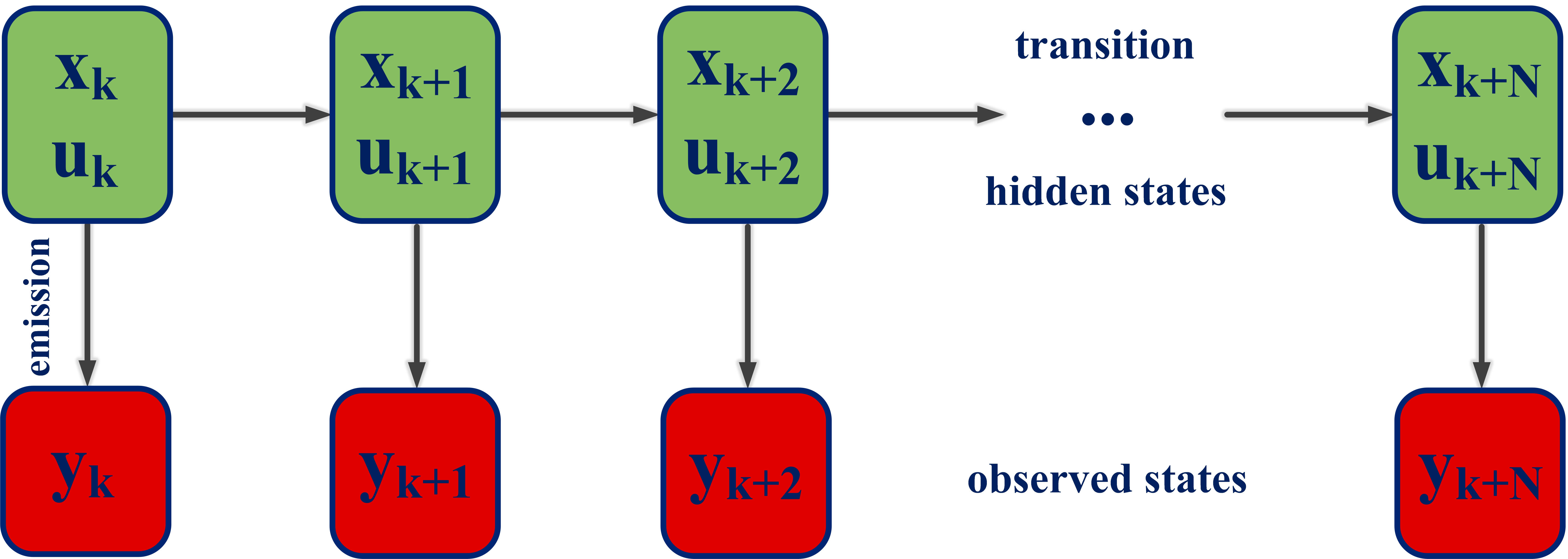}
     \caption{Markov Decision Process: Importance of current state and decisions.}
     \label{MDP}
\end{figure}

According to Markov Process (MP) \cite{dynkin1965markov}, as shown in Fig. \ref{MP}, the current state of the system $x_k$ is all we need for the system behavior prediction. $x_k$ summarizes the history of the system, i.e., previous experiences from the initial time to the current time $k$. If we must make a decision at each time step of MP, it is known as the Markov Decision Process (MDP) \cite{bellman1957markovian}, where you need both the current state and the decided action sequence (inputs of the system) to predict the system behavior, as shown in Fig. \ref{MDP}. According to MP and MDP, if we have access to the current state of the system, we can ignore the history of the system since it includes all of the previous experiences inside itself. However, the current state of the system is generally not measurable, and we need to estimate it. To accurately estimate the current state of the system, one needs the whole history of both inputs and outputs of the system, which causes a huge computation. 

\begin{figure}[!ht]
     \centering
     \includegraphics[width=8.8cm, height=2.6cm]{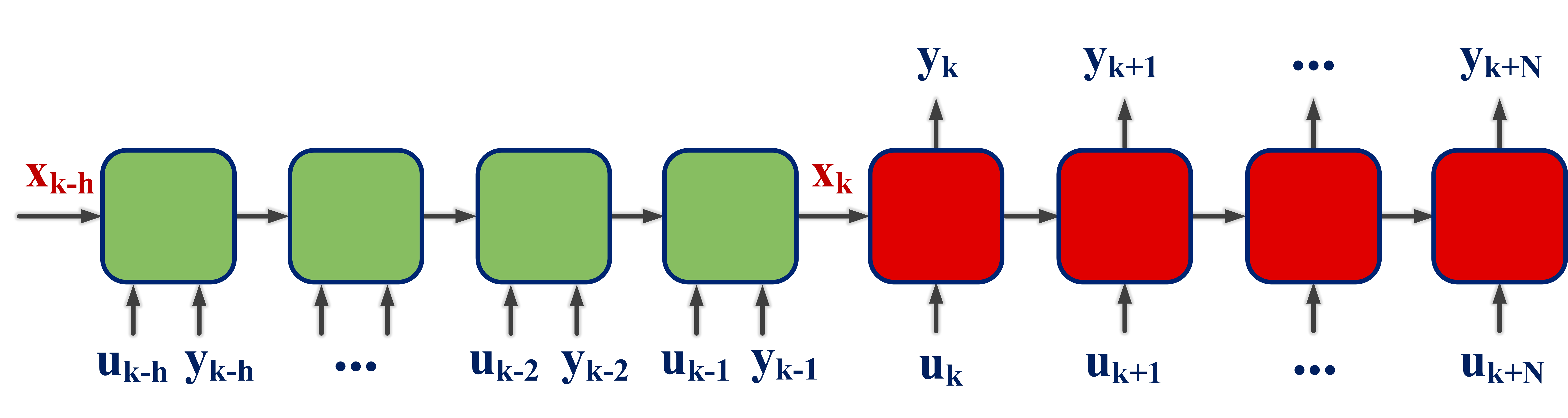}
     \caption{Memory Augmentation: Importance of history for current state estimation.}
     \label{History}
\end{figure}

\begin{figure}[!ht]
     \centering
     \includegraphics[width=8.8cm, height=5.9cm]{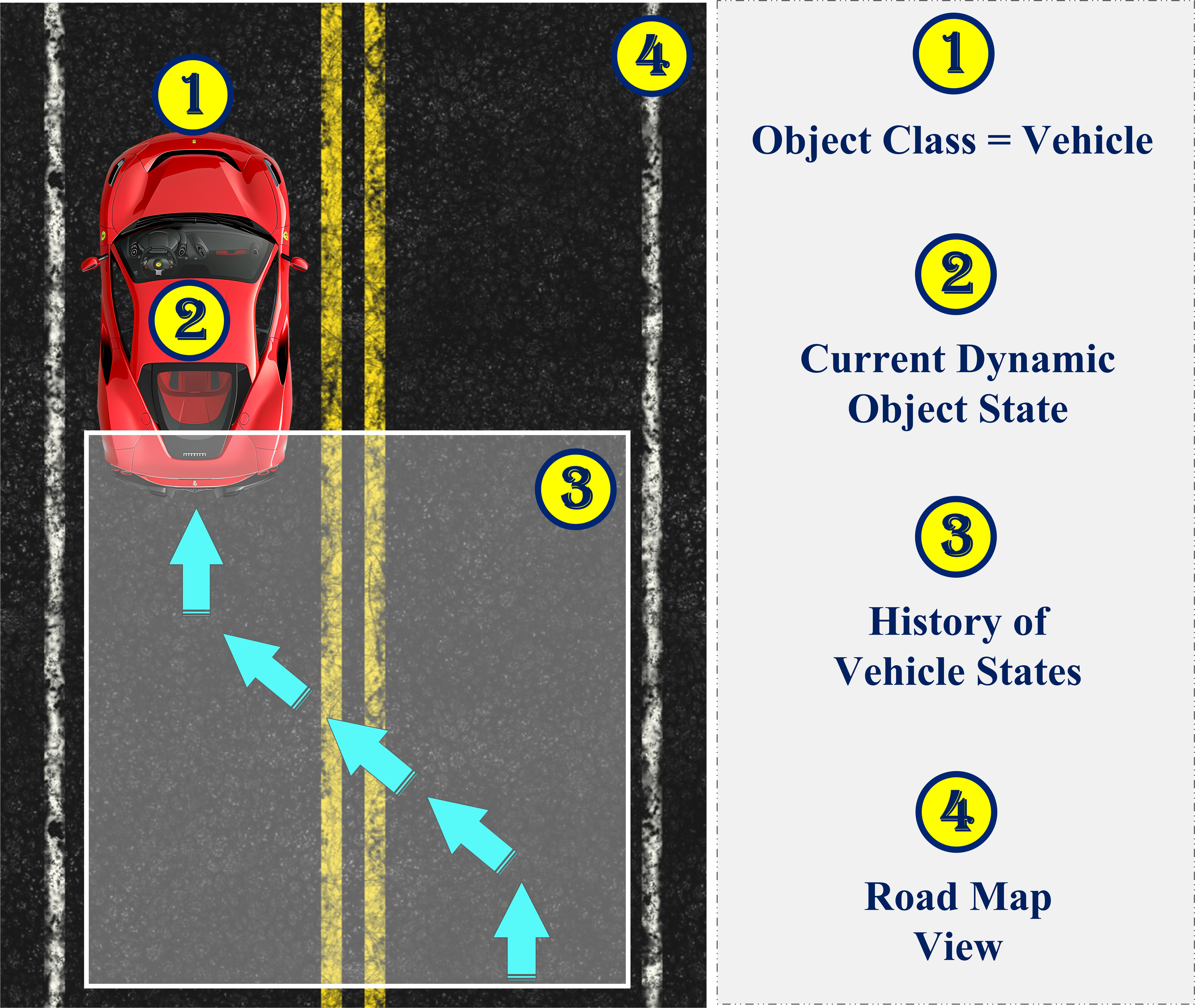}
     \caption{Memory Augmentation: Importance of history for current state estimation of dynamic object.}
     \label{History2}
\end{figure}

To overcome this issue, we consider a finite-length segment of the system’s most recent history to efficiently estimate the current state, as shown in Fig. \ref{History}. The more history we consider, the more accurate the estimation becomes, but at the cost of higher computation. One can see the use of history for dynamic object's motion prediction in the environment perception part of autonomous vehicle software, as shown in Fig. \ref{History2}. The tracker module provides not only the current pose of the dynamic object, but also the history of its path through the environment. Then, the prediction module uses this information as the current state of the dynamic object to predict its future path. Now, the question is how we can estimate the current state of the system using the most recent history of I/O data?

Consider a finite-dimensional linear time-invariant (LTI) system. We use the following input/state/output representation as:
\begin{equation}
  \begin{aligned}
    \label{system}
    & x_{k+1} = Ax_{k}+Bu_{k},\\
    & y_{k} = Cx_{k}+Du_{k},
  \end{aligned}
\end{equation}
where $k \in \mathbb{N}^+$ denotes the time step, $x \in \mathbb{R}^n$ represents the state vector, $u \in {\mathbb{R}^m}$ is the input vector, and $y \in \mathbb{R}^p$ denotes the output vector. Moreover, $A$, $B$, $C$, and $D$ are the system model matrices with appropriate dimensions. An input/state/output representation is called minimal if its order (state dimension) is as small as possible over all input/state/output representations of the system. The minimal state dimension $n(\mathcal{S})$ is called the \emph{order} of the system and is independent of the representation. Let this model be a minimal representation of order $n$. Moreover, lag $\ell(\mathcal{S})$ is called the observability index which is the smallest integer $\ell$ such that the observability matrix
\[
O_\ell(C,A) := col \bigl(C,CA,\dots,CA^{\ell-1}\bigr)
\]
has full column rank $n$, where $\ell \leq n$. Specifically, $\ell$ represents the minimum number of consecutive outputs we need to observe to uniquely reconstruct the full state of the system.

A block lower-triangular Toeplitz matrix on the impulse response is defined as:
\[
T_k(Y^{ir}) =
\begin{bmatrix}
 y^{ir}_0 & 0 & \cdots & 0 \\
 y^{ir}_1 &  y^{ir}_0 & \ddots & \vdots \\
\vdots & \ddots & \ddots & 0 \\
 y^{ir}_{k-1} & \cdots &  y^{ir}_1 &  y^{ir}_0
\end{bmatrix},
\]
where $Y^{ir} = [{y^{ir}_0}^\top, {y^{ir}_1}^\top, \dots, {y^{ir}_{k-1}}^\top]^\top$ is the impulse response, i.e., under an impulse input $u_0 = 1$, $u_k = 0$ with $k \ge 1$, and $x_0 = 0$. Therefore,
\begin{equation*}
  \begin{aligned}
    & y^{ir}_0 = D, \\
    & y^{ir}_k = C A^{k-1} B, \quad k \ge 1.
  \end{aligned}
\end{equation*}

Consequently, one can derive
\[
y_{0:k-1} = O_{k}(C,A)x_0 + T_k(Y^{ir})\,u_{0:k-1},
\]
where represents the outputs of the system based on the initial state and the given inputs.

Based on the above equation, one can predict the future behavior of the system using the current state of the system and the given inputs as follows:
\[
y_{k:k+N-1} = O_{N}(C,A)x_k + T_{N}(Y^{ir})\,u_{k:k+N-1},
\]
where the current state $x_k$ is generated by the recent sequence
\[
y_{k-h:k-1} = O_h(C,A)x_{k-h} + T_h(Y^{ir})\,u_{k-h:k-1}.
\]

Now, to obtain a unique $x_k$, we need a long enough history $h$ such that it is greater than the observability index, i.e., $h \geq \ell$. Under this condition, the observability matrix $O_h(C,A)$ is full column rank, and the above equation has a unique solution for $x_{k-h}$, which makes the uniqueness of $x_k$ according to $x_k = A^h x_{k-h} + \begin{bmatrix} A^{h-1}B & A^{h-2}B & \cdots & A\,B \end{bmatrix}\,u_{k-h:k-1}$. Consequently, we exactly obtain the current state of the LTI system using a long enough history of I/O data. However, for a data-driven system representation, as the system is unknown, we do not have $A$, $B$, $C$, and $D$ matrices to solve the equation, and also the system is probably stochastic/nonlinear/time-varying; therefore, we cannot achieve the state of the system exactly even with $h \geq \ell$. However, as we discussed above, we know that the current state of the system is a function of the recent I/O data; therefore, we can estimate it by considering a long enough history of I/O data. Another approach that proves our claim is Kalman filter and its variations, which start with a nominal estimated state and correct their state estimation through time using the I/O data.

As we discussed the important role of the most recent history of I/O data samples for estimating the current state of the system, we skip the static linear/nonlinear I/O mapping functions and start with memory-augmented linear models as the simplest choice for system dynamics modeling.

\textbf{Memory-Augmented Linear Model:} Fitting linear models to data can be achieved, both conceptually and algorithmically, by solving a system of equations $Y = WU$, where feature/input matrix $U$ and output matrix $Y$ are constructed from the collected I/O data samples, and the matrix $W$ parameterizes the model and needs to be found. Since it is a memory-augmented model not a static model, for each sample of the output matrix $Y$, the input matrix $U$ includes a signal/sequence of the corresponding input and the most recent history of I/O data with lebgth $h$ for that sample such that we have $y_i = W [u^\top_{i-h}, y^\top_{i-h}, ..., u^\top_{i-1}, y^\top_{i-1}, u^\top_{i}]^\top$. Assuming a data set in the plane, the aim of the line fitting problem is to find a line passing through the origin that best matches the given data points. Moreover, to consider a general case including offset from origin, the memory-augmented linear model is considered as:
\[
Y = WU + b,
\]
where $b$ is called the bias.

In this classical approach, the main tools are the least squares method. Geometrically, the least squares method minimizes the sum of the squared vertical distances from the data points to the fitting line. Formally, the least squares loss is defined as:
\begin{align*}
\mathcal{L}(W, b) &= \| Y - (WU + b) \|_2^2 =\sum_{i=1}^T (y_i - WU_i - b)^2,
\end{align*}
where $U_i = [u^\top_{i-h}, y^\top_{i-h}, ..., u^\top_{i-1}, y^\top_{i-1}, u^\top_{i}]^\top$, $T$ is the sample number, and the corresponding cost function is typically taken as its average,
\[
J(W, b) = \frac{1}{2T} \sum_{i=1}^T(y_i - W U_i - b)^2.
\]
The factor $\tfrac{1}{2}$ is included for convenience since it simplifies the gradient expressions. A closed-form solution exists by solving the equation
\[
Y \bar{U}^\top = \bar{W} (\bar{U} \bar{U}^\top),
\]
where $\bar{W} = [W, \hspace{1 mm} b]$ and $\bar{U} = [U^\top, \hspace{1 mm} \mathbf{1}_T]^\top$ with $\mathbf{1}_T \in \mathbb{R}^T$. Thus, the closed-form solution is
\[
\bar{W} = Y \bar{U}^\top (\bar{U} \bar{U}^\top)^{-1} ,
\]
provided that $\bar{U} \bar{U}^\top$ is invertible. However, in practice, this inversion step may not be feasible. For very large input matrices $U$, direct inversion becomes computationally expensive, and in cases where $\bar{U} \bar{U}^\top$ is ill-conditioned or singular, inversion may not even be possible. To address these limitations, iterative optimization methods such as gradient descent \cite{cauchy1847methode} are widely used to solve the model fitting problem. The gradient descent updates the parameters $W$ and $b$ iteratively by moving them in the direction opposite to the gradient of the cost function,
\begin{equation*}
  \begin{aligned}
    & W \leftarrow W - \eta \hspace{1 mm} \nabla_w J(W, b), \\
    & b \leftarrow b - \eta \hspace{1 mm} \nabla_b J(W, b),
  \end{aligned}
\end{equation*}
where $\eta > 0$ is the learning rate, controlling the step size. This approach avoids explicit matrix inversion, scales well with large datasets, and handles situations where the closed-form least squares solution is numerically unstable.

It should be noted that the LTI system \eqref{system} is identifiable if the parameter estimation is consistent \cite{records2013numerical}. A sufficient condition is that the regressor information matrix 
\begin{equation*}
      \begin{aligned}
        & I \;=\; \lim_{T \to \infty} \frac{1}{T} \sum_{i=1}^T \bar{U_i}\bar{U_i}^\top
      \end{aligned}
    \end{equation*}
is positive definite, which is satisfied with a suitable persistently exciting (PE) input data.

\begin{definition} [Persistently Exciting - Information Matrix \cite{records2013numerical}]
	The sequence $u_{1:T}$ is PE of order $K$ if the limits
    \begin{equation*}
      \begin{aligned}
        & \bar{u} = \lim_{T \to \infty} \frac{1}{T} \sum_{k=1}^{T} u_k, \\
        & R_i = \lim_{T \to \infty} \frac{1}{T}   \sum_{k=1}^{T}
          \bigl(u_k - \bar{u})(u_{k + i} - \bar{u}),
      \end{aligned}
    \end{equation*}
    exist, and the input autocorrelation information matrix $I_K =\{a_{ij} = R_{i-j}\}, \; i,j = 1, 2, ..., K$ is positive definite.
\end{definition}

\begin{lemma}[Identifiable System \cite{records2013numerical}]
\label{Identifiable System}
	To guarantee identifiability of the $n$-dimensional LTI system \eqref{system}, the input sequence must be PE of order at least $2n$, i.e., $K \ge 2n$. 
\end{lemma}

The PE of order $K$ ensures that the input data contains enough richness (frequency content, variability) to excite all system modes up to that order. If the input data is not sufficiently exciting, then the system dynamics cannot be identified consistently. $K \ge 2n$ arises because the regression vector $U$ involves past I/O data, and the correlation structure requires excitation of up to twice the system order to make the regressor information matrix $I$ nonsingular (positive definite) \cite{ljung1998signal}. Therefore, a positive-definite input autocorrelation information matrix $I_K$ with $K \ge 2n$ generates a positive-definite regressor information matrix $I$. The PE condition is sufficient to ensure consistent parameter estimation for least squares, maximum likelihood, and maximum likelihood with white measurement noise.

\vspace{2 mm}
\textbf{Memory-Augmented Deep Neural Network:} Consider a general unknown nonlinear system. We use the following general input/state/output representation as:
\begin{equation}
  \begin{aligned}
    \label{system-Nonlinear}
     &x_{k+1} = f(x_{k},u_{k},v^p_{k}),\\
     &y_{k} = g(x_{k},u_{k},v^m_{k}),
  \end{aligned}
\end{equation}
where $v^p \in \mathbb{R}^n$ and $v^m \in \mathbb{R}^p$ denote the process and measurement noises, respectively. In addition, $f:\mathbb{R}^n\times \mathbb{R}^m \times \mathbb{R}^n \rightarrow \mathbb{R}^n$ indicates the system dynamics, and $g:\mathbb{R}^n\times \mathbb{R}^m \times \mathbb{R}^p \rightarrow \mathbb{R}^p$ denotes the output dynamics.

\begin{figure}[!ht]
     \centering
     \includegraphics[width=8.8cm, height=3.2cm]{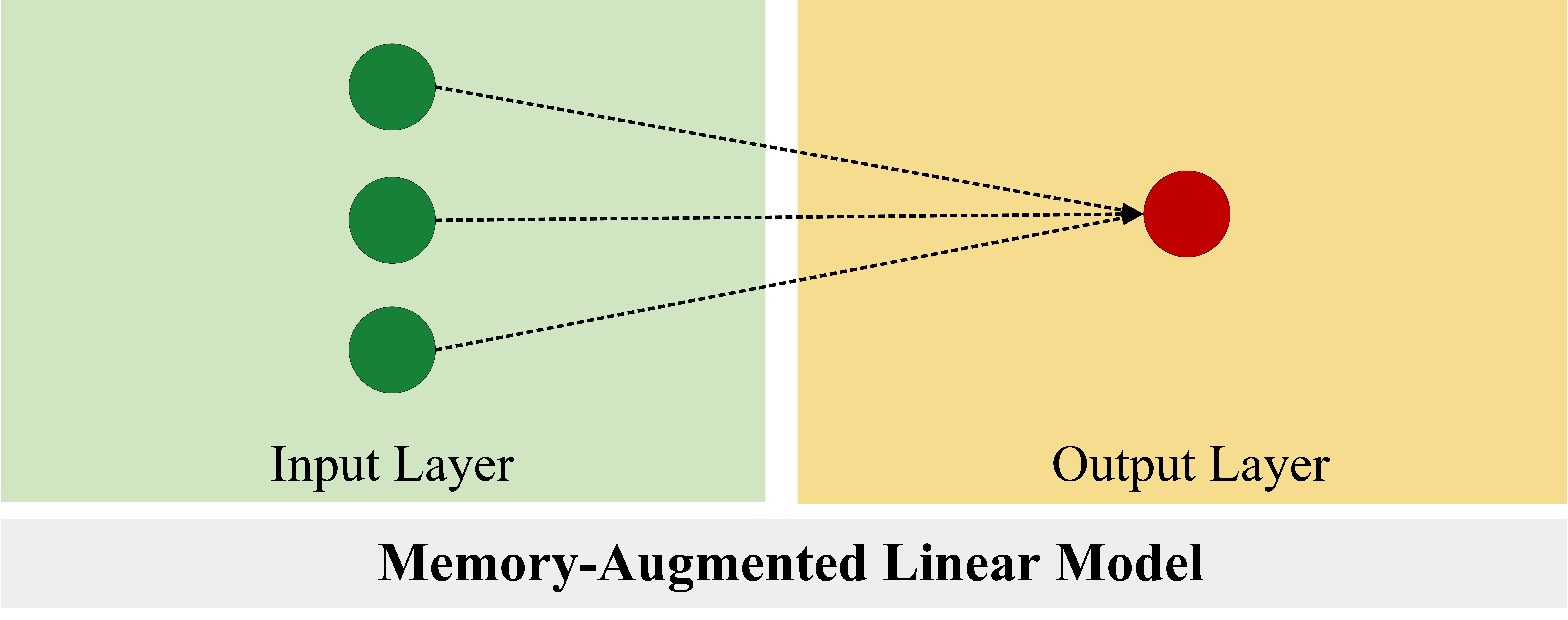}
     \includegraphics[width=8.8cm, height=4.2cm]{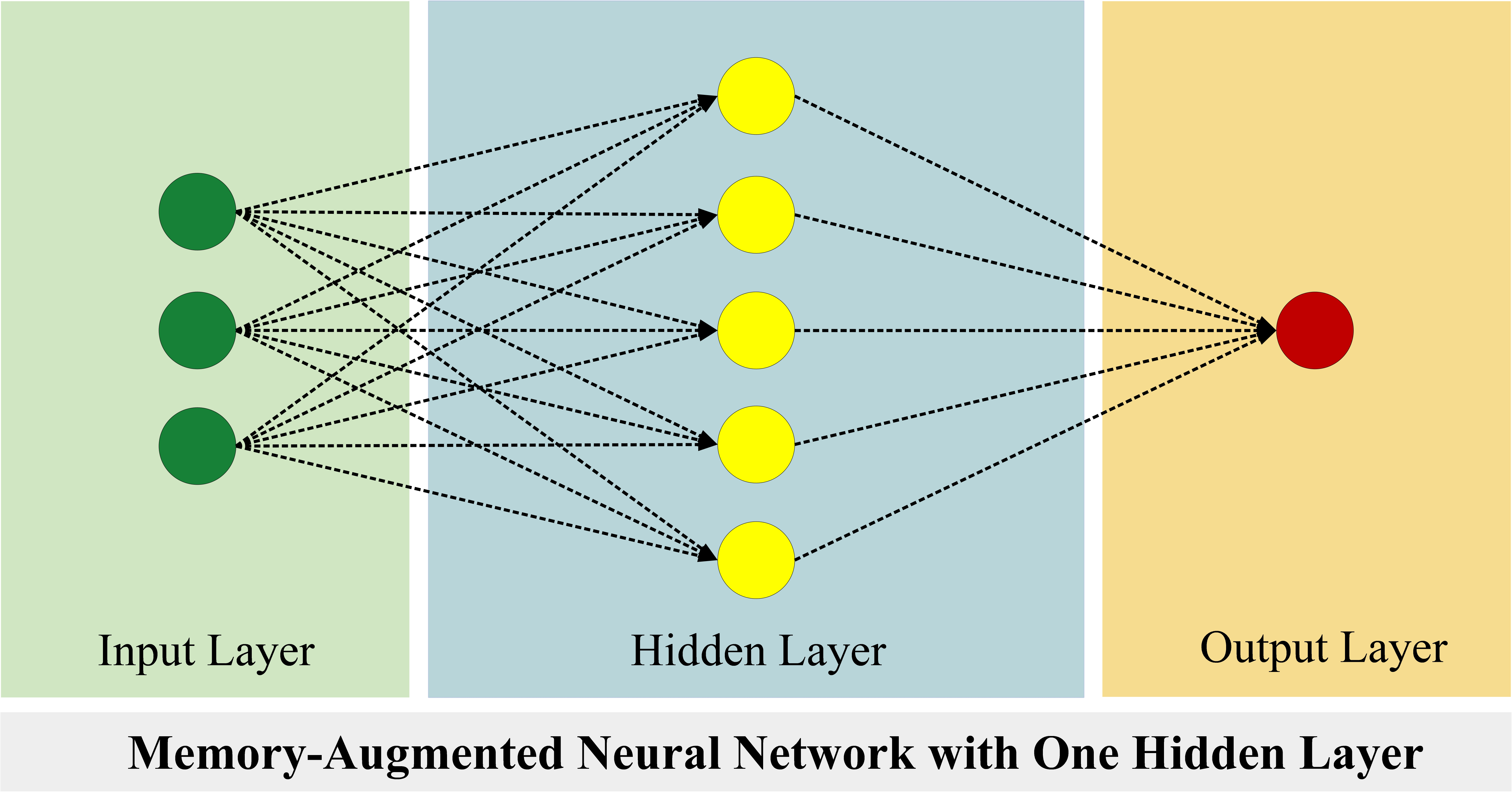}
     \includegraphics[width=8.8cm, height=4.2cm]{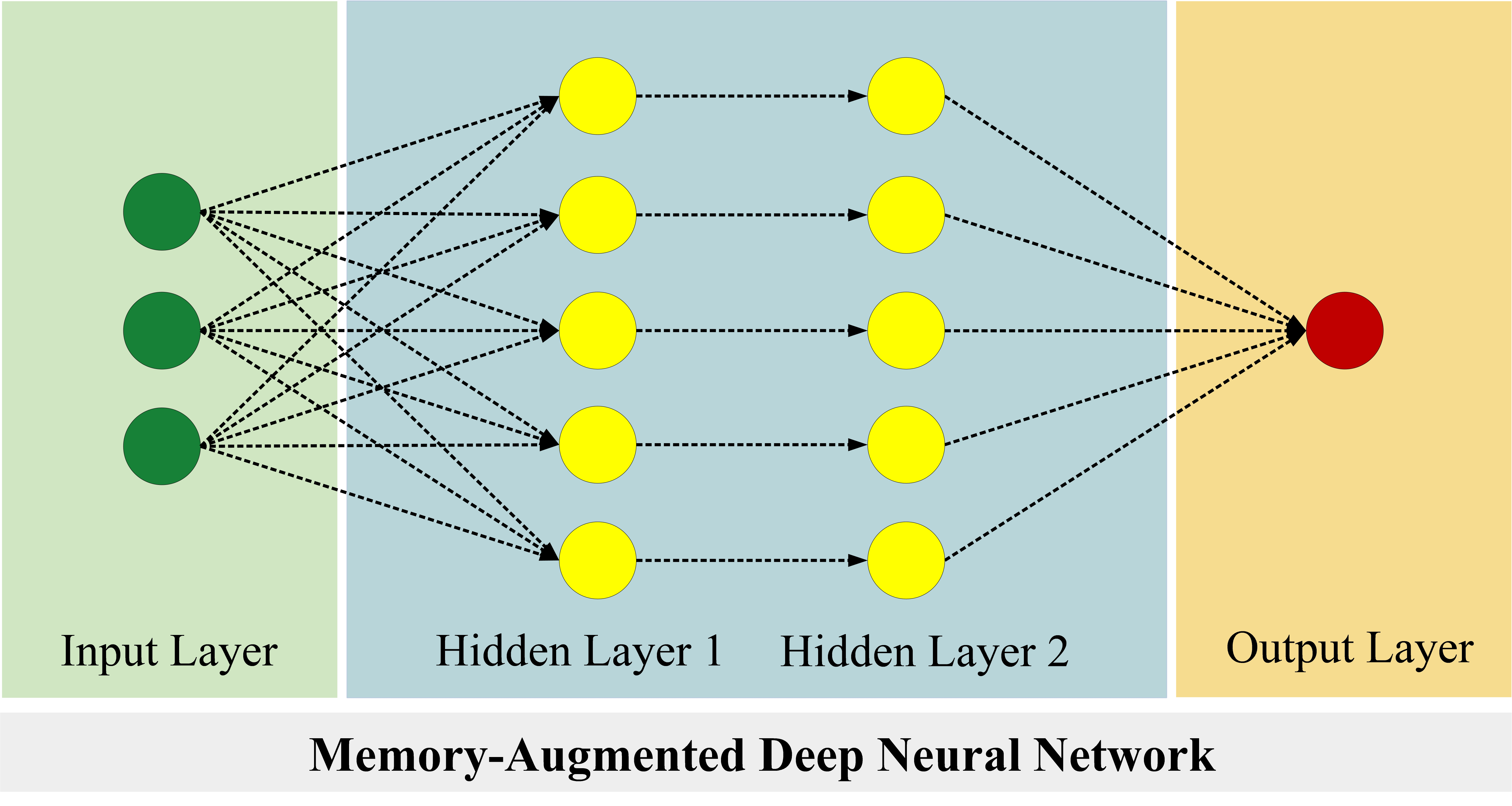}
     \caption{Memory-Augmented Linear Model, Memory-Augmented Neural Network, and Memory-Augmented Deep Neural Network.}
     \label{Machine Learning Models}
\end{figure}

The memory-augmented linear model, which assumes a fixed linear relationship between inputs and outputs, cannot learn the behavior of the nonlinear system \eqref{system-Nonlinear} accurately and shows a high bias (underfitting). Deep neural network (DNN) is a subfield of machine learning that models high-level abstractions in data using architectures composed of multiple layers of nonlinear processing units, or neurons \cite{fukushima1980neocognitron, rumelhart1986learning}.  
By stacking many such layers, these networks can automatically discover rich features and patterns without manual engineering. 
Unlike linear models, DNNs can capture complex nonlinear dependencies, making them highly expressive and flexible. Structurally, as illustrated in Fig. \ref{Machine Learning Models}, a neural network is composed of an input layer, one or more hidden layers, and an output layer. A simple linear model corresponds to a network with no hidden layers, whereas a DNN contains many hidden layers. Each hidden layer is composed of neurons that apply a linear transformation followed by a nonlinear activation function. The linear transformation maps the input of the hidden layer $l$, i.e., $A^{[l-1]}$ (the output of the previous hidden layer), to $Z^{[l]}$:
\[
Z^{[l]} = W A^{[l-1]} + b, \quad   0 < l \leq \mathcal{L},
\]
where $A^{[0]}$ and $A^{[\mathcal{L}]}$ are the collected inputs $U$ and outputs $Y$, respectively. Then, the nonlinear activation functions try to capture the nonlinear behavior of the system \eqref{system-Nonlinear}, where sigmoid, hyperbolic tangent \cite{rumelhart1986learning}, and ReLU \cite{nair2010rectified} are the most popular activation functions:
\begin{itemize}
  \item Sigmoid: \[
A^{[l]} = \sigma(Z^{[l]}) = \frac{1}{1 + e^{-Z^{[l]}}}
\]
  \item Hyperbolic Tangent: \[
A^{[l]} =  \tanh(Z^{[l]}) = \frac{e^{Z^{[l]}} - e^{-Z^{[l]}}}{e^{Z^{[l]}} + e^{-Z^{[l]}}}
\]
  \item Rectified Linear Unit (ReLU): \[
A^{[l]} = \text{ReLU}(Z^{[l]}) = \max(0, Z^{[l]})
\]
\end{itemize}

The hidden layers act as an automatic feature engineering mechanism, with each layer progressively transforming the input into increasingly abstract and higher-level representations. For instance, in robotic and autonomous vehicle perception, lower layers may capture geometric features such as edges, corners, or motion cues from raw sensor data, while deeper layers extract higher-level representations such as obstacles, drivable regions, or semantic categories of surrounding objects. 
Despite their power, training DNNs introduces challenges. One central issue is bias-variance trade-off. Networks with too few layers or parameters may suffer from underfitting (high bias), while overly large networks can overfit the training data (high variance). Regularization approaches such as $\ell_1$ or $\ell_2$ penalties \cite{lecun1989backpropagation}, early stopping \cite{prechelt2002early}, and dropout \cite{srivastava2014dropout} are commonly used to mitigate overfitting. In addition, DNN learns to transform raw input (i.e., $U$) into increasingly abstract and useful internal representations through a process called backpropagation \cite{rumelhart1986learning}, which iteratively adjusts the connection weights via the gradient descent to minimize the learning error. However, this training process introduces another challenge, known as the vanishing and exploding gradient problem, where gradients either shrink toward zero or grow uncontrollably as they are propagated back through many layers, leading to unstable optimization. Careful weight initialization, appropriate activation functions (e.g., ReLU variants), and normalization techniques are commonly employed to alleviate this issue. 

Optimization in DNN often relies on variants of gradient descent. In the standard batch gradient descent method, parameters are updated after computing the gradient over the entire dataset, which can be computationally expensive for large-scale problems. To address this, stochastic gradient descent \cite{robbins1951stochastic} updates the parameters after computing the gradient on a single randomly chosen training example at each iteration. While this introduces noise in the updates, it significantly reduces computation per iteration and can help the optimizer escape shallow local minima. A common compromise between these two extremes is mini-batch gradient descent \cite{lecun2002efficient}, which processes small subsets of the data at a time, thereby balancing computational efficiency and the stability of convergence. Enhancements to mini-batch/stochastic gradient descent include the use of momentum \cite{polyak1964some} which accelerates convergence by dampening oscillations, RMSProp \cite{tieleman2012rmsprop} which adapts learning rates for each parameter based on recent gradient magnitudes, and Adam \cite{kingma2014adam} which combines momentum and RMSProp to provide robust and efficient convergence across a wide range of tasks. Finally, batch normalization \cite{ioffe2015batch} has become a standard component of modern architectures. It normalizes intermediate activations within each mini-batch, reducing internal covariate shift, stabilizing training, and allowing the use of higher learning rates. This technique not only speeds up training but also improves generalization, making it easier to train very deep networks. Fueled by large datasets, powerful GPUs, and these optimization and regularization techniques, deep learning has emerged as the foundation of most state-of-the-art AI systems. While linear models remain valuable for their simplicity and interpretability, DNNs dominate when the goal is to capture complex, nonlinear patterns in high-dimensional data.

\begin{figure}[!ht]
     \centering
     \includegraphics[width=8.8cm, height=7.4cm]{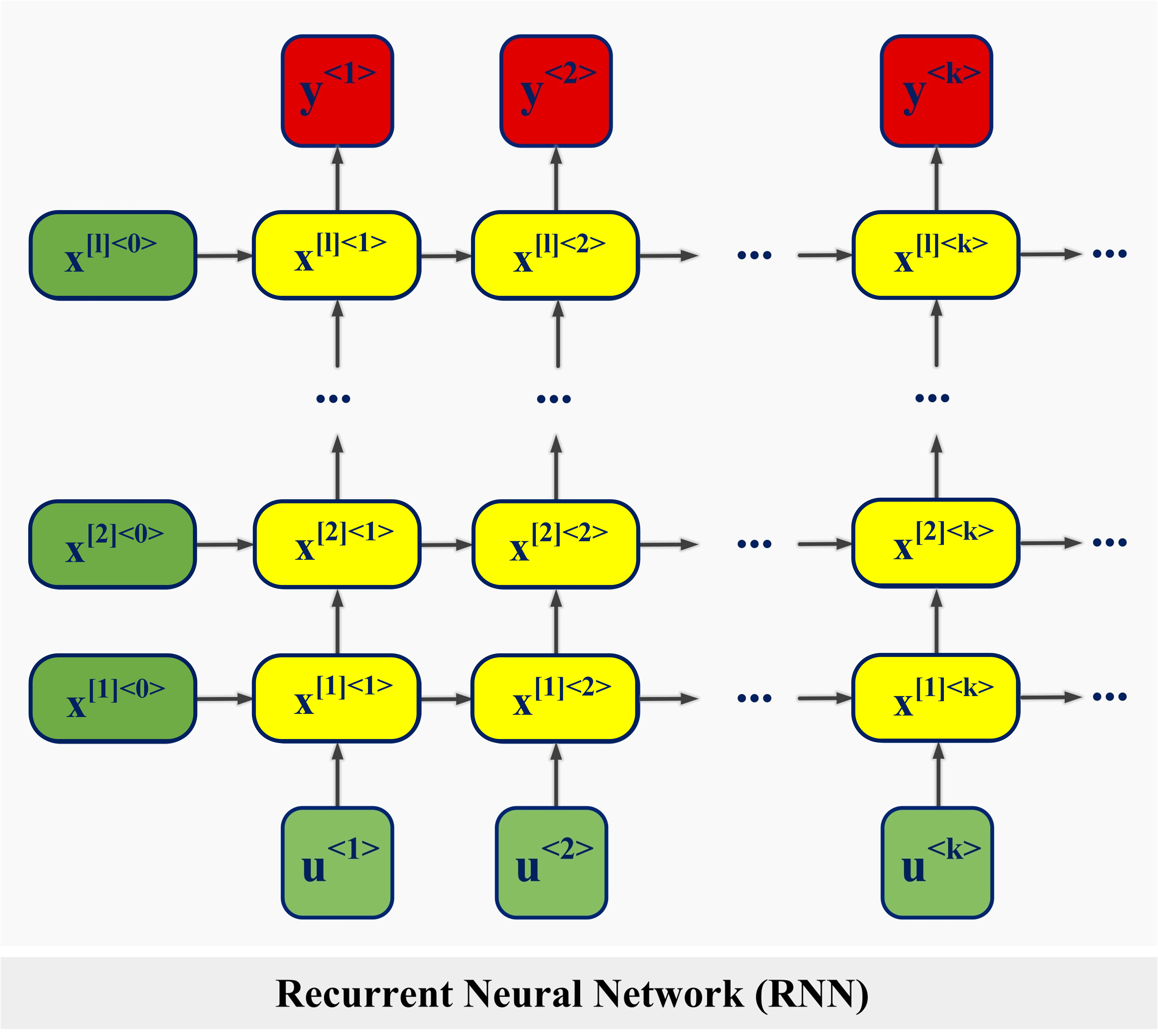}
     \includegraphics[width=8.8cm, height=7.4cm]{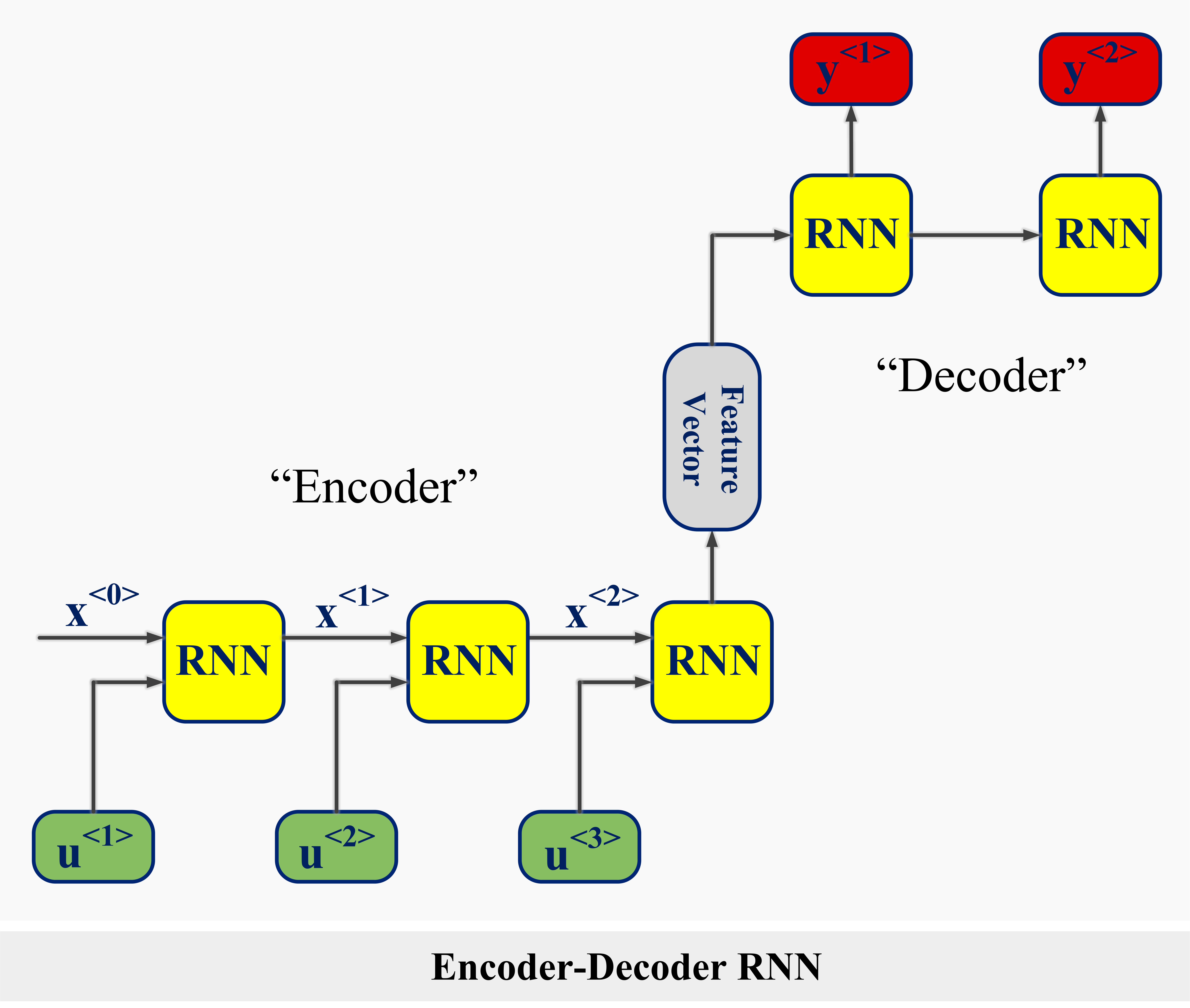}
     \includegraphics[width=8.8cm, height=5.6cm]{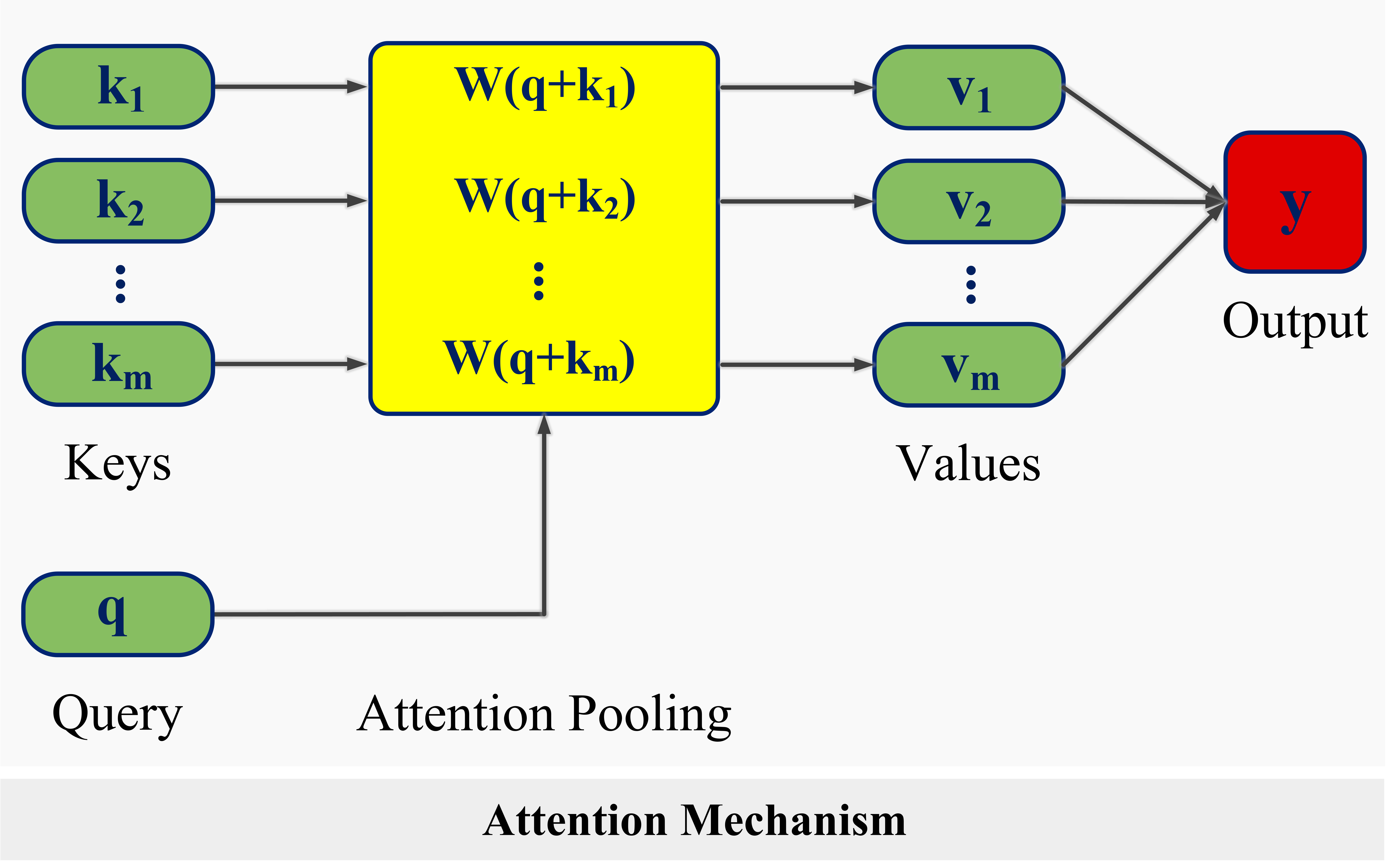}
     \caption{Recurrent Neural Network (RNN), Encoder-Decoder RNN, and Attention Mechanism.}
     \label{RNN Models}
\end{figure}

\textbf{Recurrent Neural Network:} Recurrent neural network (RNN) \cite{williams1989learning}, especially when enhanced with attention mechanisms \cite{bahdanau2014neural} (focusing on the most relevant parts of the sequence), pushes dynamics modeling even further. Inspired by the state-space models, RNN maintains a hidden state that evolves over time to effectively capture context and order:
\begin{equation*}
  \begin{aligned}
     &x^{[l] <i>} = \psi \big( W_{xx} x^{[l] <i-1>} + W_{xu} x^{[l-1] <i>} + b_x \big), \quad 0 < i \leq k,\\
     &y^{<i>} = W_{yx} x^{[\mathcal{L}] <i>} + b_y,
  \end{aligned}
\end{equation*}
where $x^{[l] <0>} = 0$, $x^{[0] <i>} = u^{<i>}$, and $\psi$ is the nonlinear activation function, as illustrated in Fig. \ref{RNN Models}.

Vanilla RNNs often struggle with long-term dependencies due to the vanishing and exploding gradient problems. To overcome these limitations, advanced variants such as Long Short-Term Memory (LSTM) \cite{hochreiter1997long} and Gated Recurrent Unit (GRU) \cite{cho2014learning} were developed. LSTMs introduce a sophisticated gating mechanism composed of input, forget, and output gates, along with a dedicated cell state. This structure allows the network to selectively retain or discard information over long time horizons, making LSTMs highly effective for long-sequence forecasting. GRUs, on the other hand, simplify the gating mechanism by combining the input and forget gates into a single update gate and merging the cell state with the hidden state. This makes GRUs computationally lighter and faster to train, while still retaining the ability to capture long-term dependencies, often achieving performance comparable to LSTMs in practice. Building upon these recurrent structures, Encoder-Decoder architectures \cite{cho2014learning, sutskever2014sequence} extend dynamics modeling to input-output mapping tasks where the input and output sequences may differ in length. The encoder processes the input sequence into a fixed-length hidden representation (context vector), while the decoder generates the output sequence step by step, conditioned on this context. This framework has proven especially powerful in multi-step time-series prediction. With the integration of attention mechanisms, Encoder-Decoder models overcome the limitations of fixed-length context vectors by allowing the decoder to dynamically focus on different parts of the input sequence at each decoding step, greatly improving performance on long and complex sequences. While these models provide strong capabilities in capturing temporal dynamics and context, they come with trade-offs: higher complexity, increased computational overhead, and sensitivity to hyperparameter choices. Nonetheless, LSTMs, GRUs, and Encoder-Decoder frameworks remain foundational in modern time-series modeling and continue to inspire more advanced architectures.

In parallel, alternative dynamics modeling frameworks have been developed. One notable example is Prophet \cite{taylor2018forecasting}, which emphasizes interpretability and usability in time-series applications. 
Another emerging alternative is Mamba \cite{gu2023mamba}, a structured state-space model (SSM) designed for efficient long-sequence modeling. Mamba uses fast recurrent-style updates with linear-time complexity, making it more memory- and compute-efficient for very long sequences. Recently, Transformers \cite{vaswani2017attention} have revolutionized time-series modeling. Transformers rely entirely on self-attention mechanisms instead of recurrence, allowing them to model long-range dependencies in parallel and with greater efficiency than RNNs. Their scalability, combined with pretraining on massive datasets, makes them the backbone of modern time-series modeling. However, Transformers are computationally demanding, with training requiring significant GPU resources and careful optimization. 

\textbf{Behavioral Systems Theory:}
In the classical methods for data fitting, the choice of the system representation determines the fitting criterion. This feature of the classical methods is undesirable, and it is more natural to specify a desired fitting criterion instead of a system representation. The underlying goal is that given a set of points in $\mathbb{R}^d$, find a subspace of $\mathbb{R}^d$, i.e., $\mathcal{B} \subset \mathbb{R}^d$, that has the least $2$-norm distance to all points. Such a subspace is an optimal fitting model (in the $2$-norm sense). The most general way to represent any subspace in $\mathbb{R}^d$ is to use the image or kernel of the data matrix, as the classical least squares exclude some subspaces \cite{markovsky2008structured}. This suggests that the image and kernel representations are better suited for data fitting. Contrary to the common perception that a model (system representation) is an equation, e.g., $Y = WU$, behavioral systems theory \cite{markovsky2021behavioral} views a model as a set, e.g., a line passing through the origin in the line fitting problem. The dimension of the subspace $\mathcal{B}$ is a measure for the complexity of the model. The larger the subspace is, the more complicated (and therefore less useful) the model is. However, the larger the subspace is, the better the fitting accuracy could be, so that there is a trade-off between complexity and accuracy. The behavioral systems theory decouples the system behavior from its representation and works with trajectories (signals) instead of data samples. It considers a system as a set of trajectories, which is applied to discrete-event as well as continuous-event, discrete-time as well as continuous-time, linear as well as nonlinear, time-invariant as well as time-varying, and lumped as well as distributed systems. A dynamical system postulates which signals $w$ from the universe of signals are possible to be observed. Those signals that are possible to be observed are called trajectories of the system. The set of all trajectories, denoted by $\mathcal{B}$, is called the system behavior. We identify the system with its behavior and use the terms system and behavior interchangeably. In signal processing and control, we need to specify that "a signal $w$ is a trajectory of a system $\mathcal{B}$".

Behind every linear model learning problem, there is a (hidden) low-rank approximation problem. The model imposes relations on the data, which renders the data matrix rank deficient. Therefore, fitting data by a linear model (bounded complexity) is equivalent to the low-rank approximation of the data matrix. The rank constraint in the low-rank approximation problem corresponds to the constraint that the data are fitted by a linear model. Therefore, the question of representing the rank constraint in the low-rank approximation problem corresponds to the question of choosing the model representation in the data fitting problem. A low-rank approximation by an unstructured data matrix corresponds to fitting the data by a static LTI model, and a low-rank approximation by a Hankel structured matrix corresponds to fitting the data by a dynamic LTI model. A key result of behavioral systems theory is that under the PE condition, the image of the Hankel matrix (column/range space) equals the set of finite-length trajectories of the LTI system, i.e., the system behavior $\mathcal{B}$. The behavioral approach is naturally suited for direct data-driven optimal policy because it separates the notion of a system from a representation. In contemporary ML language, the behavioral approach is non-parametric and unsupervised since the data does not have to be labeled into inputs and outputs. Moreover, it is conceptually simple, intuitively clear, and computationally efficient. 


\begin{definition}[Hankel Matrix]
\label{Hankel Matrix_w}
Given a signal $w\in \mathbb{R}^q$, the Hankel Matrix of depth $K \leq T$ is defined as:
\begin{equation}
  \begin{aligned}
    \label{Hankel_w}
    & \mathcal{H}_K(w_{1:T}):= \begin{bmatrix}
    w_1 & w_2 & \cdots & w_{T-K+1}\\ 
    w_2 & w_3 & \cdots & w_{T-K+2}\\
    \vdots & \vdots & \ddots & \vdots\\ 
    w_K & w_{K+1} & \cdots & w_T
    \end{bmatrix}.
  \end{aligned}
\end{equation}
Let $L= T-K+1$, then we have $\mathcal{H}_K(w_{1:T})\in \mathbb{R}^{qK\times L}$.
\end{definition}

\begin{definition} [Persistently Exciting - Hankel Matrix \cite{willems2005note}]
	The sequence $u_{1:T}$ is PE of order $K$ if $\mathcal{H}_{K}(u_{1:T})$ has full row rank, i.e., $rank(\mathcal{H}_{K}(u_{1:T})) = mK$. 
\end{definition}

\begin{lemma}[Fundamental Lemma \cite{willems2005note}]\label{Fundamental Lemma}
	Consider the deterministic LTI system \eqref{system} as a controllable system with a pre-collected I/O sequence $(u_{1:T}^{\mathrm{d}}, y_{1:T}^{\mathrm{d}})$ of length $T$. Providing a PE input sequence $u^{\mathrm{d}}_{1:T}$ of order $K+n$, any length-$K$ sequence  $(u_{1:K}, y_{1:K})$ is an I/O trajectory of the LTI system if and only if we have
	\begin{equation} \label{fundamental lemma_w}
		\begin{bmatrix}
			\mathcal{H}_{K}(u^{\mathrm{d}}_{1:T}) \\
			\mathcal{H}_{K}(y^{\mathrm{d}}_{1:T})
		\end{bmatrix} g = \begin{bmatrix}
			u_{1:K} \\ y_{1:K}
		\end{bmatrix}
	\end{equation}
	for some real vector $g \in \mathbb{R}^L$.
\end{lemma}

\begin{remark} [Rank of Hankel Matrix]
\label{Rank of Hankel Matrix_w}
Considering $r:= \rm{rank}\left(
			\mathcal{H}_{K}(u^{\mathrm{d}}_{1:T},y^{\mathrm{d}}_{1:T}) \right)$, a PE control input sequence of order $K+n$ ensures that $mK+1 \leq r \leq mK+n$ and requires $T \ge (m + 1)(K + n) - 1$. Furthermore, $r = mK+n$ if $K \geq \ell$, where $\ell$ is the observability index. See \cite{markovsky2021behavioral} for more details.
\end{remark}

The Fundamental Lemma requires a long, continuous trajectory to construct the Hankel matrix. The result in \cite{van2020willems} extends this framework to multiple short trajectories, which we refer to as the Extended Fundamental Lemma. This extension is formulated through the mosaic-Hankel matrix, which provides a structured representation for datasets composed of multiple short, discontinuous trajectories.

\begin{definition}[Mosaic-Hankel Matrix] 
\label{Mosaic Hankel Matrix}
Given a set of $z$ discontinuous signals $ \mathcal{W} = \{w_{1:T_{1}}^{1}, \cdots, w_{1:T_{z}}^{z} \}$, the mosaic-Hankel matrix of depth $K \leq \min(T_1, \cdots, T_{z})$ is defined as:
\begin{equation*}
  \begin{aligned}
    \label{Mosaic Hankel w}
    & \mathcal{M}_K(\mathcal{W}) = [\mathcal{H}_K(w_{1:T_1}^{1}), \mathcal{H}_K(w_{1:T_2}^{2}), \cdots, \mathcal{H}_K(w_{1:T_z}^{z})].
  \end{aligned}
\end{equation*}
Let $T=\sum_{i=1}^{z} T_{i}$ and $L=T-z(K-1)$, then we have $\mathcal{M}_K(\mathcal{W}) \in \mathbb{R}^{qK \times L}$.
\end{definition}

\begin{lemma}[Extended Fundamental Lemma \cite{van2020willems}]
\label{Discontinuous Fundamental Lemma}
Consider the deterministic LTI system \eqref{system} with a pre-collected I/O sequence $(u_{1:T}^{\mathrm{d}}, y_{1:T}^{\mathrm{d}})$ of length $T$ consisting of multiple discontinuous signal trajectories. Providing a rank condition 
\begin{equation*}
  \begin{aligned}
    & \rm{rank}\left(\begin{bmatrix}
			\mathcal{M}_{K}(u^{\mathrm{d}}_{1:T}) \\
			\mathcal{M}_{K}(y^{\mathrm{d}}_{1:T})
		\end{bmatrix}\right) = mK+n,
  \end{aligned}
\end{equation*} 
where $K \geq l$, any length-$K$ sequence  $(u_{1:K}, y_{1:K})$ is an I/O trajectory of the LTI system if and only if we have
	\begin{equation*}
		\begin{bmatrix}
			\mathcal{M}_{K}(u^{\mathrm{d}}_{1:T}) \\
			\mathcal{M}_{K}(y^{\mathrm{d}}_{1:T})
		\end{bmatrix} g = \begin{bmatrix}
			u_{1:K} \\ y_{1:K}
		\end{bmatrix}
	\end{equation*}
for some real vector $g \in \mathbb{R}^L$.
\end{lemma}

In the Extended Fundamental Lemma, the rank condition on the mosaic-Hankel matrix guarantees that $u^d_{1:T}$ is PE of order $K+n$ and requires $T \geq (m+z)K+n-z$. It is worth noting that the Hankel matrix \eqref{Hankel_w} represents a special case of the mosaic-Hankel matrix with $z=1$. In addition, the Extended Fundamental Lemma is more sample-efficient than the Fundamental Lemma, requiring $mn$ fewer data samples under $z=1$. In this survey, we focus on the Fundamental Lemma; however, all materials can be easily applied to the Extended Fundamental Lemma. For the Hankel matrix $\mathcal{H}$, we denote its image (column/range space) by $\operatorname{im}(\mathcal{H})$ and its kernel (null space) by $\operatorname{ker}(\mathcal{H})$. When a basis of $\operatorname{ker}(\mathcal{H})$ needs to be computed, we write $\mathcal{N} = \operatorname{null}(\mathcal{H})$, which denotes a matrix $\mathcal{N}$ of appropriate dimensions such that $\mathcal{H} \mathcal{N} = 0$. Consequently, a finite‐dimensional LTI system $\mathcal{B}$ admits two representations: 

\hspace{-3.5 mm}\textbf{Image Representation:} By exploiting the image structure (column space, i.e., all linear combinations of the columns) of the Hankel matrix $\operatorname{im}(\mathcal{H})$, the Fundamental Lemma establishes that the Hankel matrix spans the space of all length-$K$ trajectories generated by a deterministic LTI system, under the conditions that the collected input sequence is PE of order $K+n$ and the underlying system is controllable.

\hspace{-3.5 mm}\textbf{Kernel Representation:} By exploiting the kernel structure (null space) of the Hankel matrix $\operatorname{ker}(\mathcal{H})$, one can retrieve a kernel representation of a deterministic LTI system directly from the data. We will provide more details on the kernel representation in Section IV.

To employ the Fundamental Lemma for simulation and control, \cite{markovsky2008data} introduces the notions of the initial trajectory and the prediction trajectory. Accordingly, the state-space model \eqref{system} is replaced by the algebraic equation \eqref{fundamental lemma_w} for the deterministic LTI system, which relates a length-$T_{ini}$ initial I/O trajectory and a length-$N$ predicted I/O trajectory using the collected data encoded in the Hankel matrix. Here, $T_{ini} \ge \ell$ plays the role of the history $h \ge \ell$ in the parametric models (i.e., state-space model and RNN), yielding a unique current state estimate and, consequently, a unique prediction. To illustrate this version of the Fundamental Lemma \eqref{fundamental lemma_w}, we set $K = T_{ini}+N$ and partition the Hankel matrices into Past and Future subblocks as:
\begin{equation*}
  \begin{aligned}
    &\begin{bmatrix}
      U_P\\ U_F 
     \end{bmatrix} =: \mathcal{H}(u^d), \hspace{5 mm}
     \begin{bmatrix}
      Y_P\\ Y_F 
     \end{bmatrix} =: \mathcal{H}(y^d),
  \end{aligned}
\end{equation*}
where $U_P \in \mathbb{R}^{mT_{ini} \times L}$, $U_F \in \mathbb{R}^{mN \times L}$, $Y_P \in \mathbb{R}^{pT_{ini} \times L}$, and $Y_F \in \mathbb{R}^{pN \times L}$.

\begin{figure}[!ht]
     \centering
     \includegraphics[width=8.8cm, height=7cm]{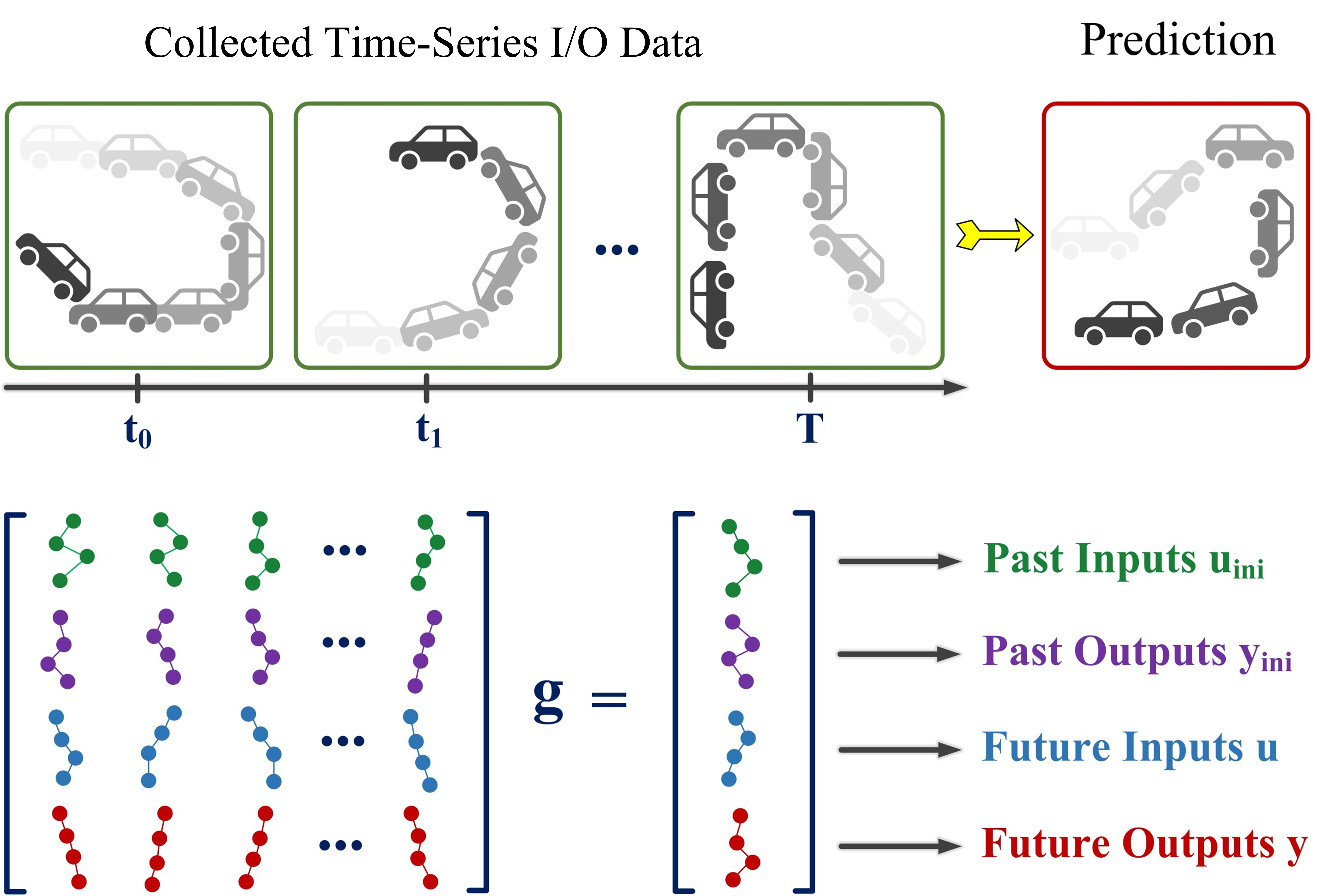}
     \caption{Behavioral Systems Theory: Fundamental Lemma for simulation and control.}
     \label{Behavioral System Theory}
\end{figure}

Now, as illustrated in Fig. \ref{Behavioral System Theory}, for a given initial trajectory $(u_{ini}, y_{ini})$, one can rewrite the Fundamental Lemma \eqref{fundamental lemma_w} as:
\begin{equation}
  \begin{aligned}
    \label{data-driven sc}
    &\begin{bmatrix}
      U_P\\ Y_P\\ U_F\\ Y_F 
     \end{bmatrix} g = 
     \begin{bmatrix}
      u_{ini}\\ y_{ini}\\ u\\ y 
     \end{bmatrix},
  \end{aligned}
\end{equation}
where $(u_{ini}, y_{ini})$ is the length-$T_{ini}$ initial trajectory, and $(u,y)$ is the length-$N$ future trajectory. Moreover, the fixed data matrices $U_P$, $Y_P$, $U_F$, and $Y_F$ are obtained offline (from collected data). For the general nonlinear system \eqref{system-Nonlinear}, we will present a modified version of the algebraic equation \eqref{data-driven sc} in the next section.

\section{Data-Driven Optimal Policy}
The pursuit of data-driven optimal policies is reshaping control, moving beyond purely model-based traditions toward frameworks that learn directly from data. One direction enhances model predictive control (MPC) with machine learning (ML), using learned dynamics. Reinforcement learning (RL) offers a fully data-driven alternative, achieving strong performance but often with poor sample efficiency and safety concerns. Hybrid MPC-based RL methods combine MPC’s stability and safety with RL’s adaptability, while data-enabled predictive control (DeePC) bypasses modeling entirely, deriving policies directly from I/O trajectories. Together, these approaches highlight diverse strategies for leveraging data in optimal control policy, each balancing reliability, scalability, and robustness in distinct ways. In the following, we examine each of these approaches in detail, analyzing their underlying principles, practical benefits, and limitations.

\vspace{2 mm}
\textbf{Machine Learning-based Model Predictive Control:}
Considering the discrete-time LTI system \eqref{system}, $\mathcal{X}$, $\mathcal{U}$, and $\mathcal{Y}$ represent the set of all possible states, inputs, and outputs such that $x \in \mathcal{X}$, $u \in \mathcal{U}$, and $y \in \mathcal{Y}$, respectively. In addition, the system is under a safety constraint
\begin{equation*}
    \label{safety}
  E(y_k,u_k) \leq 0,
\end{equation*}
where $E:\mathbb{R}^p\times \mathbb{R}^m \rightarrow \mathbb{R}^l$.

\begin{definition}[Closed-Loop Performance]
\label{def1}
Consider the LTI system \eqref{system} under a tracking control problem with the desired trajectory $r$. Starting from the current state $x_k$, the closed-loop system performance over $N$ steps is characterized by the following cost term:
\begin{equation}
    \label{cost}
    \begin{aligned}
  &J(\bf{y},\bf{u}) = \sum^{k+N-1}_{t=k} \phi(y_t,u_t),
  \end{aligned}
\end{equation}
where $\mathbf{u} = \left[ u_k,\, u_{k+1},\, \cdots,\, u_{k+N-1}  \right]$, $\mathbf{y} = \left[ y_k,\, y_{k+1},\, \cdots,\, y_{k+N-1}  \right]$, and $\phi(y,u)$ is the stage cost.
\end{definition}

\begin{figure}[!ht]
     \centering
     \includegraphics[width=8.8cm, height=7cm]{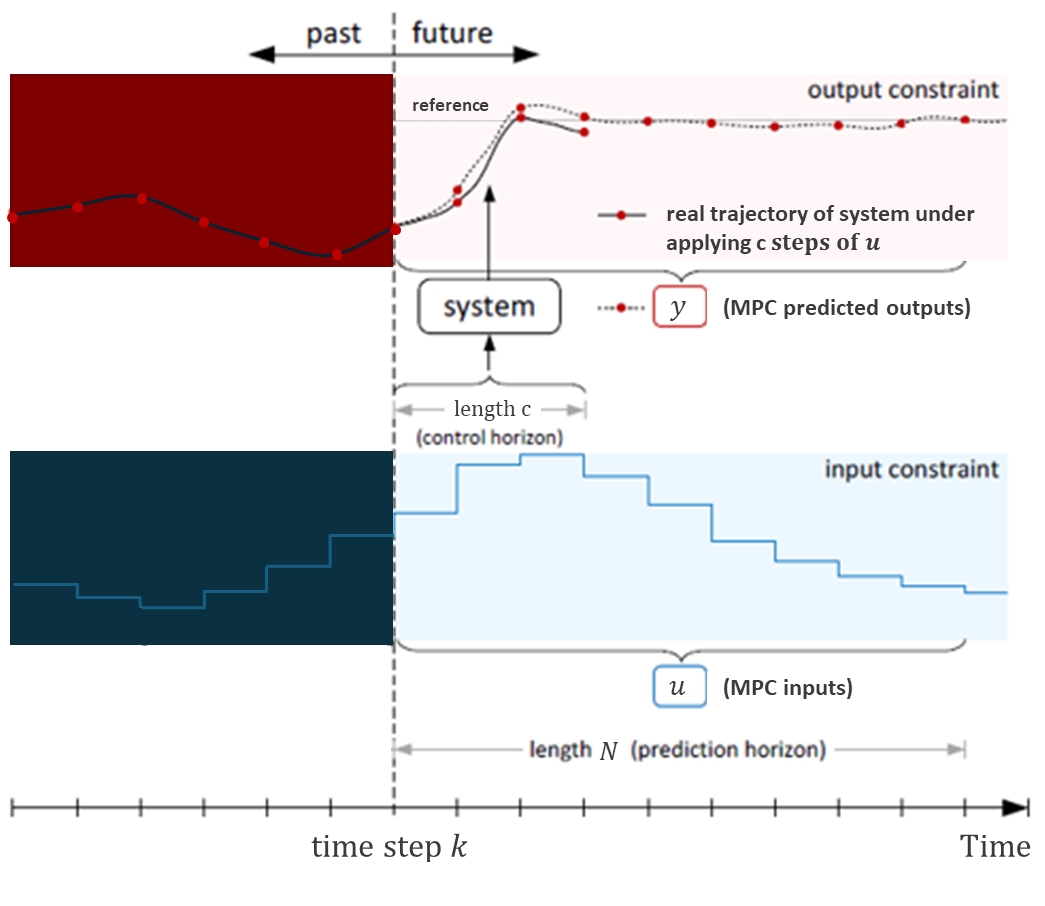}
     \includegraphics[width=0.99\linewidth]{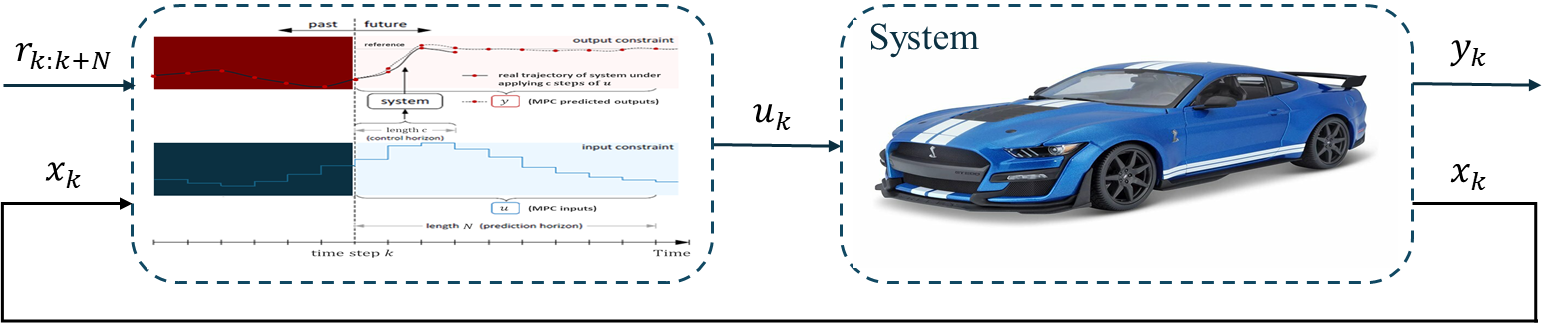}
     \caption{Model Predictive Control.}
     \label{Model Predictive Control}
\end{figure}

As a widely adopted optimal control strategy for the real system \eqref{system}, at each time step $k$, MPC aims at optimizing the system performance \eqref{cost} over a length-$N$ prediction horizon. Under a perfect system model, MPC applies the entire control sequence $\mathbf{u}$ to the system in an open-loop manner. However, to account for model uncertainties and external disturbances, as shown in Fig. \ref{Model Predictive Control}, MPC typically implements only the first $c$ control inputs $u_{k:k+c}$, e.g., $c=0$, in a closed-loop manner, after which the optimization problem is re-solved based on the updated system state. This procedure can be formulated as the following constrained optimization problem:
\begin{equation}
  \begin{aligned}
    \label{NMPC}
    &(\mathbf{y}^{*},\mathbf{u}^{*}) = \underset{\mathbf{y},\mathbf{u}}{\arg\min} \hspace{1 mm} J(\mathbf{y},\mathbf{u})\\
    &s.t. \hspace{5 mm} x_{t+1} = Ax_t+Bu_t\\
    & \hspace{10 mm} y_t = Cx_t+Du_t\\
    & \hspace{10 mm} E(y_t,u_t) \leq 0.
  \end{aligned}
\end{equation}

Substituting $y_t = Cx_t+Du_t$ into the cost function $J(\mathbf{y},\mathbf{u})$ and the constraint $E(y_t,u_t)$, the Hamiltonian function and the augmented cost function are defined as follows:
\begin{equation*}
  \begin{aligned}
    \label{Hamiltonian}
    &H_t = \phi(Cx_t+Du_t,u_t) + \lambda^{T}_{t+1} (Ax_t+Bu_t)\\
    &\hspace{7.5 mm} + \mu^{T}_t E^a(Cx_t+Du_t,u_t),
  \end{aligned}
\end{equation*}

\begin{equation*}
  \begin{aligned}
    \label{aug cost}
    &\bar{J} = \sum^{k+N-1}_{t=k} (H_t - \lambda^{T}_{t+1} x_{t+1}),
    \end{aligned}
\end{equation*}
where $E^{a}(x_t,u_t)$ represents the active constraints at the prediction step $t$. It is worth noting that $E^{a}(x_t,u_t)$ is an empty vector for inactive constraints, and $E^{a}(x_t,u_t) \in \mathbb{R}^{l^a}$ if we have $l^a$ (out of $l$) active constraints. Furthermore, $\mu_t \in \mathbb{R}^{l^a}$ is the Lagrange multiplier associated with the active constraints, and $\lambda_t \in \mathbb{R}^n$ represents the Lagrange multiplier for the system dynamics \eqref{system}. The Lagrange multipliers $\mu_t$ and $\lambda_t$ are also referred to as co-states.

Now, the KKT  conditions \cite{karush2013minima, kuhn2013nonlinear} for the augmented cost function are given as:
\begin{equation*}
  \begin{aligned}
    \label{KKT}
    &{H_{u}}_t = 0, \\
    &\lambda_t = {H_{x}}_t, \\
    &\lambda_N = 0, \\
    &\mu_t \geq 0,
  \end{aligned}
\end{equation*}
where the subscripts $u$ and $x$ stand for the partial derivatives of a function with respect to $u$ and $x$, respectively. By jointly satisfying these conditions, we can determine the optimal solution. In practice, solving the KKT system yields the control sequence $\mathbf{u}$ that minimizes the cost function subject to dynamics and constraints.

We can reformulate MPC \eqref{NMPC} for the nonlinear system \eqref{system-Nonlinear}, which is referred to as nonlinear MPC (NMPC) \cite{mayne1988receding, chen1998nonlinear}:
\begin{equation}
  \begin{aligned}
    \label{NMPC2}
    &(\mathbf{y}^{*},\mathbf{u}^{*}) = \underset{\mathbf{y},\mathbf{u}}{\arg\min} \hspace{1 mm} J(\mathbf{y},\mathbf{u})\\
    &s.t. \hspace{5 mm} x_{t+1} = f(x_t, u_t)\\
    & \hspace{10 mm} y_t = g(x_t,u_t)\\
    & \hspace{10 mm} E(y_t,u_t) \leq 0,
  \end{aligned}
\end{equation}
which is solved similarly to the KKT conditions above.

The cost function \eqref{cost} can take a quadratic form as follows:
\begin{equation}
  \begin{aligned}
    \label{Cost format}
    & J(\mathbf{y},\mathbf{u}) = \|\mathbf{y}-\mathbf{r}\|^2_Q  + \|\mathbf{u}\|^2_R,
  \end{aligned}
\end{equation}
where $\mathbf{r} = \left[ r_k,\, r_{k+1},\, \cdots,\, r_{k+N-1} \right]$, and $Q$ and $R$ are positive-definite matrices of compatible dimensions. Under the above quadratic cost function, MPC \eqref{NMPC} with a linear safety constraint $E(y,u)$ and NMPC \eqref{NMPC2} correspond to a quadratic program (QP) and a nonlinear program (NLP), respectively. Both formulations require iterative solvers
; however, if MPC is unconstrained, i.e., there is no inequality constraint $E(y,u)$, the optimization problem reduces to an unconstrained QP. In this case, a closed-form solution of significantly lower computational burden exists, which yields a linear feedback control law of the form $\mathbf{u} = K_m^r \mathbf{r} + K_m^x x_k$, as shown in \cite{lewis2012optimal}. MPC \eqref{NMPC} generally exhibits a reasonable computational cost, as its complexity scales linearly with both the prediction horizon $N$ and the system dimension (input/state/output variables). In contrast, NMPC \eqref{NMPC2} incurs a significantly higher computational burden, since it requires solving a nonconvex optimization problem at each time step $k$. 

Nevertheless, both MPC and NMPC rely on an accurate parametric system representation to ensure the desired closed-loop performance \eqref{cost}. However, for complex, nonlinear systems, it is often difficult to derive a reliable first-principles model; therefore, ML techniques are increasingly employed to construct a nominal model. In the presence of model uncertainties, these controllers may still perform well on the nominal model; however, the resulting control sequence $\mathbf{u}$, computed under an inaccurate model, can lead to instability or divergence when applied to the real system \eqref{system} or \eqref{system-Nonlinear}. To address these issues, robust MPC \cite{campo1987robust, mayne2005robust, lopez2019dynamic} and adaptive MPC \cite{garcia1986quadratic, morningred1992adaptive, draeger1995model} have been proposed to handle bounded model uncertainties and time-varying system dynamics, respectively.

\textbf{Reinforcement Learning:} RL is a powerful ML paradigm where an agent interacts with an environment to learn optimal decision-making policies \cite{sutton1998reinforcement}. Unlike supervised learning, which relies on labeled datasets, RL agents operate in an (uncertain) environment and receive feedback in the form of scalar rewards for their actions. The agent's objective is to maximize its cumulative long-term reward (return) by iteratively exploring the environment, selecting actions, and observing the resulting consequences. Through trial and error, the agent leverages these reward signals to refine its policy, a mapping function from states to actions, ultimately converging to optimal policy that consistently yield high rewards for the environment. This makes RL particularly well-suited for scenarios where the desired outcome is clear but the specific actions to achieve it are unknown.

Markov Decision Processes (MDPs) provide the mathematical foundation for RL, formalizing RL problems and enabling algorithms to discover optimal policies within this framework \cite{guo2024cooperative}. An MDP is formally defined as a 5-tuple:
\begin{equation}
  \begin{aligned}
    \label{MDPEQ}
    &\mathcal{M} = (\mathcal{S}, \mathcal{A}, P, R, \gamma),
  \end{aligned}
\end{equation}
where
\begin{itemize}
    \item \(\mathcal{S}\): The set of all possible states the agent can be in. 
    
    \item \(\mathcal{A}\): The set of all possible actions the agent can take.  
    Actions may depend on the current state, so sometimes written as \(\mathcal{A}(s)\).
    
    \item \(P\): The state transition probability function, which defines the environment dynamics:
    \[
    P(s' \mid s, a) = \Pr(S_{k+1} = s' \mid S_k = s, A_k = a).
    \]
    This gives the probability of reaching state \(s'\) when taking action \(a\) in state \(s\).
    
    \item \(R\): The reward function, which defines the expected immediate reward:
    \[
    R(s, a, s') = \mathbb{E}[R_{k+1} \mid S_k = s, A_k = a, S_{k+1} = s'].
    \]
    
    \item \(\gamma\): The discount factor, where \(\gamma \in [0,1]\).  
    This determines how much future rewards are valued compared to immediate ones.
    \begin{itemize}
        \item \(\gamma = 0\): The agent only cares about immediate rewards.
        \item \(\gamma = 1\): The agent strongly values long-term rewards.
    \end{itemize}
\end{itemize}

The total accumulated reward, denoted by \(G\), is referred to as finite-horizon discounted return. It is defined as:
\[
G_k = \sum^{k+N-1}_{t=k} \gamma^{t-k} R_{t+1},
\]
where $N = \infty$ for an infinite-horizon discounted return. Table I illustrates the correspondence between RL and MPC notations. In MPC, we typically consider a finite-horizon problem with discount factor $\gamma = 1$, and the RL reward $R_{t+1}$ corresponds to the stage cost $\phi(y_t,u_t)$. Unlike RL, which is usually formulated as a return maximization problem, MPC is formulated as a cost minimization problem.

\begin{table}[h!]
\centering
\caption{Correspondence between MPC and RL notations.}
\begin{tabular}{ccl}
\hline
\hline
\textbf{MPC} & \textbf{RL} & \hspace{7 mm} \textbf{Meaning} \\
\hline
$x$   & $s$                & \hspace{7 mm} State \\ \hline
$\mathcal{X}$   & $\mathcal{S}$                & \hspace{7 mm} State Set \\ \hline
$u$   & $a$                & \hspace{7 mm} Input/Action \\ \hline
$\mathcal{U}$   & $\mathcal{A}$                & \hspace{7 mm} Input/Action Set \\ \hline
\eqref{system-Nonlinear} & $P$     & \hspace{7 mm} System/Environment \\ \hline
$\phi$   & $R$             & \hspace{7 mm} Stage-Cost/Reward \\ \hline
$J$   & $G$                & \hspace{7 mm} Closed-Loop-Performance/Return \\ 
\hline
\end{tabular}
\end{table}

In the context of MDP, policy $\pi$ is formalized as a probability distribution over actions given a state:
\[
\pi(a \mid s) = \Pr(A_k = a \mid S_k = s).
\]

For a deterministic policy, one has \(\pi(s) = a\), a mapping function from the state $s$ to the action $a$. This mapping function specifies that under the policy $\pi$, the action $a$ will be consistently selected when the RL agent occupies the state $s$. The goal of RL is to find a deterministic optimal policy \(\pi^*\) by exploring the environment under a stochastic policy \(\pi\) where actions are chosen according to a probability distribution. Given a policy \(\pi\), state value function (or simply value function) is defined as the expected return when starting from state \(s\) and following policy \(\pi\):
\[
V^{\pi}(s) = \mathbb{E}_{\pi}\left[ G_k \mid S_k = s \right].
\]

The value function $V^{\pi}(s)$ represents the value of all states under the policy \(\pi\). The expectation arises because the policy and/or the environment may be stochastic. To improve the policy \(\pi\) and converge to the deterministic optimal policy \(\pi^*\), state-action value function (or Q-function) is defined as the expected return when starting from state \(s\), taking action \(a\), and thereafter following the policy \(\pi\):
\[
Q^{\pi}(s,a) = \mathbb{E}_{\pi}\left[ G_k \mid S_k = s, A_k = a \right].
\]

These functions provide a way to evaluate how good it is to be in a state (or to take an action in a state) under a given policy \(\pi\). To find the optimal policy, RL algorithms use Bellman expectation equations, where the value function and the Q-function satisfy the following recursive relationships:
\begin{equation*}
  \begin{aligned}
    & V^{\pi}(s) = \mathbb{E}_{\pi}\left[ R_{k+1} + \gamma V^{\pi}(S_{k+1}) \mid S_k = s \right]\\
    & \hspace{10 mm} = \sum_{a \in \mathcal{A}} \pi(a \mid s) 
\Big[ R(s,a) + \gamma \sum_{s' \in \mathcal{S}} P(s,a,s') V^{\pi}(s') \Big],
  \end{aligned}
\end{equation*}
\begin{equation*}
  \begin{aligned}
    & Q^{\pi}(s,a) = \mathbb{E}_{\pi}\left[ R_{k+1} + \gamma Q^{\pi}(S_{k+1}, A_{k+1}) \mid S_k = s, A_k = a \right]\\
    & \hspace{10 mm} = R(s,a) + \gamma \sum_{s' \in \mathcal{S}} P(s,a,s') 
\sum_{a' \in \mathcal{A}} \pi(a' \mid s') Q^{\pi}(s',a').
  \end{aligned}
\end{equation*}

The agent aims to identify an optimal policy
$\pi^*$ that maximizes the value function across all states. This optimal policy must fulfill Bellman optimality equations:
\begin{equation*}
  \begin{aligned}
    & V^{*}(s) = \max_{a \in \mathcal{A}} \; \mathbb{E}\left[ R_{k+1} + \gamma V^{*}(S_{k+1}) \mid S_k = s, A_k = a \right]\\
    & \hspace{10 mm} = \max_{a \in \mathcal{A}} 
\Big[ R(s,a) + \gamma \sum_{s' \in \mathcal{S}} P(s,a,s') V^{*}(s') \Big],
  \end{aligned}
\end{equation*}
\begin{equation*}
  \begin{aligned}
    & Q^{*}(s,a) = \mathbb{E}\left[ R_{k+1} + \gamma \max_{a'} Q^{*}(S_{k+1}, a') \mid S_k = s, A_k = a \right]\\
    & \hspace{13.5 mm} = R(s,a) + \gamma \sum_{s' \in \mathcal{S}} P(s,a,s') 
\max_{a' \in \mathcal{A}} Q^{*}(s',a').
  \end{aligned}
\end{equation*}

Here, \(V^{*}(s)\) and \(Q^{*}(s,a)\) are the optimal value function and Q-function, where the maximization reflects choosing the best action to maximize expected return. Therefore, the optimal policy $\pi^{*}$ is obtained as:
\[
\pi^{*} \in \arg\max_{\pi} V^{\pi}(s), \quad \forall s \in \mathcal{S},
\]
\[
\pi^{*}(s) \in \arg\max_{a \in \mathcal{A}} Q^{*}(s,a), \quad \forall s \in \mathcal{S},
\]
where we use the symbol "$\in$" because the operator $\arg\max$ may return a set of optimal solutions rather than a unique one. In such cases, the optimal policy $\pi^*$ is one element of this set. On the other hand, the symbol "$=$" can be used if we make the additional assumption that there exists a unique policy that maximizes the value function. However, since uniqueness is not guaranteed in general, the more precise and widely applicable notation is to use "$\in$".

Note that under deterministic policies and environments, the Bellman expectation and optimality equations simplify as:
\[
V^{\pi}(s) = R(s, \pi(s)) + \gamma \, V^{\pi}\!\big( P(s, \pi(s)) \big),
\]
\[
Q^{\pi}(s,a) = R(s,a) + \gamma \, V^{\pi}\!\big( P(s,a) \big),
\]
\[
V^{*}(s) = \max_{a \in \mathcal{A}} \Big[ R(s,a) + \gamma \, V^{*}\!\big(P(s,a) \big) \Big],
\]
\[
Q^{*}(s,a) = R(s,a) + \gamma \, \max_{a' \in \mathcal{A}} Q^{*}\!\big(P(s,a), a' \big).
\]

\begin{figure}[!ht]
     \centering
     \includegraphics[width=0.99\linewidth]{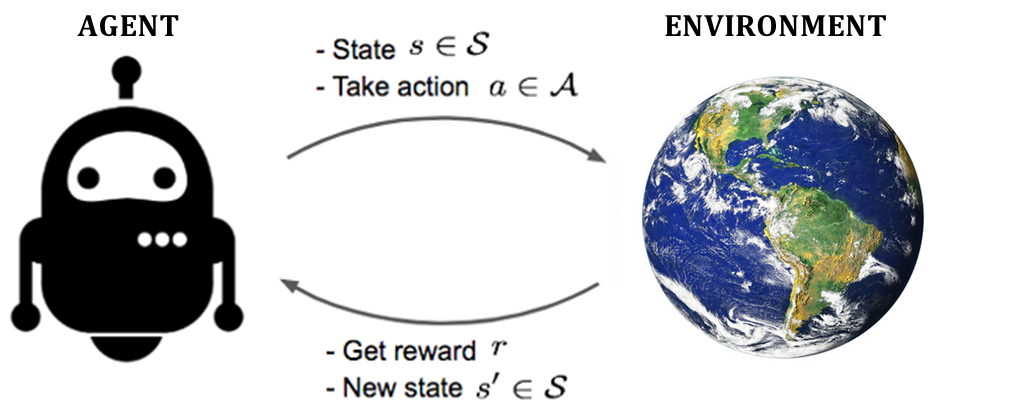}\\
     \vspace{0.05 \linewidth}
     \includegraphics[width=8.8cm, height=3.8cm]{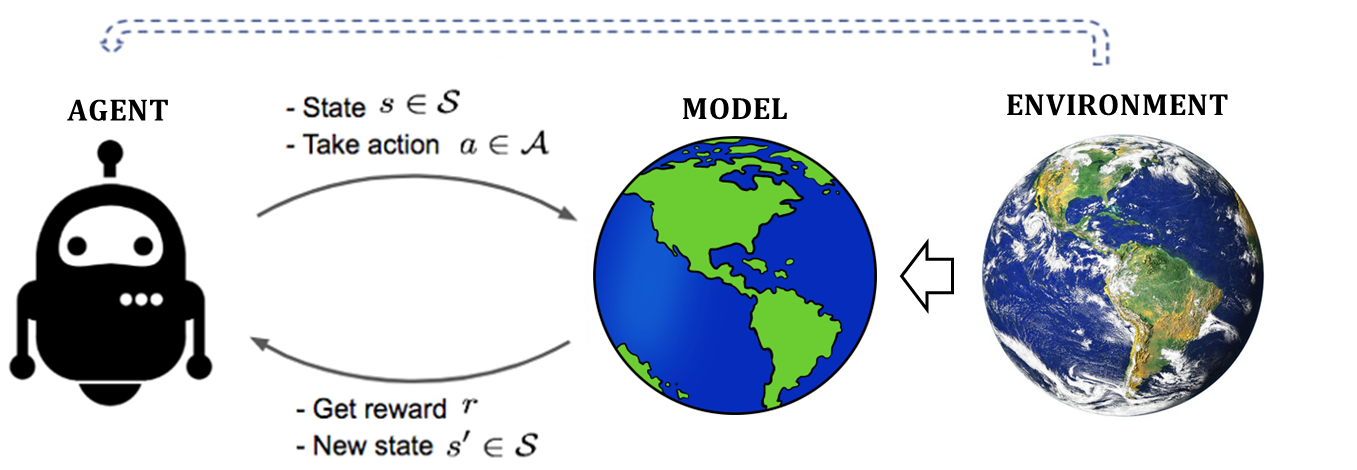}
     \caption{Reinforcement Learning: Model-free vs Model-based Approaches.}
     \label{Model-based vs Model-free}
\end{figure}

Based on the availability of the environment model, RL methods are broadly categorized into model-based and model-free approaches \cite{swazinna2022comparing}, as shown in Fig. \ref{Model-based vs Model-free}. Model-based methods, such as Dynamic Programming (DP), require complete knowledge of the transition and reward functions. In contrast, model-free methods learn directly from interaction with the environment without assuming access to such models, making them more suitable for complex or unknown domains \cite{swazinna2022comparing}. When the transition and reward functions are fully known, the Bellman equations can be directly leveraged through DP \cite{wei2015value}. DP methods, i.e., Value Iteration and Policy Iteration, compute the optimal value function and policy by repeatedly applying the Bellman equations. Value Iteration iteratively updates value estimates using the Bellman optimality operator, while Policy Iteration alternates between evaluating a policy and improving it based on the current value function \cite{wang2023recent}. These methods are exact and efficient when applied to small, fully known environments. However, they are impractical in real-world settings with large or unknown state spaces, motivating the development of model-free approaches that learn directly from experience.

One of the earliest model-free approaches is Monte Carlo (MC) methods \cite{wei2022monte}. MC methods estimate value functions by averaging returns obtained from complete episodes of experience. They operate without a model and rely on sampled trajectories to update value estimates. Two common variants are first-visit and every-visit Monte Carlo, which differ in how they handle repeated state visits within an episode. While MC methods are simple and effective, they require episodes to terminate before updates can be made (no bootstrapping). This limits their applicability in continuing (non-episodic) tasks or environments with very long episodes. Moreover, their reliance on full returns (cumulative rewards until episode end) leads to high variance in estimates, which can slow convergence. Despite these limitations, MC methods laid the foundation for more efficient bootstrapped methods and remain useful for environments where precise return estimates are critical (e.g., policy evaluation in finite episodes) \cite{shakya2023reinforcement}. To overcome the limitations of MC methods, Temporal Difference (TD) learning was introduced as a model-free approach that combines ideas from both MC and DP \cite{shakya2023reinforcement}. Unlike MC, TD methods update value estimates incrementally at each time step, without waiting for the end of an episode. This allows learning to proceed in ongoing tasks and improves data efficiency. The core idea is to bootstrap: the current estimate is updated using the observed reward and the estimate of the next state \cite{wang2023solving}. TD learning methods are divided into on-policy and off-policy approaches \cite{sutton2018reinforcement}. On-policy methods, such as SARSA \cite{yao2025improved}, learn the value of the policy currently being executed, incorporating the actions actually taken during training. Off-policy methods, such as Q-learning \cite{shi2022pessimistic}, learn the value of an optimal policy independently of the agent's behavior, by updating toward the greedy action even if a different action was executed. The off-policy methods generally offer better convergence properties and enable learning from arbitrary experiences, including those generated by other policies or replay buffers \cite{wang2024reliable}.

Although tabular TD methods like SARSA and Q-learning are effective for small, discrete environments, they become impractical when the state or action space is large or continuous. To address this, Deep Reinforcement Learning (Deep RL) integrates RL algorithms with DNNs to approximate value functions or policies \cite{tang2025deep}. One of the most prominent Deep RL algorithms is Deep Q-Network (DQN), which extends Q-learning by using a DNN to approximate the Q-function \cite{vivek2024integrating}. DQN introduced key innovations such as experience replay and target networks to stabilize training, enabling successful learning from high-dimensional inputs like raw images. However, DQN is limited to discrete action spaces and struggles with continuous control tasks, while many real-world problems, such as robotics and autonomous vehicles, require optimality over continuous actions. To address this, several Deep RL algorithms have been developed specifically for continuous control tasks. An alternative class of Deep RL methods is based on policy gradient techniques, which directly optimize a parameterized policy by maximizing the expected return. REINFORCE algorithm is a foundational example, using sampled trajectories to estimate gradients \cite{williams1992simple}. While simple, it suffers from high variance. To address this, Actor-Critic methods combine a learned policy (actor) with a value function (critic) to provide more stable learning \cite{zhou2023natural}. These methods are suitable for high-dimensional and continuous action spaces, setting the stage for advanced Deep RL algorithms designed for real-world robotics and autonomous vehicles. Deep Deterministic Policy Gradient (DDPG) algorithm extends the actor-critic framework to deterministic policies and continuous action spaces by learning both a policy network (actor) and a Q-function (critic) \cite{sumiea2024deep}. However, DDPG is known to be sensitive to hyperparameters and prone to overestimation bias. To improve stability, Twin Delayed DDPG (TD3) introduces three key enhancements: clipped double Q-learning to reduce overestimation, target policy smoothing for better generalization, and delayed policy updates to stabilize learning \cite{xu2024td3}. On the other hand, Proximal Policy Optimization (PPO) is a widely used on-policy method that strikes a balance between ease of implementation and training stability \cite{zamfirache2024adaptive}. It improves upon vanilla policy gradient methods by using a clipped surrogate objective that limits policy updates within a trust region, reducing destructive updates and improving robustness. Another powerful approach is Soft Actor-Critic (SAC), an off-policy actor-critic method that maximizes both expected return and policy entropy \cite{wang2024adaptive}. This encourages exploration by promoting stochastic policies and has shown strong performance across a range of continuous control benchmarks. Together, these algorithms form the backbone of modern continuous Deep RL.

In some settings, directly interacting with the environment to learn optimal policy may be unsafe, expensive, or inefficient. Instead, learning from expert demonstrations becomes a practical alternative. Imitation Learning aims to mimic expert behavior without access to explicit reward signals \cite{lin2024behavioral}. A simple approach is Behavior Cloning, where a supervised learning model is trained to map observed states to expert actions. Although it is effective in structured tasks, it can suffer from compounding errors due to distributional shift when the agent visits unfamiliar states \cite{lin2024behavioral}. To address this, Inverse Reinforcement Learning (IRL) focuses on recovering the underlying reward function that best explains expert behavior \cite{8103164}. Once the reward is inferred, standard RL algorithms can be applied to learn a policy that optimizes it. IRL provides a more principled foundation than direct imitation, especially in scenarios where understanding the expert's intent or generalizing to new situations is important. Although computationally more intensive, IRL methods are particularly useful in domains such as robotics and autonomous vehicles, where reward design is challenging. 

\begin{figure}[!ht]
     \centering
     \includegraphics[width=0.99\linewidth]{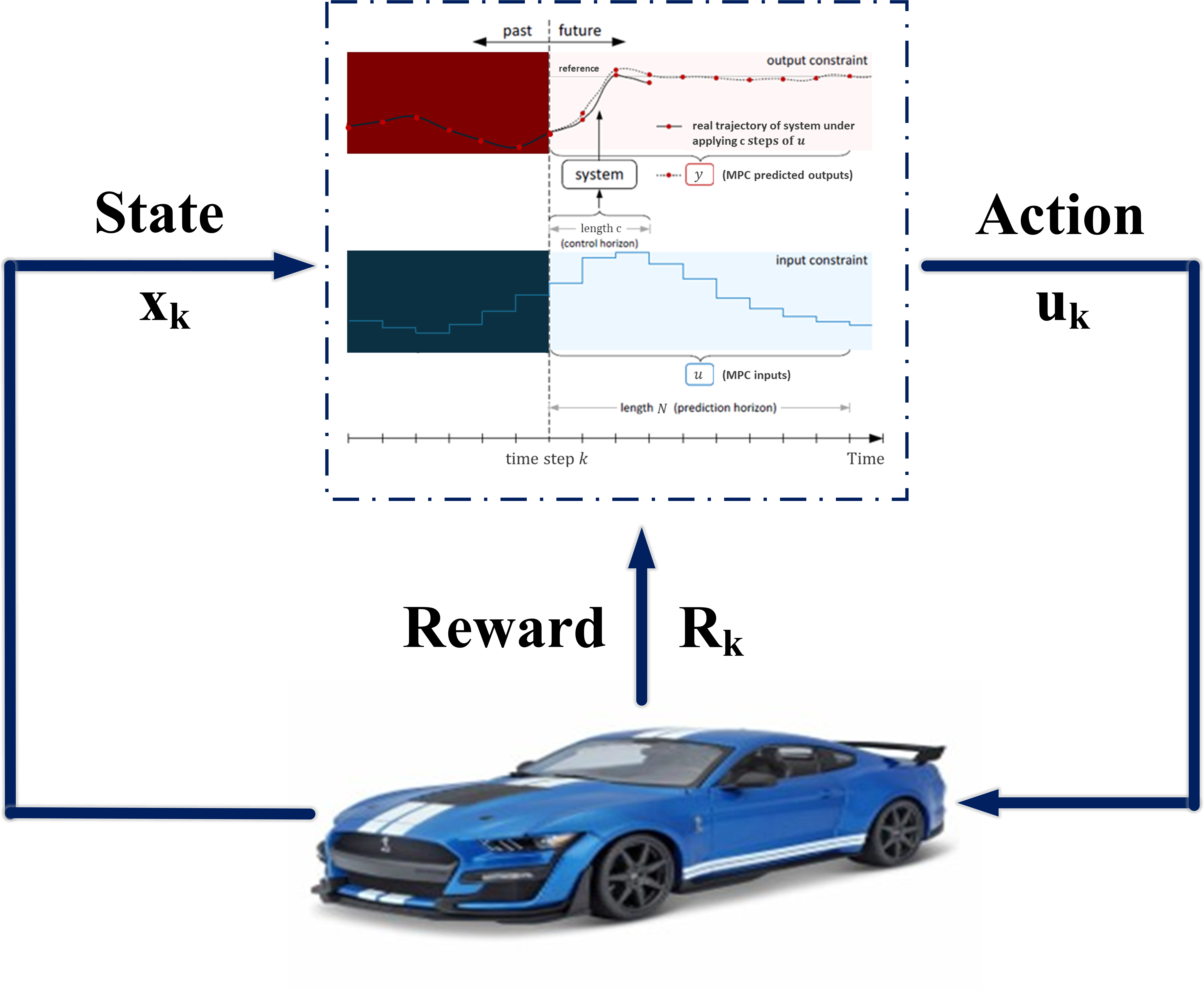}
     \caption{Model Predictive Control-based Reinforcement Learning.}
     \label{MPC-based RL}
\end{figure}

\textbf{MPC-based Reinforcement Learning:}
Both MPC and RL present optimal policies; however, MPC requires a parametric model for its optimization problem, and RL does not guarantee system safety during its policy learning \cite{kokolakis2022safe, kamalapurkar2018reinforcement}. MPC-based RL methods combine the strengths of both paradigms, integrating the safety handling of MPC with the adaptability of RL, which allows policies to adjust more flexibly while still benefiting from optimization-based safeguards \cite{wabersich2018safe}. Thus, MPC-based RL methods provide a more deployment-ready solution that offers a sample-efficient and safer alternative for applications where direct trial-and-error learning may be infeasible. In MPC-based RL, we use MPC, instead of DNN, as a function approximator for the value functions to guarantee system safety during optimal policy learning. Therefore, one can define the optimization problem \eqref{NMPC2} as a parametric state value function $V_\theta(x_k)$, where the cost weights, the model parameters, and the constraint parameters are represented as unknown parameters $\theta$ and are trained using model-free RL algorithms (e.g., TD or policy gradient) \cite{zanon2020safe, gros2019data, zanon2019practical, gros2020safe}:
\begin{equation}
  \begin{aligned}
    \label{NMPC_V}
    &V^{\pi_{mpc}}_\theta(x_k) = \underset{\mathbf{y},\mathbf{u}}{\min} \hspace{1 mm} J(\mathbf{y},\mathbf{u})\\
    &s.t. \hspace{5 mm} x_{t+1} = f(x_t, u_t)\\
    & \hspace{10 mm} y_t = g(x_t,u_t)\\
    & \hspace{10 mm} E(y_t,u_t) \leq 0,
  \end{aligned}
\end{equation}
which represents expected return when starting from state $x_k$ and following MPC policy $\pi_{mpc}$, as illustrated in Fig. \ref{MPC-based RL}.

Consequently, the parametric Q-function $Q^{\pi_{mpc}}_\theta(x_k,u_k)$ is:
\begin{equation}
  \begin{aligned}
    \label{NMPC_Q}
    &Q^{\pi_{mpc}}_\theta(x_k,u_k) = \underset{\mathbf{y},\mathbf{u}}{\min} \hspace{1 mm} J(\mathbf{y},\mathbf{u})\\
    &s.t. \hspace{5 mm} x_{t+1} = f(x_t, u_t)\\
    & \hspace{10 mm} y_t = g(x_t,u_t)\\
    & \hspace{10 mm} E(y_t,u_t) \leq 0,
  \end{aligned}
\end{equation}
which represents expected return when starting from state $x_k$, taking action $u_k$, and thereafter following MPC policy $\pi_{mpc}$. In Q-function, $\mathbf{u}$ denotes the control sequence without $u_k$.

With an MPC parametrization $\theta$ capable of representing the true optimal policy, one can formulate optimal policy learning as the problem of identifying the optimal parameters $\theta^*$:
\begin{equation*}
  \begin{aligned}
    \label{optimal_values}
    &Q^*(x_k,u_k) = Q^{\pi_{mpc}}_{\theta^*}(x_k,u_k), \\
    &V^*(x_k) = V^{\pi_{mpc}}_{\theta^*}(x_k) = \underset{u_k}{\min} \hspace{1 mm} Q^{\pi_{mpc}}_{\theta^*}(x_k,u_k), \\
    & \pi^*(x_k) = \pi_{\theta^*}(x_k) \in \underset{u_k}{\arg\min} \hspace{1 mm} Q^{\pi_{mpc}}_{\theta^*}(x_k,u_k),
  \end{aligned}
\end{equation*}
where $\pi_{mpc} = \pi_\theta$. We adopt minimization instead of maximization since MPC is formulated as a cost minimization problem. Moreover, we use the symbol "$\in$" for the optimal policy because minimization problem may return a set of minimizers rather than a unique solution. This situation naturally arises in NMPC, where the underlying optimization problem is nonconvex and can admit multiple locally optimal control actions for the same state $x_k$; therefore, the optimal policy is chosen from this set of minimizers. Nevertheless, since numerical solvers return a specific solution that is treated as the control input to implement, the convention in control community is to denote it with "$=$", even if the solution is only locally optimal. In contrast, for MPC problems where the cost and constraints are convex, the optimizer is guaranteed to be unique, and the equality sign "$=$" is used without ambiguity.

MPC-based Q-Learning: From the TD perspective, MPC-based RL for safe Q-learning is formulated as follows:
\begin{equation*}
  \begin{aligned}
    \label{MPC-RL}
    &\text{Initialize parameters} \hspace{1 mm} \theta, \\
    & \tau_k = \bar{J}^c_{\theta_{k}} + \gamma^{c+1} {V^{\pi_{mpc}}_{\theta_{k}}}(x_{k+c+1}) - {Q^{\pi_{mpc}}_{\theta_{k}}}(x_{k},u_{k}), \\
    & \theta_{k+1} = \theta_{k} + \alpha \tau_k {\nabla_\theta} {Q^{\pi_{mpc}}_{\theta_{k}}}(x_{k},u_{k}),
  \end{aligned}
\end{equation*}
where $\tau$ denotes TD error, ${\bar{J}^{c}_{\theta_{k}}}$ is the augmented cost function associated to the MPC problem \eqref{NMPC_Q} evaluated at its first $c$ optimal solution (typically $c=0$, with $0 \leq c \leq N-1$) under the parameters ${\theta_{k}}$, $\gamma$ is the discount factor, $\alpha$ is the learning rate used in the gradient descent updates, and 
\begin{equation*}
  \begin{aligned}
    \label{gardient_Q}
    &{\nabla_\theta} {Q^{\pi_{mpc}}_{\theta_{k}}}(x_k,u_k) = {\nabla_\theta} \bar{J}_{\theta_{k}}.
  \end{aligned}
\end{equation*}
where $u_k$ is selected according to the MPC policy $\pi_{mpc}$ with the possible addition of occasional random exploratory moves.

The above MPC-based RL framework naturally extends to continuous control tasks, since the MPC policy $\pi_{mpc}$ is obtained by solving a continuous optimization problem and directly produces real-valued control inputs. In this setting, exploration is not carried out by selecting arbitrary random actions, which could easily violate safety constraints, but rather by perturbing the MPC solution through structured noise or parameter variations, ensuring that exploratory actions remain feasible. 
This stands in contrast to DQN, where exploration is typically implemented using an $\epsilon$-greedy strategy, i.e., selecting a random action with probability $\epsilon$ and the greedy action otherwise. While this is effective for discrete action spaces, DQN cannot be applied directly to continuous control tasks because computing $\max_{a} Q_{\theta}(s,a)$ is intractable for DNN function approximators without discretization. MPC-based RL avoids this limitation by embedding the maximization within the MPC optimization problem, thereby combining continuous control, safety guarantees, and the ability to incorporate exploration in a controlled manner.

It should be noted that the objective of Q-learning is to approximate the optimal Q-function $Q^\star$ by fitting the parametrized function $Q^{\pi_{mpc}}_\theta$, with the expectation that $Q^{\pi_{mpc}}_\theta \approx Q^\star$ will also imply $\pi_\theta \approx \pi^\star$. In the ideal case where $Q^\star$ is recovered exactly, the corresponding MPC policy is indeed optimal. In practice, however, a close approximation of $Q^\star$ by $Q^{\pi_{mpc}}_\theta$ does not necessarily ensure that the induced policy $\pi_\theta$ is itself close to the optimal policy $\pi^\star$. Thus, it is often advantageous to employ the policy gradient methods, which aim at directly improving the closed-loop performance of the policy rather than relying solely on the accuracy of the value function approximation. In what follows, we focus on DDPG framework, which can be expressed as:
\begin{equation*}
\nabla_\theta \bar{J}^{\pi_{mpc}}_\theta = \mathbb{E}\!\left[\nabla_\theta \pi_\theta(x_k)\,\nabla_{u_k} Q^{\pi_{mpc}}_\theta(x_k,u_k)\right],
\label{eq:policy_gradient}
\end{equation*}
where $\bar{J}^{\pi_{mpc}}_\theta$ denotes the expected, augmented closed-loop performance under the MPC policy $\pi_\theta$, and $Q^{\pi_\theta}_\theta$ is the corresponding Q-function. The expectation $\mathbb{E}$ is taken over trajectories of the real system evolving under the MPC policy $\pi_\theta$. A necessary condition for the optimality of $\pi_\theta$ is then:
\begin{equation*}
\nabla_\theta \bar{J}^{\pi_{mpc}}_\theta = 0.
\label{eq:optimality_condition}
\end{equation*}

MPC-based DDPG: In practice, DDPG algorithms are often implemented within a TD Actor–Critic framework as follows:
\begin{equation*}
\begin{aligned}
&\text{Initialize parameters} \ \theta \ \text{and} \ \vartheta, \\
& \tau_k = \bar{J}^{\pi_{mpc}} + \gamma^{c+1} Q_{\vartheta_k}(x_{k+c+1},u_{k+c+1}) - Q_{\vartheta_k}(x_k,u_k), \\
& \vartheta_{k+1} = \vartheta_k + \alpha_\vartheta \tau_k \nabla_\vartheta Q_{\vartheta_k}(x_k,u_k), \\
& \theta_{k+1} = \theta_{k} + \alpha_\theta \nabla_\theta \pi_{\theta_{k}}(x_k)\,\nabla_{u_k} Q_{\vartheta_k}(x_k,u_k),
\end{aligned}
\label{eq:actor_critic}
\end{equation*}
where $u_{k+1}$ is obtained from the MPC policy $\pi_{\theta_k}(x_{k+1})$, $\alpha_\vartheta$ and $\alpha_\theta$ denote the critic and actor learning rates, respectively, and $Q_\vartheta \approx Q^{\pi_\theta}$ serves as an approximator of the true Q-function. Moreover, the gradient of the MPC policy with respect to $\theta$ can be expressed as:
\begin{equation*}
\nabla_\theta \pi_{\theta_{k}}(x_k) = - \nabla_\theta \xi^{\pi_{mpc}}_{\theta_{k}}(x_k,o^\star)\,
\left(\nabla_o \xi^{\pi_{mpc}}_{\theta_{k}}(x_k,o^\star)\right)^{-1}\,
\frac{\partial o}{\partial u_k},
\label{eq:policy_gradient_NMPC}
\end{equation*}
where $\xi^{\pi_{mpc}}_{\theta_{k}}(x_k,o^\star)$ collects the KKT conditions associated with the MPC problem \eqref{NMPC_V} under its primal–dual optimal solution $o^\star = (x^\star, u^\star, \lambda^\star, \mu^\star)$.

MPC with the optimal parameters $\theta^*$ thus constitutes the optimal policy $\pi^*_{mpc}$, and the role of Q-learning or DDPG is to guide the update of $\theta$ toward this optimum. At each learning iteration, the MPC problem must be solved to both determine the action $u_k$ applied to the system and to evaluate the gradients, which serve as the learning signals. While this integration ensures that safety constraints are respected throughout the learning process, it also introduces significant computational demands, since repeatedly solving the MPC problem can be costly in practice. Moreover, to guarantee system safety during training under non-optimal parameters $\theta$, model uncertainty, and disturbances, a robust MPC formulation is required, ensuring that constraint satisfaction is maintained even in the presence of uncertainties. Finally, by continually adapting the parameters $\theta$ through learning, this framework naturally extends to an adaptive MPC scheme, which is particularly advantageous for controlling time-varying systems.

\begin{figure*}[!t]
     \centering
     \includegraphics[width=0.99\linewidth]{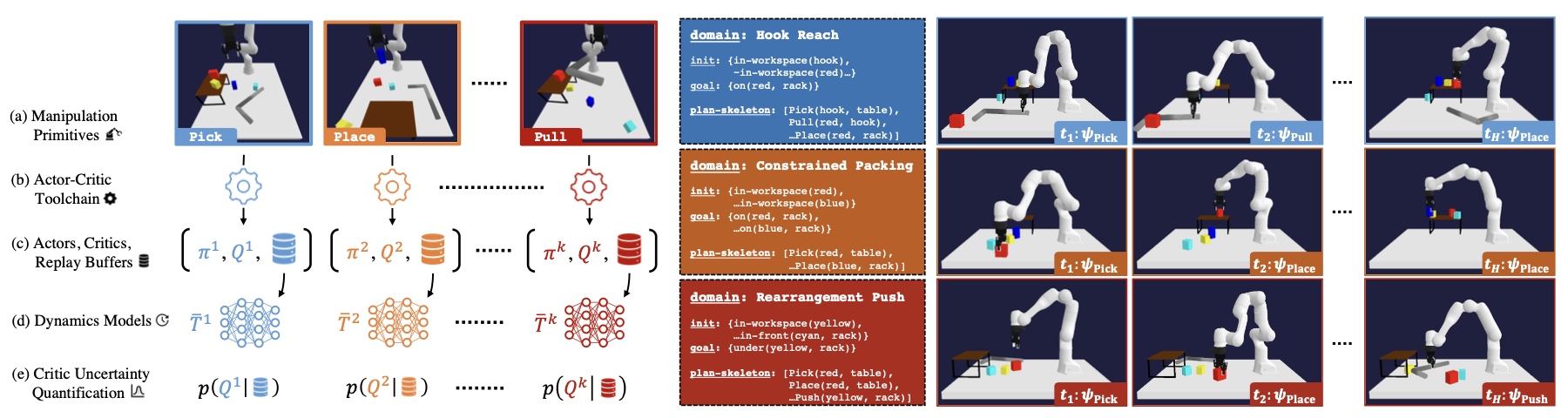}\\
     \caption{STAP: Sequencing Task-Agnostic Policies \cite{agia2022stap}.}
     \label{STAP}
\end{figure*}

\textbf{Large Language Model Agents:} Large Language Models (LLMs) have recently emerged as a promising tool for robot/vehicle motion control, particularly in the context of translating high-level natural language instructions into structured task plans \cite{lin2023text2motion, cui2024personalized}. While RL and MPC provide strong guarantees in terms of feasibility and optimality for motion planning and control in robotics and autonomous vehicles \cite{jebellat2024motion, jebellat2025designing, jebellat2024anti}, they are limited in their ability to process flexible, open-ended instructions from humans. LLMs, in contrast, can reason over natural language and produce structured symbolic sequences, offering a direct interface between human instructions and robotic control. However, a key challenge in using LLMs for planning lies in the mismatch between symbolic plans and the embodied execution of those plans. Symbolic reasoning produced by LLMs tends to be agnostic to geometric feasibility, temporal constraints, or the specific action affordances of a robot in a given environment. This problem becomes especially acute in long-horizon tasks, where even a single infeasible step may cascade into failure of the entire sequence. Thus, LLM-based motion planning must be tightly integrated with feasibility verification mechanisms that can prune unrealistic plans before execution. To address this, LLM planners often make use of a skill library, where each skill abstracts a learned policy or primitive action that can be reused across tasks \cite{felip2013manipulation}. A skill is modeled as a contextual bandit MDP \eqref{MDPEQ} such that each skill $\psi$ is associated with a policy $\pi(a|s)$, a Q-function $Q^\pi(s,a)$, and a primitive parameterization $\phi(a)$. Execution of a skill is treated as binary, yielding success or failure, which provides a modular and compositional building block for constructing longer plans. This formulation allows symbolic task specifications from the LLM to be grounded in reusable, trained RL policies.  

Given a natural language instruction $i$ and an initial state $s$, the planning problem is to find a sequence of skills $\psi_{1:H}$ that achieves the instruction with high probability. This probability decomposes as:  
\begin{equation*}
p(\psi_{1:H}, R_{1:H} | i, s) = p(\psi_{1:H} | i, s) \; p(R_{1:H} | i, s, \psi_{1:H}), 
\end{equation*}
where the reward $R$ denotes a binary success indicator for each skill, taking the value $1$ if the skill completes successfully and $0$ otherwise. The decomposition highlights two components: the symbolic plan generation from the LLM, $p(\psi_{1:H} | i, s)$, and the execution feasibility, $p(R_{1:H} | i, s, \psi_{1:H})$. Since success is ultimately determined by the sequence of state-action rollouts, one has:  
\begin{equation*}
p(R_{1:H} | i, s, \psi_{1:H}) = p(R_{1:H} | s, A_{1:H}), 
\end{equation*}
where the action $A$ is the parameterization of the skill primitive. This formulation reveals why LLM planning alone is insufficient: LLM may propose symbolically coherent plans, but without grounding in action parameters, there is no guarantee of feasibility. Thus, feasibility is evaluated through a geometric planner that searches over $A_{1:H}$. A key approach is sequencing task-agnostic policies (STAP) \cite{agia2022stap}, as shown in Fig. \ref{STAP}, which selects the actions $A_{1:H}$ to maximize expected success:  
\begin{equation*}
A^{*}_{1:H} = \arg\max_{A_{1:H}} \; \mathbb{E} \Bigg[\prod_{k=1}^H p(R_k | S_k, A_k)\Bigg]. 
\end{equation*}

This expectation over successor states captures the fact that feasibility is not independent across steps: an action feasible in isolation may still render later steps impossible. The optimization is approximated using Q-functions associated with each skill, producing a tractable surrogate for feasibility evaluation:  
\begin{equation*}
p(R_{1:H} | s, A_{1:H}) \approx \prod_{k=1}^H Q^{\pi_k}(S_k, A^*_k), 
\label{eq:5}
\end{equation*}
where by chaining Q-values, the planner effectively evaluates the joint probability of multi-step success, pruning sequences that may be symbolically correct but geometrically infeasible.  

Beyond feasibility, another critical issue in LLM planning is termination. In many robotic tasks, the instruction does not specify a fixed horizon but rather an outcome condition (e.g., ``place the red block on the blue block’’). Instead of relying on handcrafted stop-skills or arbitrary time horizons, LLMs can be prompted to predict symbolic goal propositions $G = \{g_1, \dots, g_j\}$ representing the intended outcomes of the instruction. A plan is deemed successful if there exists a state along the trajectory that satisfies one of the predicted goals:  
\begin{equation*}
\exists s \in S_{2:H+1} : F^{G}_{\text{sat}}(s) = 1, 
\label{eq:6}
\end{equation*}
which enables a more natural alignment between language instructions and robotic execution, as it captures termination in terms of goal satisfaction rather than mechanical completion.

\begin{figure}[!ht]
     \centering
     \includegraphics[width=0.99\linewidth]{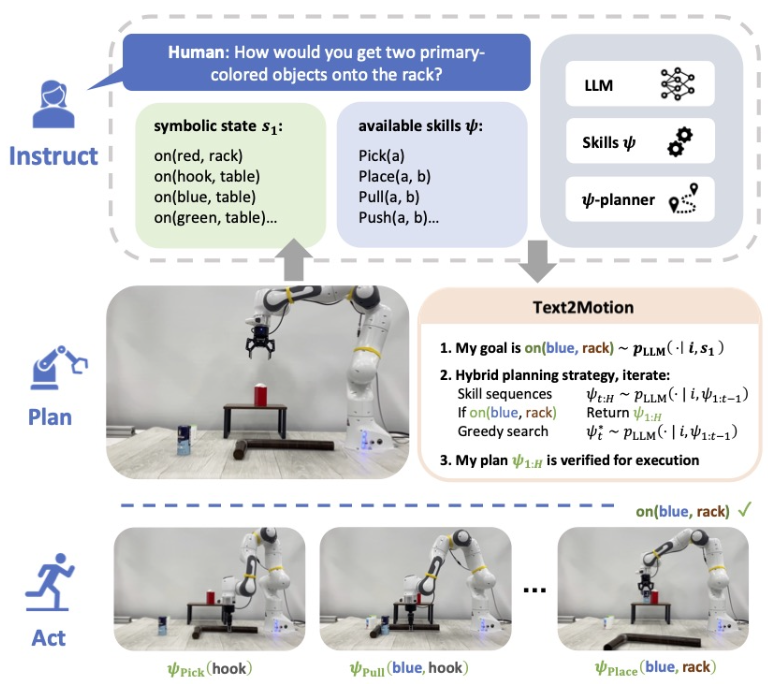}\\
     \caption{Text2Motion: From Natural Language Instructions to Feasible Plans \cite{lin2023text2motion}.}
     \label{Text2Motion}
\end{figure}

Text2motion framework \cite{lin2023text2motion}, as shown in Fig. \ref{Text2Motion}, has proposed two broad strategies for coupling LLM-based planning with feasibility verification. In shooting-based planning, LLM generates full candidate sequences, which are subsequently validated by the feasibility planner. This method is efficient when the LLM has a strong prior over feasible skill sequences, but its success rate diminishes in domains with large search spaces or subtle geometric dependencies. In contrast, search-based planning constructs plans incrementally: at each step, LLM proposes candidate skills, and the feasibility planner evaluates them before extending the plan. This greedy approach is more robust in scenarios with partial observability, lifted goal conditions, or environments with complex affordances, although it can incur higher computational costs due to repeated feasibility evaluations. In practice, hybrid strategies combine these two approaches, leveraging the efficiency of shooting when feasible sequences can be generated directly, and falling back to search when the sampled candidates fail. Out-of-distribution detection mechanisms can also be employed to reject skill proposals outside the scope of the library, further improving robustness. The result is a planner that balances the flexibility of LLM reasoning with the grounded reliability of skill-based feasibility evaluation. This integration of LLM planning with skill libraries and geometric verification illustrates a broader theme in robot decision-making: combining the symbolic reasoning capabilities of LLMs with the execution guarantees of RL-based skills and feasibility-aware planning. From the perspective of MPC, the feasibility evaluation serves a role analogous to trajectory optimization, systematically pruning infeasible action sequences before they are executed. In this sense, LLM planning does not replace MPC or RL but instead complements them, offering a powerful mechanism to interpret natural language and propose symbolic plans, while simultaneously grounding those plans in the execution capabilities of the robot. Natural language provides a flexible and intuitive interface for specifying tasks, whereas feasibility verification and goal prediction ensure that generated plans remain both reliable and executable. The result is a unified system that bridges high-level task specification with low-level motion execution, complementing MPC and RL for robotic motion planing and control.   

\begin{figure}[!ht]
     \centering
     \includegraphics[width=0.99\linewidth]{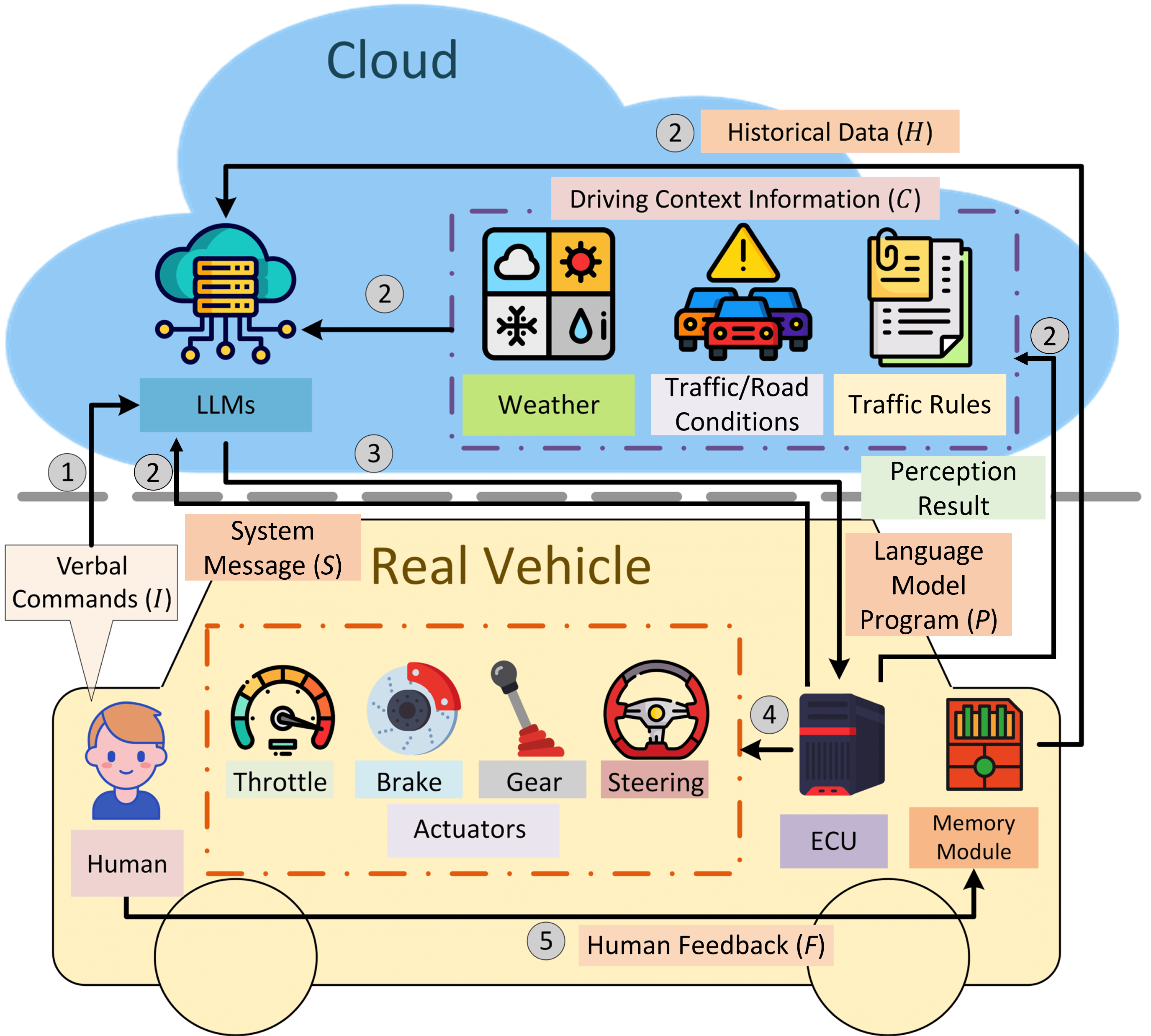}\\
     \caption{Talk2Drive: Personalized Autonomous Driving with Large Language Models \cite{cui2024personalized}.}
     \label{Talk2Drive}
\end{figure}

Talk2Drive framework \cite{cui2024personalized}, as shown in Fig. \ref{Talk2Drive}, demonstrates how LLMs can translate verbal commands into executable control programs for real vehicles. The pipeline begins with speech recognition that converts driver requests into textual form, which are then augmented with contextual information such as weather, road geometry, traffic conditions, and traffic rules. Instead of operating directly on numerical state vectors, the system constructs textual descriptions of the driving context, thereby enabling LLM to reason within its natural domain. Through in-context learning and chain-of-thought prompting, LLM interprets the driver’s instructions together with system-level constraints, producing so-called Language Model Programs (LMPs). These programs adjust high-level control parameters such as target velocity, look-ahead distance, and trajectory following configuration, which are transmitted to the electronic control unit for execution. Two layers of verification ensure feasibility: one checking the syntactic validity of generated commands, and another evaluating their safety (e.g., rejecting policies that exceed speed limits). To enable personalization, Talk2Drive incorporates a memory module that logs every past interaction, including commands, generated LMPs, and human feedback. This historical archive allows the system to refine its responses over time and adapt to individual driving styles, effectively learning personal preferences for comfort, safety, and aggressiveness. In contrast to simulation-only studies, these experiments represent one of the first demonstrations of LLM-driven personalization in real-world autonomous driving, emphasizing how language-based planning and reasoning can serve as an adaptive bridge between human intent and executable robotic control.

\begin{figure}[!ht]
     \centering
     \includegraphics[width=8.8cm, height=7cm]{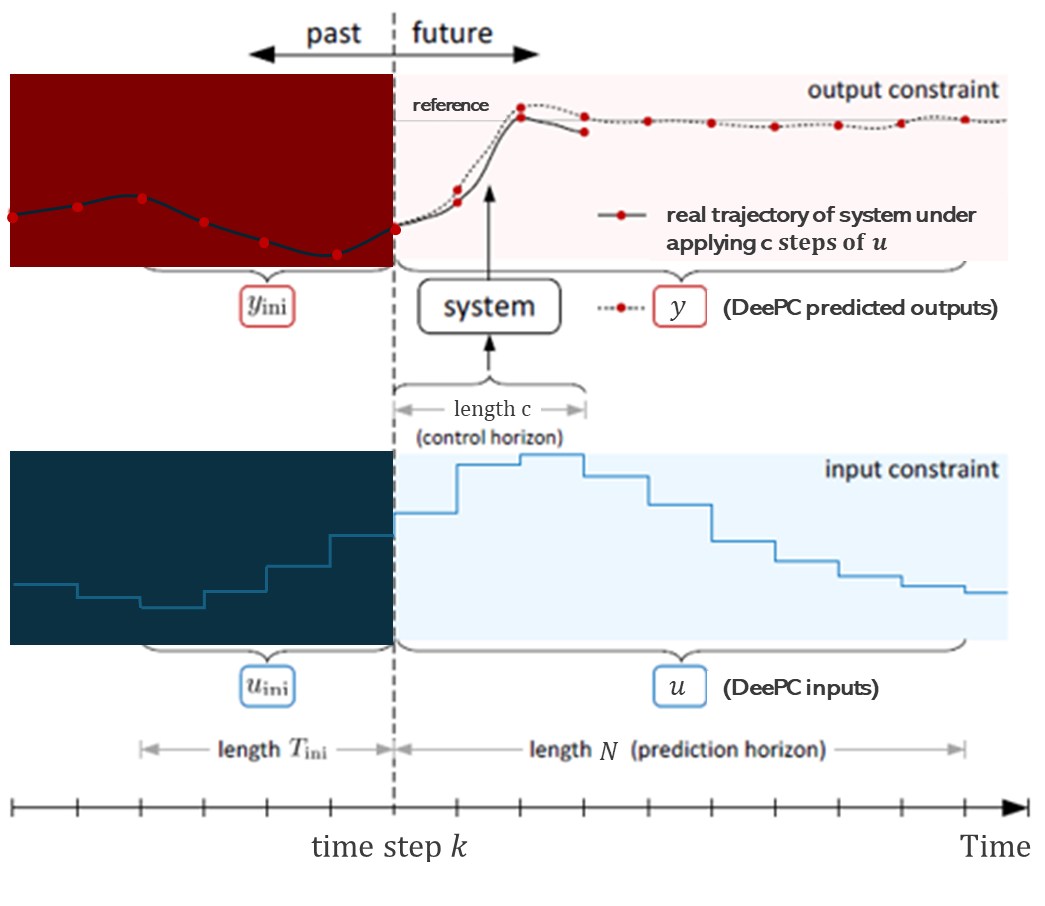}
     \includegraphics[width=0.99\linewidth]{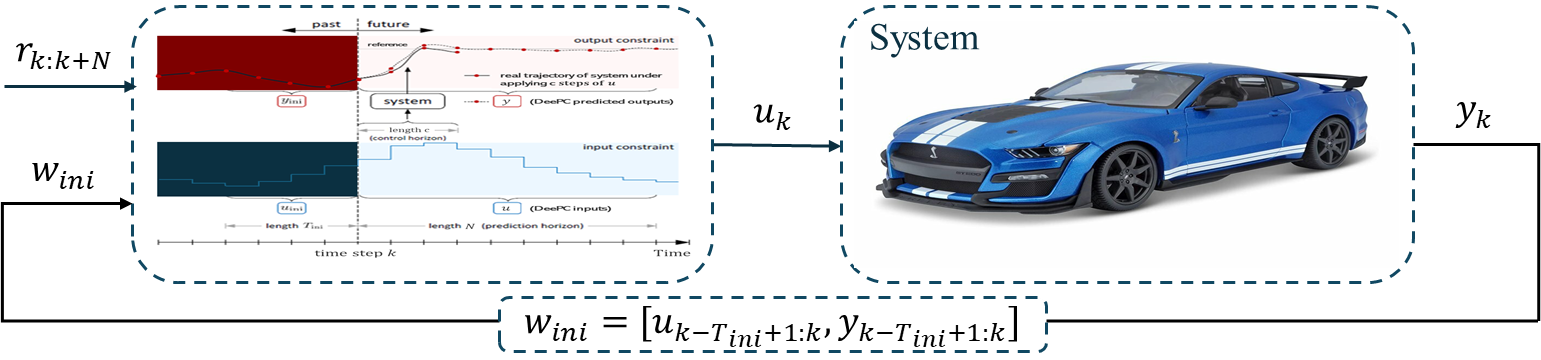}
     \caption{Data-Enabled Predictive Control.}
     \label{Data-Enabled Predictive Control}
\end{figure}

\textbf{Data-Enabled Predictive Control:}
The design of data-enabled predictive control (DeePC) is purely data driven: the collected data are used directly in the controller through the behavioral systems framework. Unlike MPC and RL, which require a state estimator when the full system state is unavailable, DeePC naturally operates on system outputs and therefore does not need an explicit state estimator. In this framework, the state-space model \eqref{system} is replaced by the algebraic constraint \eqref{data-driven sc}, which links the collected data with the past $T_{ini}$-long I/O trajectory and the future $N$-long I/O trajectory. When $T_{ini} \geq \ell$, this construction ensures a unique current state estimation, and hence a unique prediction. As shown in Fig. \ref{Data-Enabled Predictive Control}, given an initial trajectory $w_{ini}=(u_{ini}, y_{ini})$, the optimization problem \eqref{NMPC} can then be reformulated in the DeePC framework \cite{coulson2019data}:
\begin{equation}
  \begin{aligned}
    \label{DeePC}
    &(\mathbf{g}^{*},\mathbf{y}^{*},\mathbf{u}^{*}) = \underset{\mathbf{g},\mathbf{y},\mathbf{u}}{\arg\min} \hspace{1 mm} J(\mathbf{y},\mathbf{u})\\
    &s.t. \hspace{5 mm} \begin{bmatrix}
      U_P\\ Y_P\\ U_F\\ Y_F 
     \end{bmatrix} g = 
     \begin{bmatrix}
      u_{ini}\\ y_{ini}\\ u\\ y 
     \end{bmatrix}\\
    & \hspace{10 mm} E(y,u) \leq 0,
  \end{aligned}
\end{equation}
where $(u_{ini}, y_{ini})$ is the $T_{ini}$-long initial trajectory, and $(u,y)$ is the $N$-long future trajectory. Moreover, the fixed data matrices $U_P$, $Y_P$, $U_F$, and $Y_F$ are obtained offline (from collected data), the equality constraint represents the Fundamental Lemma, and the inequality constraint is the safety constraints.

Using $y = Y_F g$ and $u = U_F g$ as free optimization variables, one can rewrite \eqref{DeePC} as:
\begin{equation}
  \begin{aligned}
    \label{DeePC g}
    &\mathbf{g}^{*} = \underset{\mathbf{g}}{\arg\min} \hspace{1 mm} J(Y_F\mathbf{g},U_F\mathbf{g})\\
    &s.t. \hspace{5 mm} \begin{bmatrix}
  U_P\\ Y_P
  \end{bmatrix} g = 
  \begin{bmatrix}
  u_{ini}\\ y_{ini}
  \end{bmatrix}\\
    & \hspace{10 mm} E(Y_F g,U_F g) \leq 0,
  \end{aligned}
\end{equation}
where we only have $g$ as the optimization variable. For the above optimization problem, the augmented cost function is:
\begin{equation*}
  \begin{aligned}
    \label{Aug Cost DeePC}
    & \bar{J} = J(Y_F g,U_F g) + \lambda^{T} \left(\begin{bmatrix}
  Y_P\\ U_P 
  \end{bmatrix} g - 
  \begin{bmatrix}
  u_{ini}\\ y_{ini} 
  \end{bmatrix} \right) \\
  & \hspace{5.5 mm} + \mu^\top E^a(Y_F g,U_F g),
  \end{aligned}
\end{equation*}
and the KKT conditions are given as:
\begin{equation*}
  \begin{aligned}
    \label{KKT DeePC}
    &\bar{J}_{g} = 0,\\
    &\mu \geq 0.
  \end{aligned}
\end{equation*}

To streamline the discussion of the different approaches, the following notations will be used:
\begin{equation*}
  \begin{aligned}
    &w_{ini} = \begin{bmatrix}
  u_{ini}\\ y_{ini} 
  \end{bmatrix}, \mathcal{H}_P = \begin{bmatrix}
      U_P\\ Y_P 
     \end{bmatrix}, \mathcal{H}_{PU} = \begin{bmatrix}
      U_P\\ Y_P\\ U_F 
     \end{bmatrix}, \mathcal{H} = \begin{bmatrix}
      U_P\\ Y_P\\ U_F\\ Y_F 
     \end{bmatrix},
  \end{aligned}
\end{equation*}
where $w_{ini} \in \mathbb{R}^{(m+p)T_{ini}}$, $\mathcal{H}_P\in \mathbb{R}^{(m+p)T_{ini}\times L}$, $\mathcal{H}_{PU}\in \mathbb{R}^{(m+p)T_{ini}+mN\times L}$, $\mathcal{H}\in \mathbb{R}^{(m+p)K\times L}$, $K = T_{ini}+N$, $L = T-z(K-1)$, and $z$ represents the number of collected trajectories in the Hankel matrix.

If $w_{ini} \notin \operatorname{im}(\mathcal{H}_P)$, where is the case for stochastic and/or nonlinear system \eqref{system-Nonlinear}, DeePC is infeasible. To address stochastic and/or nonlinear systems, DeePC is reformulated as:
\begin{equation}
  \begin{aligned}
    \label{DeePC-Nonlinear}
    &(\mathbf{g}^{*},\mathbf{y}^{*},\mathbf{u}^{*}, \mathbf{\sigma_u}^{*}, \mathbf{\sigma_y}^{*}) = \underset{\mathbf{g},\mathbf{y},\mathbf{u},\mathbf{\sigma_u},\mathbf{\sigma_y}}{\arg\min} \hspace{1 mm} J(\mathbf{y},\mathbf{u}, \mathbf{\sigma_u},\mathbf{\sigma_y}, \mathbf{g})\\
    &s.t. \hspace{5 mm} \begin{bmatrix}
      U_P\\ Y_P\\ U_F\\ Y_F 
     \end{bmatrix} g = 
     \begin{bmatrix}
      u_{ini}\\ y_{ini}\\ u\\ y 
     \end{bmatrix} +
     \begin{bmatrix}
      \sigma_u\\ \sigma_y\\ 0\\ 0 
     \end{bmatrix}\\
    & \hspace{10 mm} E(y,u) \leq 0,
  \end{aligned}
\end{equation}
where $\sigma_u \in \mathbb{R}^{mT_{ini}}$ and $\sigma_y \in \mathbb{R}^{pT_{ini}}$ are auxiliary slack variables to ensure the feasibility of the optimization problem due to noises and/or nonlinearities. These slack variables are the residuals we are allowed to use to explain noise (variance) and/or error (bias) in the past data. 
In practice, we often keep only $\sigma_y$ and set $\sigma_u = 0$ since past inputs are commanded and recorded exactly. Under the standard PE condition, the Past Input Hankel matrix $U_p$ has full row rank $mT_{ini}$, therefore, any initial input segment $u_{ini}$ lies in $col(U_p)$. Hence, allowing $\sigma_u \neq 0$ would let the optimizer rewrite the applied inputs to fit the data, which is undesirable. Only in rare cases—logging errors, time misalignment, or unmodeled input saturation—$\sigma_u$ might be used. Moreover, $J(\mathbf{y},\mathbf{u}, \mathbf{\sigma_u},\mathbf{\sigma_y}, \mathbf{g})$ represents a modified cost function including two penalty terms for slack variables and also a regularization term on optimization variable $g$ to avoid overfitting issues, i.e., to use the most informative features (columns) of the Hankel matrix. The modified cost function can take a quadratic form as follows:
\begin{equation}
  \begin{aligned}
    \label{Cost format-modified}
    & J(\mathbf{y},\mathbf{u}, \mathbf{\sigma_u},\mathbf{\sigma_y}, \mathbf{g}) = \|\mathbf{y} - \mathbf{r}\|^2_Q  + \|\mathbf{u}\|^2_R \\
    & \hspace{29 mm} + \lambda_u \|\mathbf{\sigma_u}\|^2_2 + \lambda_y \|\mathbf{\sigma_y}\|^2_2 + \lambda_g \|\mathbf{g}\|^2_2,
  \end{aligned}
\end{equation}
where $\lambda_u$, $\lambda_y$, and $\lambda_g$ are regularization parameters. Note that we consider $\ell_2$ regularization to keep the optimization problem quadratic; however, $\ell_1$ regularization is another option, which is more robust to outliers. We refer the interested readers to \cite{shang2024convex} for further details on different regularization terms to handle stochastic and/or nonlinear systems.

Using $\sigma_u = U_P g - u_{ini}$ and $\sigma_y = Y_P g - y_{ini}$, one can rewrite \eqref{DeePC-Nonlinear} as follows:
\begin{equation}
  \begin{aligned}
    \label{DeePC g-nonlinear}
    &\mathbf{g}^{*} = \underset{\mathbf{g}}{\arg\min} \hspace{1 mm} J(Y_F\mathbf{g},U_F\mathbf{g},U_P \mathbf{g} - u_{ini}, Y_P \mathbf{g} - y_{ini}, \mathbf{g})\\
    &s.t. \hspace{5 mm} E(Y_F g,U_F g) \leq 0,
  \end{aligned}
\end{equation}
where we only have $g$ as the optimization variable. For the above optimization problem, the augmented cost function is defined as:
\begin{equation}
  \begin{aligned}
    \label{Aug Cost DeePC-nonlinear}
    & \bar{J} = J(Y_F g,U_F g, U_P g - u_{ini}, Y_P g - y_{ini}, g) \\
  & \hspace{5.3 mm} + \mu^\top E^a(Y_F g,U_F g).
  \end{aligned}
\end{equation}

\begin{figure*}[!b]
\begin{equation}
  \begin{aligned}
    \label{SPC}
    &(\mathbf{y}^{*},\mathbf{u}^{*},\mathbf{\sigma_u}^{*},\mathbf{\sigma_y}^{*}) = \underset{\mathbf{y},\mathbf{u},\mathbf{\sigma_u},\mathbf{\sigma_y}}{\arg\min} \hspace{1 mm} J \left( \mathbf{y},\mathbf{u},\mathbf{\sigma_u},\mathbf{\sigma_y}, H_{ini} \begin{bmatrix}
  u_{ini} + \mathbf{\sigma_u}\\ y_{ini} + \mathbf{\sigma_y}
  \end{bmatrix} + H_{u} \mathbf{u} \right)\\
    &s.t. \hspace{5 mm} y = S_{ini} \begin{bmatrix}
  u_{ini} + \sigma_u\\ y_{ini} + \sigma_y
  \end{bmatrix} 
  + S_u u\\
    & \hspace{10 mm} E(y,u) \leq 0.
  \end{aligned}
\end{equation}
\end{figure*}

\begin{figure*}[!b]
\begin{equation}
  \begin{aligned}
    \label{SPC 2}
    &(\mathbf{u}^{*},\mathbf{\sigma_u}^{*},\mathbf{\sigma_y}^{*}) = \underset{\mathbf{u},\mathbf{\sigma_u},\mathbf{\sigma_y}}{\arg\min} \hspace{1 mm} J \left( S_{ini} \begin{bmatrix}
  u_{ini}+\mathbf{\sigma_u}\\ y_{ini}+\mathbf{\sigma_y} 
  \end{bmatrix} 
  + S_u \mathbf{u},\mathbf{u},\mathbf{\sigma_u},\mathbf{\sigma_y}, H_{ini} \begin{bmatrix}
  u_{ini} + \mathbf{\sigma_u}\\ y_{ini} + \mathbf{\sigma_y}
  \end{bmatrix} + H_{u} \mathbf{u} \right) \\
    & s.t. \hspace{5 mm} E\left(S_{ini} \begin{bmatrix}
  u_{ini} + \sigma_u\\ y_{ini}+\sigma_y 
  \end{bmatrix} 
  + S_u u,u\right) \leq 0.
  \end{aligned}
\end{equation}
\end{figure*}

DeePC \eqref{DeePC-Nonlinear} can handle mildly nonlinear and/or time-varying behavior. However, for strongly nonlinear or time-varying systems, one needs online DeePC \cite{baros2022online, teutsch2023online, shi2024efficient, vahidi2024online}, which updates the Hankel matrix using real-time data. In both formulations, \eqref{DeePC} and \eqref{DeePC-Nonlinear}, DeePC reduces to a quadratic program (QP) under a quadratic cost function, and thus requires an online QP solver to compute the optimal control input via the decision variable $g$. If the constraint $E(y,u)$ is absent, the problem becomes unconstrained DeePC, which admits a closed-form solution with significantly reduced computational burden. In this case, the control input is given by $\mathbf{u} = U_F \mathbf{g} = K_d^r \mathbf{r} + K_d^{ini} w_{ini}$. Such a closed-form solution essentially exists only for convex quadratic problems with equality constraints. For more general optimization problems, a closed form is typically unavailable, and DeePC must be solved numerically. This can lead to high computational cost because the dimension of the decision variable $g$ scales with the data length $T$ in the Hankel matrix. In addition, DeePC requires storing the Hankel matrix $\mathcal{H}$, which can become very large for long datasets, leading to substantial memory demands. This poses a fundamental limitation: while more data typically improves representation of the nonlinear system \eqref{system-Nonlinear}, it also makes DeePC increasingly expensive to implement. To address this trade-off, several efficient variants have been proposed to improve the scalability and practicality of DeePC.

\section{Efficient Data-Driven Optimal Policy}
We review and compare eight approaches to improve the efficiency of the data-driven optimal policies while preserving their performance guarantees. Although the primary focus is on DeePC, most of the methods are also applicable to MPC, ML-based MPC, RL, MPC-based RL, and LLM agents.

\vspace{2 mm}
\textbf{Subspace Predictive Control:}
In subspace predictive control (SPC), the state-space model \eqref{system-Nonlinear} is replaced by an algebraic relation that relates the future output trajectory $y$ with the initial trajectory $w_{ini}$ and the future input signal $u$. Within the DeePC framework, SPC retains the Fundamental Lemma as a hard constraint in \eqref{DeePC-Nonlinear}, isolates $y = Y_F g$ from it, and constructs an ARX predictor for $y$ as follows \cite{favoreel1999spc, huang2008dynamic, huang2019data}:
\begin{equation*}
  \begin{aligned}
    \label{ARX Predictor g}
    g=\begin{bmatrix}
  U_P\\ Y_P\\ U_F 
  \end{bmatrix}^\dagger 
  \begin{bmatrix}
  u_{ini} + \sigma_u\\ y_{ini}+ \sigma_y\\ u 
  \end{bmatrix} = H_{ini} \begin{bmatrix}
  u_{ini} + \sigma_u\\ y_{ini} + \sigma_y
  \end{bmatrix} + H_{u} u,
  \end{aligned}
\end{equation*}
\begin{equation*}
  \begin{aligned}
    \label{ARX Predictor}
    &y = Y_F \begin{bmatrix}
  U_P\\ Y_P\\ U_F 
  \end{bmatrix}^\dagger 
  \begin{bmatrix}
  u_{ini}+ \sigma_u\\ y_{ini}+ \sigma_y\\ u 
  \end{bmatrix}= S_{ini} \begin{bmatrix}
  u_{ini} + \sigma_u\\ y_{ini} + \sigma_y
  \end{bmatrix} 
  + S_u u,
  \end{aligned}
\end{equation*}
where $^\dagger$ represents the Moore-Penrose pseudoinverse operator, $H_{ini} \in \mathbb{R}^{L \times (m+p)T_{ini}}$, $H_{u} \in \mathbb{R}^{L \times mN}$, $S_{ini} \in \mathbb{R}^{pN \times (m+p)T_{ini}}$, and $S_{u} \in \mathbb{R}^{pN \times mN}$. Now, using the above equations, one can replace DeePC \eqref{DeePC-Nonlinear} with SPC \eqref{SPC}, where the fixed data matrices $H_{ini}$, $H_u$, $S_{ini}$, and $S_u$ are identified, from the collected data, offline. Then, substituting the ARX predictor into the cost function and the safety constraint, one can rewrite \eqref{SPC} as \eqref{SPC 2}. For the cost function \eqref{Cost format-modified}, the problem is a QP in $(u,\sigma_u,\sigma_y)$. Eliminating $\sigma_u$ and $\sigma_y$ in closed form yields a QP in $u$ only \cite{boyd2004convex}. Solving the tracking problem then requires an online QP solver for the decision variable $u$. Therefore, the computational burden depends on the dimension of $u$, that is $mN$, instead of the dimension of $g$, that is $L$. In addition, we only need to save $S_{ini}$ and $S_u$ (we can skip the last term of the cost function in SPC), which requires much less memory compared to saving the Hankel matrix $\mathcal{H}$ since they do not depend on the length of the collected data $T$. Moreover, if the constraint $E(y,u)$ is neglected, the problem is referred to as unconstrained SPC, and one can achieve a closed-form solution by setting the gradient of the quadratic cost to zero $(J_u = 0)$ as $\mathbf{u} = K_s^r \mathbf{r} + K_s^{ini} w_{ini}$.

\begin{figure*}[!b]
\begin{equation}
  \begin{aligned}
    \label{NPC}
    &(\mathbf{z}^{*},\mathbf{\sigma_u}^{*},\mathbf{\sigma_y}^{*}) = \underset{\mathbf{z},\mathbf{\sigma_u},\mathbf{\sigma_y}}{\arg\min} \hspace{1 mm} J \left( Y_F \left( \mathcal{H}_P^\dagger \begin{bmatrix}
  u_{ini}+\mathbf{\sigma_u}\\ y_{ini}+\mathbf{\sigma_y} 
  \end{bmatrix} 
  + \Phi \mathbf{z} \right),U_F \left( \mathcal{H}_P^\dagger \begin{bmatrix}
  u_{ini}+\mathbf{\sigma_u}\\ y_{ini}+\mathbf{\sigma_y} 
  \end{bmatrix} 
  + \Phi \mathbf{z} \right),\mathbf{\sigma_u},\mathbf{\sigma_y},\mathcal{H}_P^\dagger \begin{bmatrix}
  u_{ini}+\mathbf{\sigma_u}\\ y_{ini}+\mathbf{\sigma_y} 
  \end{bmatrix} 
  + \Phi \mathbf{z} \right) \\
    & s.t. \hspace{5 mm} E\left( Y_F \left( \mathcal{H}_P^\dagger \begin{bmatrix}
  u_{ini}+\mathbf{\sigma_u}\\ y_{ini}+\mathbf{\sigma_y} 
  \end{bmatrix} 
  + \Phi \mathbf{z} \right),U_F \left( \mathcal{H}_P^\dagger \begin{bmatrix}
  u_{ini}+\mathbf{\sigma_u}\\ y_{ini}+\mathbf{\sigma_y} 
  \end{bmatrix} 
  + \Phi \mathbf{z} \right) \right) \leq 0.
  \end{aligned}
\end{equation}
\end{figure*}

Although SPC reduces the computational cost of DeePC, it does not qualify as a fully direct data-driven control method, since the predictor matrices $S_{ini}$ and $S_u$ are identified offline via least-squares estimation. Furthermore, because SPC relies on a linear predictor, it is not well-suited for addressing the nonlinear system \eqref{system-Nonlinear}. For the deterministic LTI system \eqref{system}, ML-based MPC, DeePC, and SPC coincide and yield the same $y$. Beyond this case, they differ since ML-based MPC and SPC allow any $u$ and predict $y$ using the parametric model and the ARX predictor, respectively; however, DeePC restricts $u$ to $u = U_F g$ and predicts $y$ as $y = Y_F g$. This distinction reflects a bias–variance trade-off. ML-based MPC and SPC reduce variance by relying on a fixed linear predictor, which enables them to handle measurement noise in the linear system \eqref{system} effectively. Yet, when the true system is nonlinear \eqref{system-Nonlinear}, this fixed predictor incurs model bias, leading to systematic prediction error. The restriction lowers variance, as it limits the degrees of freedom and prevents overfitting to noise, but increases bias if the feasible set is too small. By contrast, DeePC does not commit to a single linear parametric model, which reduces model bias when the system departs from linearity. This flexibility allows DeePC to perform better on nonlinear systems compared to ML-based MPC with linear models or SPC. The downside, however, is greater sensitivity to noise, which is mitigated by regularization and slack penalties.

\vspace{2 mm}
\textbf{Nullspace Predictive Control:}
This framework was introduced in \cite{carlet2022data}, although the authors did not assign it a specific name. We refer to it as nullspace predictive control (NPC), since it exploits the null-space representation of the Past submatrix in the Fundamental Lemma. Within the DeePC formulation \eqref{DeePC-Nonlinear}, the decision variable $g$ can be decomposed into a lower-dimensional variable, allowing the future trajectory $(u, y)$ to be computed with reduced online computational cost. In analogy to \eqref{DeePC g}, NPC enforces the Past component of the Fundamental Lemma as a hard constraint in \eqref{DeePC-Nonlinear} and represents the null-space parametrization of $g$ as follows:
\begin{equation*}
  \begin{aligned}
    \label{UDeePC g}
    & g = \begin{bmatrix}
  U_P\\ Y_P
  \end{bmatrix}^\dagger
  \begin{bmatrix}
  u_{ini} + \sigma_u\\ y_{ini} + \sigma_y
  \end{bmatrix} + \Phi z = \mathcal{H}_P^\dagger \begin{bmatrix}
  u_{ini} + \sigma_u\\ y_{ini} + \sigma_y
  \end{bmatrix} + \Phi z,
  \end{aligned}
\end{equation*}
where $\mathcal{H}_P^\dagger \in \mathbb{R}^{L \times (m+p)T_{ini}}$, $\Phi \in \mathbb{R}^{L \times (L-r_p)}$, $z \in \mathbb{R}^{L-r_p}$, and $r_p = rank(\mathcal{H}_P)$. $\Phi$ denotes a basis of the kernel (null space) of $\mathcal{H}_P$, i.e., $\mathcal{H}_P \Phi = 0$. Therefore, one has
\begin{equation*}
  \begin{aligned}
 & \begin{bmatrix}
  U_P\\ Y_P
  \end{bmatrix} g = 
  \begin{bmatrix}
  U_P\\ Y_P
  \end{bmatrix} \begin{bmatrix}
  U_P\\ Y_P
  \end{bmatrix}^\dagger
  \begin{bmatrix}
  u_{ini} + \sigma_u\\ y_{ini}+ \sigma_y
  \end{bmatrix} + \begin{bmatrix}
  U_P\\ Y_P
  \end{bmatrix} \Phi z \\
  & \hspace{11 mm} = \begin{bmatrix}
  u_{ini}+ \sigma_u\\ y_{ini}+ \sigma_y
  \end{bmatrix} + 0.
  \end{aligned}
\end{equation*}
    
Both $\mathcal{H}_P^\dagger$ and $\Phi$ can be computed offline using standard linear algebra routines. This decomposition allows one to express the future trajectory $(u, y)$ as a function of the lower-dimensional variable $z$, which is useful in the online phase of DeePC. Therefore, one can rewrite DeePC \eqref{DeePC-Nonlinear} as NPC \eqref{NPC}. If $r_p = L$, we have $\Phi = 0$, and $z$ disappears; thus, NPC collapses to DeePC. For the cost function \eqref{Cost format-modified}, the problem is a QP in $(z,\sigma_u,\sigma_y)$. Similarly to SPC, one can solve out $\sigma_u$ and $\sigma_y$ in closed form and get a QP in $z$ only, which requires an online QP solver to solve the tracking problem for the optimization variable $z$. Therefore, the computational burden depends on the dimension of $z$, that is $L-r_p$, instead of the dimension of $g$, that is $L$. However, NPC can easily be bigger/slower than SPC since the absolute reduction $r_p$ is a small fraction of $L$. When $L$ is large, and we need an efficient DeePC, NPC cuts only a small slice off $L$. Moreover, considering their memory burden, storing and using $\mathcal{H}_P^\dagger$ and $\Phi$, which depend on the length of the collected data $T$, is typically much larger than $S_{ini}$ and $S_u$. It should be noted that if the constraint $E(y,u)$ is neglected, the problem is referred to as unconstrained NPC, and one can achieve a closed-form solution by setting the gradient of the quadratic cost to zero $(J_z = 0)$. The resulting online controller is a linear feedback of the form $\mathbf{u} = K_n^r \mathbf{r} + K_n^{ini} w_{ini}$, which is obtained from $z$. On the other hand, the constrained NPC problem requires an online QP solver, which has a high computational cost since the dimension of the decision variable $z$ can be large as it depends on the number of samples $T$ used in the Hankel matrix. It is worth noting that in general convex optimization problems \cite{boyd2004convex} and nonconvex optimization problems \cite{nocedal2006numerical}, although $\sigma_u$ and $\sigma_y$ can be eliminated, retaining them explicitly as decision variables is often preferable for numerical stability and theoretical guarantees.

NPC is a change of variables that reduces the dimensionality of the DeePC problem without altering its statistical properties. Since the feasible set remains unchanged, NPC preserves DeePC’s bias–variance behavior. In particular, computing $\mathcal{H}_P^\dagger$ and $\Phi$ offline merely precomputes linear algebra operations that would otherwise be performed online, without modifying the underlying optimization problem. Unlike SPC, which defines a different feasible set through regression-based identification of predictor matrices, NPC enforces the same Hankel feasibility constraints as DeePC. Consequently, DeePC and NPC yield identical feasible pairs $(u,y)$ and therefore the same optimizer $(u^*,y^*)$; only the representation of the optimization variables differs.
Finally, while DeePC and NPC are equivalent in this sense, ML-based MPC with nonlinear models (e.g., DNNs, CNNs, RNNs) can surpass both in nonlinear settings. Nonetheless, recent online DeePC formulations \cite{baros2022online, teutsch2023online, shi2024efficient, vahidi2024online} are emerging as direct data-driven counterparts to such nonlinear ML-based MPC approaches.


\textbf{Reduced-Order Data-Enabled Predictive Control:}
The Hankel matrix $\mathcal{H}$ may have a large dimension due to the required collected data length $T$. To address this, one can apply singular value decomposition (SVD) and retain only the dominant singular components, thereby constructing a reduced-order Hankel matrix. This approach not only mitigates the effect of noise in trajectory data but also reduces computational complexity and memory requirements for data-driven control methods~\cite{shi2023efficient}. Building on this idea, \cite{zhang2023dimension} incorporates SVD directly into the DeePC framework by defining a reduced-order Hankel matrix within the Fundamental Lemma. Considering $r = \mathrm{rank}(\mathcal{H})$, one can formulate the SVD of the Hankel matrix $\mathcal{H}$ as follows:
\begin{equation*}
  \begin{aligned}
    \label{SVD M}
    &\mathcal{H} = U \Sigma V^{\top}= [U_r \hspace{2 mm} U_{qK-r}] \begin{bmatrix}
    \Sigma_r & 0\\
    0 & 0
    \end{bmatrix} [V_r \hspace{2 mm} V_{L-r}]^{\top},
  \end{aligned}
\end{equation*}
where $q=m+p$, $\Sigma \in \mathbb{R}^{qK \times L}$ is the singular matrix, and $U \in \mathbb{R}^{qK \times qK}$ and $V \in \mathbb{R}^{L \times L}$ are left and right singular vectors, respectively, such that $UU^{\top}=U^{\top}U=I_{qK}$ and $VV^{\top}=V^{\top}V=I_L$. Moreover, $\Sigma_r$ contains the top $r$ non-zero singular values, $U_r \in \mathbb{R}^{qK \times r}$, $U_{qK-r} \in \mathbb{R}^{qK \times (qK-r)}$, and $V_r \in \mathbb{R}^{L \times r}$, $V_{L-r} \in \mathbb{R}^{L \times (L-r)}$. 
Therefore, one has
\begin{equation*}
  \begin{aligned}
    \label{reduced M}
    &\mathcal{H} g = U_r \Sigma_r V_r^{\top} g = \mathcal{H}' g',
  \end{aligned}
\end{equation*}
where $\mathcal{H}' = U_r \Sigma_r \in \mathbb{R}^{qK \times r}$ and $g' = V_r^{\top} g \in \mathbb{R}^r$. If the pre-collected data is sufficient rich, then we have $r = mK+n$ for the deterministic LTI system \eqref{system}, and $mK+n \leq r \leq \mathrm{min}(qK, L)$ for the general nonlinear system \eqref{system-Nonlinear}. Thus, one can approximate the Hankel matrix using a rank order $mK+n \leq r_a \leq r$ as follows:
\begin{equation*}
  \begin{aligned}
    \label{minimum M}
    &\mathcal{H} g \approx U_{r_a} \Sigma_{r_a} V_{r_a}^{\top} g = \mathcal{H}'' g'',
  \end{aligned}
\end{equation*}
where $\mathcal{H}'' = U_{r_a} \Sigma_{r_a} \in \mathbb{R}^{qK \times r_a}$ and $g'' = V_{r_a}^{\top} g \in \mathbb{R}^{r_a}$. It is worth noting that we have $\mathcal{H} g = \mathcal{H}' g' = \mathcal{H}'' g''$ for $r_a = r$.

Now, one can formulate a reduced-order DeePC as follows:
\begin{equation}
  \begin{aligned}
    \label{reduced-DeePC}
    &(\mathbf{g''}^{*},\mathbf{y}^{*},\mathbf{u}^{*}, \mathbf{\sigma_u}^{*}, \mathbf{\sigma_y}^{*}) = \underset{\mathbf{g''},\mathbf{y},\mathbf{u},\mathbf{\sigma_u},\mathbf{\sigma_y}}{\arg\min} \hspace{1 mm} J(\mathbf{y},\mathbf{u}, \mathbf{\sigma_u},\mathbf{\sigma_y}, \mathbf{g''})\\
    &s.t. \hspace{5 mm} \begin{bmatrix}
      U''_P\\ Y''_P\\ U''_F\\ Y''_F 
     \end{bmatrix} g'' = 
     \begin{bmatrix}
      u_{ini}\\ y_{ini}\\ u\\ y 
     \end{bmatrix} +
     \begin{bmatrix}
      \sigma_u\\ \sigma_y\\ 0\\ 0 
     \end{bmatrix}\\
    & \hspace{10 mm} E(y,u) \leq 0,
  \end{aligned}
\end{equation}
where the matrices $U''_P$, $U''_F$, $Y''_P$, and $Y''_F$ are obtained offline from $\mathcal{H}''$. Similarly to \eqref{DeePC g-nonlinear}, the problem is reformulated with $g''$ as the sole optimization variable. Consequently, the computational burden now depends on the dimension of $g''$, that is $r_a$, rather than on the dimension of $g$, that is $L$. Moreover, only the reduced Hankel matrix $\mathcal{H}''$ needs to be stored, requiring significantly less memory than the full Hankel matrix $\mathcal{H}$. The reduced-order DeePC is solved in exactly the same way as DeePC, and one can achieve a closed-form solution $\mathbf{u} = K_r^r \mathbf{r} + K_r^{ini} w_{ini}$ for the unconstrained QP case by setting the gradient of the quadratic cost to zero $(J_{g''} = 0)$. In terms of complexity, SPC does not offer a significant advantage over reduced-order DeePC in either computational cost or memory requirements. Importantly, the reduced-order DeePC with $r_a = r$ preserves DeePC’s low-bias behavior for nonlinear systems \eqref{system-Nonlinear}, in contrast to SPC. When truncation is applied $mk+n \leq r_a < r$, computational and memory demands are further reduced, but this introduces approximation error, thereby biasing DeePC’s behavior. Thus, reduced-order DeePC admits two regimes: if all nonzero singular directions of $\mathcal{H}'$ are retained, it is algebraically equivalent to DeePC, and the optimizer remains unchanged after a variable transformation. If the SVD is truncated $\mathcal{H}''$, the resulting optimizer generally differs—yielding lower variance but increased bias. Applying SVD and retaining only the dominant singular directions can be interpreted as fitting the data with a linear model of bounded complexity. In this sense, the choice of rank in the SVD truncation is equivalent to the choice of model complexity.

\vspace{2 mm}
\textbf{Kernel Representation:}
This framework was introduced in \cite{alsalti2024sample} under the name efficient data-driven predictive control (eDDPC). It should be noted, however, that all of the approaches discussed in this section can be viewed as forms of efficient data-driven predictive control. In the eDDPC formulation, a polynomial matrix $\mathcal{R}(z) \in \mathbb{R}^{v \times q}[z]$ of degree $l$ is defined as:
\begin{equation*}
  \begin{aligned}
    \label{kernel representation R}
    &\mathcal{R}(z) = \begin{bmatrix} r_1(z) \\ r_2(z) \\ \vdots \\ r_v(z) \end{bmatrix} = \begin{bmatrix} r_{1,0} + r_{1,1}z + \cdots + r_{1, l}z^{l} \\ 
    r_{2,0} + r_{2,1}z + \cdots + r_{2, l}z^{l} \\
    \vdots \\ r_{v,0} + r_{v,1}z + \cdots + r_{v,l}z^{l} \end{bmatrix}, \quad r_{i,j} \in \mathbb{R}^{q},
  \end{aligned}
\end{equation*}
where can be expressed as $\mathcal{R}(z) = \mathcal{R}_0 + \mathcal{R}_1 z + ...+ \mathcal{R}_l z^l$ with $\mathcal{R}_i \in \mathbb{R}^{v \times q}$. In the behavioral systems theory, forward shift operator $\sigma$ acts on a signal $w \in \mathbb{R}^{q}$ as $\sigma^j w_k = w_{k+j}$. In other words, $\sigma^j$ moves the trajectory $j$ steps forward in time. One can evaluate $\mathcal{R}(z)$ at the shift operator $\sigma$ to obtain an operator on the signal $w$ as $\mathcal{R}(\sigma) w = \mathcal{R}_0 w_k+ \mathcal{R}_1 w_{k+1} + ...+ \mathcal{R}_l w_{k+l}$. The Fundamental Lemma characterizes the behavior of the LTI system \eqref{system} as the image (column space) of the Hankel matrix. Equivalently, the same system admits a kernel (null-space) representation given by:
\begin{equation*}
  \begin{aligned}
    \label{kernel representation}
    &\mathcal{B} = \operatorname{im}(\mathcal{H}(w^d)) = \operatorname{\ker}(\mathcal{R}(\sigma)), 
  \end{aligned}
\end{equation*}
where $\mathcal{R}(\sigma) w^d = 0$ for any signal $w^d$ that belongs to the LTI system. We now use the following results to retrieve the kernel representation $\operatorname{\ker}(\mathcal{R}(\sigma))$ of the LTI system \eqref{system}. By exploiting the kernel structure of the Hankel matrix, one can obtain a full column-rank matrix whose image is equal to the finite-length behavior of the LTI system.

\begin{corollary}[Kernel Representation \cite{alsalti2024data}]
	Given $w^d$ as the signal of the controllable LTI system \eqref{system}, let $M \ge \ell + 1$. If $\operatorname{rank}(\mathcal{H}_M(w^d)) = mM + n$, the coefficients of the polynomial operator $\mathcal{R}(\sigma)$, i.e., $r_{i,j}$, are given by elements of full row-rank matrix $\mathcal{R}_M \in \mathbb{R}^{pM - n \times qM}$ as
    \begin{equation*}
    \begin{aligned}
    \mathcal{R}_M =
    \begin{bmatrix}
     r_{1,0} & r_{1,1} & \cdots & r_{1,M-1} \\
     r_{2,0} & r_{2,1} & \cdots & r_{2,M-1} \\
     \vdots  & \vdots  & \ddots & \vdots    \\
     r_{pM-n,0} & r_{pM-n,1} & \cdots & r_{pM-n,M-1}
     \end{bmatrix},
    \end{aligned}
    \end{equation*}
    which is the basis of the left null space of the Hankel matrix as $\mathcal{R}_M \mathcal{H}_M(w^d) = 0$.
\end{corollary}

\begin{corollary}[Non-Parametric Representation \cite{alsalti2024data}]
	Given $w^d$ as the signal of the controllable LTI system \eqref{system}, let $Z \ge M \ge \ell + 1$ and $\operatorname{rank}(\mathcal{H}_M(w^d)) = mM + n$. Then, $w$ is a signal of the LTI system if and only if there exists a unique vector $\beta \in \mathbb{R}^{mZ + n}$ such that
\begin{equation}
  \begin{aligned}
  \label{Non-Parametric Representation}
    & \mathcal{P} \beta = w,
  \end{aligned}
\end{equation}
where the full column-rank matrix $\mathcal{P} = \operatorname{null}(\Gamma) \in \mathbb{R}^{qZ \times mZ+n}$, the full row-rank matrix $\Gamma \in \mathbb{R}^{pZ - n \times qZ}$ is
{
\scriptsize
\begin{equation*}
  \begin{aligned}
    & \Gamma = \begin{bmatrix}
      r_{1,0} & r_{1,1} & \cdots & r_{1,M-1} & & \\
      r_{2,0} & r_{2,1} & \cdots & r_{2,M-1} & & \\
      \vdots & \vdots & \ddots & \vdots & & \\
      r_{pM-n,0} & r_{pM-n,1} & \cdots & r_{pM-n,M-1} & & \\
    & r_{1,0} & r_{1,1} & \cdots & r_{1,M-1} & \\
    & r_{2,0} & r_{2,1} & \cdots & r_{2,M-1} & \\
    & \vdots & \vdots & \ddots & \vdots & \\
    & r_{p,0} & r_{p,1} & \cdots & r_{p,M-1} &\\
    & & \ddots & \ddots & \ddots & \ddots &\\
    & & & r_{1,0} & r_{1,1} & \cdots & r_{1,M-1}\\
    & & & r_{2,0} & r_{2,1} & \cdots & r_{2,M-1}\\
    & & & \vdots & \vdots & \ddots & \vdots\\
    & & & r_{p,0} & r_{p,1} & \cdots & r_{p,M-1}\\
\end{bmatrix},
  \end{aligned}
\end{equation*}
} 
$\Gamma \mathcal{H}_Z(w^d) = 0$, and $r_{i,j}$ are the elements of the matrix $\mathcal{R}_M$ in Corollary 1.
\end{corollary}

Comparing the Fundamental Lemma and Corollary 2 under $Z = K = T_{ini} + N$, both provide a non-parametric representation of the finite-length behavior of the controllable LTI system. This is true since $\operatorname{im}(\mathcal{H}_L(w^{\text{data}})) = \operatorname{im}(\mathcal{P}) = \mathcal{B}$. However, \eqref{Non-Parametric Representation} is an efficient non-parametric representation since the Fundamental Lemma requires $T \ge (m + 1)(K + n) - 1$ data points, whereas Corollary 2 requires $T \ge (m + 1)(\ell + n + 1) - 1$. If the minimum $T$ is chosen in both cases, Corollary 2 requires $(m + 1)(K - \ell - 1)$ less samples, for any $K > \ell + 1$. By the way, it is only a big advantage for large systems and/or large $K$. Note that the Fundamental Lemma has infinitely many solutions for $g \in \mathbb{R}^{L}$, whereas Corollary 2 presents a unique solution $\beta \in \mathbb{R}^{mK+n}$ for each trajectory. 
Finally, the dimension of $\beta$ is independent of the number of collected data points, whereas the dimension of $g$ increases with increasing $T$, which helps the controller to require less computation and memory. One has $rank(\mathcal{P}) = rank(\mathcal{H}_Z(w^d)) = mZ + n$ for the deterministic linear system \eqref{system}. In addition, for the general nonlinear system \eqref{system-Nonlinear}, $M$ must be large enough to enhance the performance of the non-parametric representation \eqref{Non-Parametric Representation} on system's behavior prediction. 

Now, similarly to \eqref{DeePC-Nonlinear}, one can formulate eDDPC as follows \cite{alsalti2024sample, alsalti2024robust}:
\begin{equation}
  \begin{aligned}
    \label{eDDPC}
    &(\mathbf{\beta}^{*},\mathbf{y}^{*},\mathbf{u}^{*}, \mathbf{\sigma_u}^{*}, \mathbf{\sigma_y}^{*}) = \underset{\mathbf{\beta},\mathbf{y},\mathbf{u}, \mathbf{\sigma_u}, \mathbf{\sigma_y}}{\arg\min} \hspace{1 mm} J(\mathbf{y},\mathbf{u},\mathbf{\sigma_u},\mathbf{\sigma_y}, \mathbf{\beta})\\
    &s.t. \hspace{5 mm} \mathcal{P} \beta = 
     \begin{bmatrix}
      u_{ini}\\ y_{ini}\\ u\\ y 
     \end{bmatrix} +
     \begin{bmatrix}
      \sigma_u\\ \sigma_y\\ 0\\ 0 
     \end{bmatrix} \\
    & \hspace{10 mm} E(y,u) \leq 0,
  \end{aligned}
\end{equation}
where the data matrix $\mathcal{P}$ is full column rank and is obtained offline using Corollary 2. Moreover, similarly to \eqref{DeePC g-nonlinear}, one can partition the data matrix $\mathcal{P} = [\mathcal{P}^\top_{u_{ini}}, \mathcal{P}^\top_{y_{ini}}, \mathcal{P}^\top_u, \mathcal{P}^\top_y]^\top$ and rewrite \eqref{eDDPC} as follows:
\begin{equation*}
  \begin{aligned}
    \label{eDeePC-b}
    &\mathbf{\beta}^{*} = \underset{\mathbf{\beta}}{\arg\min} \hspace{1 mm} J(\mathcal{P}_y\mathbf{\beta},\mathcal{P}_u\mathbf{\beta},\mathcal{P}_{u_{ini}} \mathbf{\beta} - u_{ini}, \mathcal{P}_{y_{ini}} \mathbf{\beta} - y_{ini}, \mathbf{\beta})\\
    &s.t. \hspace{5 mm} E(\mathcal{P}_y\mathbf{\beta},\mathcal{P}_u\mathbf{\beta}) \leq 0,
  \end{aligned}
\end{equation*}
where is an optimization problem on $\beta$. Therefore, the computational burden depends on the dimension of $\beta$, that is $mK+n$, instead of the dimension of $g$, that is $L$. In addition, we only need to save $\mathcal{P}$, which requires much less memory compared to the Hankel matrix $\mathcal{H}$. eDDPC is exactly solved as DeePC, and one can achieve a closed-form solution $\mathbf{u} = K_e^r \mathbf{r} + K_e^{ini} w_{ini}$ for the unconstrained QP version by setting the gradient of the quadratic cost to zero $(J_{\beta} = 0)$. Comparing the reduced-order DeePC (with $\mathcal{H}'$) and eDDPC, both are in fact equivalent to DeePC for the deterministic LTI systems $\operatorname{im}(\mathcal{P}) = \operatorname{im}(\mathcal{H}') = \operatorname{im}(\mathcal{H})$; thus, the resulting closed-loop trajectories are identical. The reduced-order DeePC and eDDPC have the same improvement in computation and memory burden as $\mathcal{H}' \in \mathbb{R}^{qK \times mK+n}$ and $\mathcal{P} \in \mathbb{R}^{qK \times mK+n}$; however, eDDPC is a sample efficient one for large systems and/or large $K$ (requires less collected data points). On the other hand, eDDPC introduces new challenges: as the system complexity $(m, p, n, \ell)$ increases, some numerical problems are encountered when constructing the matrix $\Gamma$. Specifically, taking the first $p$ rows of the matrix $\mathcal{R}_M$ can sometimes result in an ill-conditioned matrix $\Gamma$, i.e., very large condition number (the ratio between the largest and smallest non-zero singular values), which might lead to errors in computing $\mathcal{P} = null(\Gamma)$. One explanation is that the kernel representation specified by the rows of $\mathcal{R}_M$ is not necessarily minimal. In addition, although the reduced-order DeePC may require a little bit more computation and memory for the nonlinear system \eqref{system-Nonlinear}, it has the same behavior as DeePC on the bias-variance trade-off. However, eDDPC is not guaranteed to have the same behavior, but it can be similar to them. Thus, one may conclude that the reduced-order DeePC is a better choice.

\vspace{2 mm}
\textbf{Range Space Equivalence:}
By exploiting the range space (column space) of the Hankel matrix, i.e., $\operatorname{im}(\mathcal{H})$, DeePC can be reformulated to reduce computational cost. This reformulation replaces the Hankel matrix with a lower-dimensional representation of its column space, while remaining algebraically equivalent to the original problem. The derivation of this reformulation relies on the following lemma:

\begin{lemma} [Range Space Equivalence \cite{faye2024computationally}]
Let $\Psi \in \mathbb{R}^{L \times L}$ be a positive-definite matrix, i.e., $\Psi > 0$. Then, for the Hankel matrix $\mathcal{H} \in \mathbb{R}^{qK \times L}$, one has $\operatorname{im}(\mathcal{H}) = \operatorname{im}(\mathcal{H} \Psi^{-1} \mathcal{H}^\top)$. In particular, with $\Psi = I_L$, $\operatorname{im}(\mathcal{H}) = \operatorname{im}(\mathcal{H}\mathcal{H}^\top)$.
\end{lemma}

Using the above lemma, one can formulate range-space DeePC as follows \cite{faye2024computationally}:
\begin{equation}
  \begin{aligned}
    \label{rDeePC}
    &(\mathbf{\alpha}^{*},\mathbf{y}^{*},\mathbf{u}^{*}, \mathbf{\sigma_u}^{*}, \mathbf{\sigma_y}^{*}) = \underset{\mathbf{\alpha},\mathbf{y},\mathbf{u}, \mathbf{\sigma_u}, \mathbf{\sigma_y}}{\arg\min} \hspace{1 mm} J(\mathbf{y},\mathbf{u},\mathbf{\sigma_u},\mathbf{\sigma_y}, \mathbf{\alpha})\\
    &s.t. \hspace{5 mm} \mathcal{G} \alpha = 
     \begin{bmatrix}
      u_{ini}\\ y_{ini}\\ u\\ y 
     \end{bmatrix} +
     \begin{bmatrix}
      \sigma_u\\ \sigma_y\\ 0\\ 0 
     \end{bmatrix} \\
    & \hspace{10 mm} E(y,u) \leq 0,
  \end{aligned}
\end{equation}
where the matrix $\mathcal{G} = \mathcal{H} \Gamma^{-1} \mathcal{H}^\top \in \mathbb{R}^{qK \times qK}$ is obtained offline. Moreover, under the quadratic regularization term $\|g\|_\Psi^2$ for DeePC, one has the quadratic regularization term $\|\alpha\|_\mathcal{G}^2$ for the range-space DeePC. Similarly to \eqref{eDDPC}, the problem is rewritten to only has $\alpha$ as the optimization variable. Therefore, the computational burden depends on the dimension of $\alpha$, that is $qK$, instead of the dimension of $g$, that is $L$. In addition, we only need to save $\mathcal{G}$, which requires much less memory compared to the Hankel matrix $\mathcal{H}$. The range-space DeePC is exactly solved as DeePC, and one can achieve a closed-form solution $\mathbf{u} = K_r^r \mathbf{r} + K_r^{ini} w_{ini}$ for the unconstrained QP version by setting the gradient of the quadratic cost to zero $(J_{\alpha} = 0)$. Both the range-space DeePC and the reduced-order DeePC with $r_a = r$ keeps DeePC's behavior on the bias-variance trade-off; however, the reduced-order DeePC requires lower or same computation and memory. In addition, if $\lVert g\rVert_\Psi^2$ is strictly convex, there is no guarantee that $\lVert \alpha\rVert_\mathcal{G}^2$ remains strictly convex, since if $\mathcal{H}$ does not have full row rank (and we only assume $\Psi\succ0$), then $\mathcal{G}\succeq0$. Finally, $\mathcal{G}$ has a worse condition number than $\mathcal{H}$ (e.g.\ with $\Psi=I$, $\mathrm{cond}(\mathcal{G})=\mathrm{cond}(\mathcal{H})^2$), which can deteriorate numerical accuracy or require more iterations in optimization algorithms. Therefore, one may prefer to stay with the reduced-order DeePC as the best choice. Note that these modified efficient versions of DeePC are guaranteed to be equivalent to DeePC only when the regularization term is a positive-definite quadratic form, such as $\|g\|_\Psi^2$ (i.e., an $\ell_2$ norm penalty). 

\vspace{2 mm}
\textbf{Discrete Fourier Transform:}
Factorizations based on the discrete Fourier transform (DFT) of the Hankel matrix enable fast and memory-efficient computations. To exploit this in DeePC, one must employ DFT-based iterative optimization algorithms (e.g., gradient-descent variants) for the augmented cost function, thereby avoiding the need to store the Hankel matrix explicitly. \cite{schmitz2024fast} proposes a two-step procedure: first, they derive a DFT-based factorization that enables efficient matrix–vector multiplications involving the Hankel matrix; then, they integrate this factorization into DeePC by reformulating the augmented cost function and solving it with an iterative optimal control algorithm. This combination allows DeePC to be solved without storing the Hankel matrix, requiring only the implementation of a matrix–vector product operator based on the data $w$ and its DFT. Unlike SVD-based reductions, the DFT-based approach preserves the full DeePC model, and its efficiency gain arises purely from computational structure rather than from modifying the feasible set.

A straightforward approach to calculate the matrix-vector product $\mathcal{H}(w^d)g$ without storing the full Hankel matrix $\mathcal{H}(w^d)$ is to make use of the structure of the Hankel matrix and access its coefficients directly from the generating sequence $w^d$. A factorization of $\mathcal{H}(w^d)$ likewise solves the memory problem and in addition is favorable for the speed of matrix-vector multiplication. A row permutation of $\mathcal{H}(w^d)$ leads to the Toeplitz matrix $\Xi \in \mathbb{R}^{qK \times L}$ as follows:
\begin{equation*}
\begin{aligned}
\label{DFT}
    & \Xi := \begin{bmatrix}
        w^d_{K} & \cdots & w^d_{T} \\
        \vdots & \ddots & \vdots \\
        w^d_1 & \cdots & w^d_{L}
         \end{bmatrix} = \begin{bmatrix}
         & & I_q \\
         & \iddots & \\
        I_q &  & 
    \end{bmatrix} \mathcal{H}(w^d),
    \end{aligned}
\end{equation*}
which can be embedded into a circulant block-Toeplitz matrix $\Upsilon \in \mathbb{R}^{qT \times T}$ with $\Xi = \left[ I_{qK} \ 0_{qK \times q(T-K)} \right] \Upsilon \left[ I_{L} \ 0_{L \times (T-L)} \right]^\top$ as:
{\small
\[
\Upsilon = \begin{bmatrix}
    w^d_{K} & w^d_{K+1} & \cdots & w^d_{T} & w^d_1 & w^d_2 & \cdots & w^d_{K-1} \\
    \vdots & \vdots & \ddots & \vdots & \vdots & \vdots & \ddots & \vdots \\
    w^d_1 & w^d_2 & \cdots & w^d_{L} & w^d_{L+1} & w^d_{L+2} & \cdots & w^d_{T}\\
    w^d_{T} & w^d_{1} & \cdots & w^d_{L-1} & w^d_L & w^d_{L+1} & \cdots & w^d_{T-1} \\
    \vdots & \vdots & \ddots & \vdots & \vdots & \vdots & \ddots & \vdots \\
    w^d_{K+1} & w^d_{K+2} & \cdots & w^d_{1} & w^d_2 & w^d_3 & \cdots & w^d_{K} 
\end{bmatrix}.
\]
}

Then, the matrix $\Upsilon$ can be factorized as $\Upsilon = \tilde{F} \Lambda F^{-1}$, where $F := [e^{-2\pi ikj/N}]_{k,j=1,\ldots,T}$ is a Fourier matrix in $\mathbb{R}^{T \times T}$, $\tilde{F} := F \otimes I_q \in \mathbb{C}^{qT \times qT}$, and $\Lambda := \operatorname{diag}(\Lambda_1, \dots, \Lambda_{T}) \in \mathbb{C}^{qT \times T}$ with
\[
\begin{bmatrix}
    \Lambda_1 \\
    \vdots \\
    \Lambda_T
\end{bmatrix} := \tilde{F} \begin{bmatrix}
    w_{K} \\
    \vdots \\
    w_{T} \\
    w_1 \\
    \vdots \\
    w_{K-1}
\end{bmatrix}.
\]

Considering $v = \begin{bmatrix} g^\top & 0^\top_{T-L} \end{bmatrix}^\top$, the matrix-vector product $\Upsilon v$ can be efficiently computed since $F^{-1} v = \frac{1}{T} F^* v$ can be evaluated in $\mathcal{O}(T \log T)$ time, $\Lambda v$ can be evaluated in $\mathcal{O}(qT)$ time, and 
\[
\tilde{F}v = (F \otimes I_q)v = \left[ \left( e^{-2 \pi i/N} \right)^{kj} I_q \right]_{k,j=1,\ldots,T} v
\]
can be evaluated in $\mathcal{O}(qT \log T)$ time. As a consequence, the matrix-vector product $\mathcal{H}(w^d)g$ can be efficiently calculated via the following three steps: 1) zero-padding $v = \begin{bmatrix} g^\top & 0^\top_{T-L} \end{bmatrix}^\top$, 2) computing $\Upsilon v$ using DFT-based factorization, and 3) truncating the result and permuting its components $\Xi$ and $\mathcal{H}(w^d)g$. Therefore, the matrix-vector multiplication $\mathcal{H}(w^d)g$ has complexity $\mathcal{O}(qT \log T)$.

\begin{remark} [Small Prime Factors]
Most standard implementations of DFT run significantly faster when the signal length $T$ has only small prime factors (not greater than $7$). Therefore, apart from the PE condition, this fact can be taken into account when choosing the length of the signal.
\end{remark}

\begin{remark} [Ill-Conditioning]
Since the convergence rate of iterative solvers often hinges on a good condition number of the Hankel matrix, \cite{schmitz2024fast} suggests Broyden-Fletcher-Golfarb-Shanno (BFGS) algorithm as iterative optimization algorithm to constructs a preconditioner of the Hankel matrix iteratively and guarantee fast convergence. To reduce the effect of inherent bad conditioning of the Hankel matrix, the limited memory Broyden-Fletcher-Golfarb-Shanno (lBFGS) algorithm is used to minimize the augmented cost function. This is due to the fact that, whereas the BFGS method is a quasi-Newton method and hence does not suffer from ill-conditioning as much as the gradient method, the preconditioner is built up successively via rank-one updates. 
While running, the BFGS algorithm constructs a preconditioner of the Hankel matrix which is improved in every iteration step. These updates of the preconditioner can be performed efficiently with the DFT-based factorizations presented before.
\end{remark}

DFT-based formulation is just a matrix-free way to solve the same DeePC. When it converges, it returns exactly the same optimizer as the standard DeePC. Thus, it is identical to DeePC in bias-variance trade-off because it only accelerates the linear algebra. However, DFT is usually usefull in online DeePC since we have $\mathcal{O}(qT \log T)$ for each DeePC solve. For DeePC, the reduced-order DeePC is a better option since we obtain $\mathcal{H}'$ offline. Moreover, online numerical SVD algorithms offer a complementary route for online DeePC \cite{vahidi2024online,shi2024efficient}. By maintaining a low-dimensional basis and updating it as new data arrive, online reduced-order DeePC can keep each solve small and is competitive when the trajectory subspace has low numerical rank and evolves slowly. In contrast, DFT-based DeePC avoids forming Hankel matrices, making it preferable when memory is tight or the data window changes rapidly. 

\vspace{2 mm}
\textbf{Learning-based Approximation:}
One can decouple control performance and model fitness in DeePC such that it can be viewed as minimizing a control cost and a score cost, where the evaluation of the latter is costly due to the size of the Hankel matrix \cite{zhou2024learning}. Therefore, the scoring cost function can be approximated via differentiable convex programming, where the parameters of the approximated function are learned offline. Finally, an approximate control law is formulated to minimize the summation of the control cost and the learned scoring function. Thus, one can reformulate DeePC to separate the cost function into two parts as follows:
\begin{equation*}
(\mathbf{y}^{*},\mathbf{u}^{*}) = \underset{\mathbf{y},\mathbf{u}}{\arg\min} \hspace{1 mm} O_c(\mathbf{y},\mathbf{u}) + O_m(\mathbf{y},\mathbf{u}),
\end{equation*}
where
\begin{equation*}
  \begin{aligned}
    &O_c(\mathbf{y},\mathbf{u}) = J(\mathbf{y},\mathbf{u})\\
    &s.t. \hspace{5 mm} E(y,u) \leq 0,
  \end{aligned}
\end{equation*}
and
\begin{equation*}
  \begin{aligned}
    &O_m(\mathbf{y},\mathbf{u}) =  \min_{\mathbf{g}, \mathbf{\sigma_u},\mathbf{\sigma_y}} \, J(\mathbf{\sigma_u},\mathbf{\sigma_y},\mathbf{g})\\
    &s.t. \hspace{5 mm} \begin{bmatrix}
      U_P\\ Y_P\\ U_F\\ Y_F 
     \end{bmatrix} g = 
     \begin{bmatrix}
      u_{ini}\\ y_{ini}\\ u\\ y 
     \end{bmatrix} + \begin{bmatrix}
      \sigma_u\\ \sigma_y\\ 0\\ 0 
     \end{bmatrix}.
  \end{aligned}
\end{equation*}
where $O_c(y,u)$ represents the cost function of the control problem, which is distinct from and unaffected by the system dynamics. Also, $O_m(y,u)$ is referred to as the scoring function, which evaluates the fitness of $w$ to the trajectory segments collected in the data matrix. Once a candidate trajectory $(y,u)$ is proposed, the scoring function $O_m(y,u)$ simply maps $(y,u)$ to a scalar. Thus, once $(y,u)$ is assembled, $O_m(y,u)$ gives its score on model fitness. $O_m(y,u)$ depends only on the system and is not influenced by any specific cost or safety constraint. Note that $(g, \sigma_u, \sigma_y)$ are not part of the control policy and only evaluate the scoring function, where the dimension of $g$ depends on the number of the collected data. Therefore, we desire to accelerate the evaluation of $O_m(y,u)$ using a learning-based approach which approximates the scoring function with fewer variables. We denote the approximate scoring function as $\hat{O}_m(y,u)$, then the overall control problem becomes:
\begin{equation*}
    \min_{\mathbf{y},\mathbf{u}} \, O_c(\mathbf{y},\mathbf{u}) + \hat{O}_m (\mathbf{y},\mathbf{u}),
\end{equation*}
where $\hat{O}_m(y,u)$ is a mapping from an I/O signal $(u,y)$ to the value of the corresponding scoring function and is learned offline. Here, the learned $\hat{O}_m(y,u)$ is used to
substitute the original scoring function $O_m(y,u)$ and efficiently solve the above control problem online. We refer the interested readers to \cite{zhou2024learning} for further details on the offline learning process of $\hat{O}_m(y,u)$. For this approach, the computation time is mainly determined by the size of the learned scoring function $\hat{O}_m(y,u)$, and there is also a trade-off between control performance and computation time. Therefore, \cite{zhang2024deep} uses deep learning to directly learn the DeePC policy as $\hat{g} = F_\theta(r, w_{ini})$, which is called Deep DeePC. However, the main advantage of DeePC is avoiding the offline learning process for control systems; thus, these two approaches may not be perfectly aligned with DeePC's goal. Moreover, they introduce a new challenge for DeePC, which is learning errors.

\vspace{2 mm}
\textbf{Data-Enabled Neighboring Extremal:}
One can use neighboring extremal (NE) adaptation \cite{ghaemi2008neighboring} to approximate the DeePC solution. Considering DeePC \eqref{DeePC-Nonlinear}, we have a nominal solution $(u^{o},y^{o})$ for an initial I/O trajectory $(u_{ini}^o,y_{ini}^o)$ and reference trajectory $r^o$. Now, for a new initial I/O trajectory $(u_{ini},y_{ini})$ and/or reference trajectory $r$, the DeePC solution is approximated by $u^* = u^{o} + \delta u$ using NE adaptation. The objective is to compute $\delta u$ using data-enabled neighboring extremal (DeeNE) for the optimization problem \eqref{DeePC-Nonlinear}. The resulting equation is a closed-form approximation of the DeePC solution, which helps us to reduce the computational time of DeePC.

Considering \eqref{Aug Cost DeePC-nonlinear}, the augmented cost function can be rewritten as:
\begin{equation}
  \begin{aligned}
    \label{aug-cost}
  &\bar{J} (w_{ini},g,r,\mu) = J (w_{ini},g,r) + \mu^{T} E^a (g).
    \end{aligned}
\end{equation}

\begin{assumption}[Active Constraints]
\label{ass2}
$E^a_{g}(g)$ is full row rank.
\end{assumption}

Since $(w^o_{ini},g^o,r^o)$ represents the nominal solution for the DeePC \eqref{DeePC-Nonlinear}, they satisfy the following necessary optimality conditions obtained from the KKT conditions for the augmented cost function \eqref{aug-cost} as follows:
\begin{equation*}
  \begin{aligned}
    \label{KKT DeePC z}
    &\bar{J}_g (w_{ini},g,r,\mu) = 0, \\
    &\mu \geq 0.
  \end{aligned}
\end{equation*}

Now, using the above KKT conditions and the nominal solution $(w^o_{ini},g^o,r^o)$, one can calculate the Lagrange multiplier $\mu$ online as:
\begin{equation*}
  \begin{aligned}
    \label{Optimal KKT}
    &J_g (w^o_{ini},g^o,r^o) + \mu^{T} E^a_g (g^o) = 0,
  \end{aligned}
\end{equation*}
and
\begin{equation*}
  \begin{aligned}
    \label{Lagrange multipliers}
    &\mu = -{(E^a_{g} {E^a_{g}}^\top)}^{-1} E^a_{g} J^\top_g.
  \end{aligned}
\end{equation*}

Note that Assumption \ref{ass2} guarantees that $E^a_{g} {E^a_{g}}^\top$ is invertible. Moreover, it is worth noting that $\mu = 0$ if the constraint $E(g^o)$ is not active. The Lagrange multiplier is considered as the nominal Lagrange multiplier $\mu^{o}$.

Adapting to initial I/O perturbation and/or reference perturbation, DeeNE seeks to minimize the second-order variation of \eqref{aug-cost} subject to linearized constraints. More specifically, given $\delta w_{ini}$ and $\delta r$, DeeNE solves the following optimization problem as:
\begin{equation}
  \begin{aligned}
    \label{DeeNE}
    &\mathbf{\delta g^{*}} = \underset{\mathbf{\delta g}}{\arg\min} \hspace{1 mm} {J}^{ne}\\ 
    & s.t. \hspace{5 mm} E^{a}_{g} \delta g = 0,
  \end{aligned}
\end{equation}
where 
\begin{equation*}
  \begin{aligned}
    \label{DeeNE-cost}
  &{J}^{ne}=\delta^{2} \bar{J} \\
  & \hspace{6 mm}= \frac{1}{2} 
  \begin{bmatrix}
  \delta w_{ini}\\ \delta g\\ \delta r
  \end{bmatrix}^{T}
  \begin{bmatrix}
  \bar{J}_{w_{ini}w_{ini}} & \bar{J}_{w_{ini}g} & \bar{J}_{w_{ini}r}\\
  \bar{J}_{gw_{ini}} & \bar{J}_{gg} & \bar{J}_{gr}\\
  \bar{J}_{rw_{ini}} & \bar{J}_{rg} & \bar{J}_{rr}
  \end{bmatrix}
  \begin{bmatrix}
  \delta w_{ini}\\ \delta g\\ \delta r
  \end{bmatrix}.
    \end{aligned}
\end{equation*}

For \eqref{DeeNE}, the augmented cost function are obtained as:
\begin{equation*}
  \begin{aligned}
    \label{DeeNE-aug-cost}
    &\bar{J}^{ne} = \frac{1}{2} 
    \begin{bmatrix}
  \delta w_{ini}\\ \delta g\\ \delta r
  \end{bmatrix}^{T}
  \begin{bmatrix}
  \bar{J}_{w_{ini}w_{ini}} & \bar{J}_{w_{ini}g} & \bar{J}_{w_{ini}r}\\
  \bar{J}_{gw_{ini}} & \bar{J}_{gg} & \bar{J}_{gr}\\
  \bar{J}_{rw_{ini}} & \bar{J}_{rg} & \bar{J}_{rr}
  \end{bmatrix}
  \begin{bmatrix}
  \delta w_{ini}\\ \delta g\\ \delta r
  \end{bmatrix} \\
    & \hspace{9 mm}+ \delta \mu^{T} E^{a}_{g} \delta g,
  \end{aligned}
\end{equation*}
where $\delta \mu$ is the Lagrange multiplier for \eqref{DeeNE}. By applying the KKT conditions to the above augmented cost function, one has
\begin{equation*}
  \begin{aligned}
    \label{DeeNE-KKT}
    &\bar{J}^{ne}_{\delta g} = 0, \\
    &\delta \mu \geq 0.
      \end{aligned}
\end{equation*}

The following theorem presents the proposed DeeNE to approximate the DeepC policy in the presence of initial input-output and reference perturbations.

\begin{theorem} [Data-Enabled Neighboring Extremal \cite{vahidi2023data}] 
\label{theo1}
Consider the optimization problem \eqref{DeeNE}. If $\bar{J}_{gg} > 0$, DeeNE policy
\begin{equation}
  \begin{aligned}
    \label{law}
    &\delta g = K^{*}_1 \delta w_{ini} + K^{*}_2 \delta r,\\
    &K^{*}_1 = -
    \begin{bmatrix}
    I & 0
    \end{bmatrix}
    K^o
    \begin{bmatrix}
    \bar{J}_{gw_{ini}}\\
    0
    \end{bmatrix},\\
    &K^{*}_2 = -
    \begin{bmatrix}
    I & 0
    \end{bmatrix}
    K^o
    \begin{bmatrix}
    \bar{J}_{gr}\\
    0
    \end{bmatrix},\\
    &K^{o} =\begin{bmatrix}
    \bar{J}_{gg} & {E^a_{g}}^\top\\
    E^a_{g} & 0
    \end{bmatrix}^{-1}
  \end{aligned}
\end{equation}
approximates the perturbed solution of the nominal DeePC \eqref{DeePC-Nonlinear} with initial I/O perturbation $\delta w_{ini}$ and reference perturbation $\delta r$.
\end{theorem}

\begin{remark} [Singularity]
\label{Singularity}
It is worth noting that $\bar{J}_{gg} > 0$ is essential for DeeNE since it guarantees the convexity of \eqref{DeeNE}. Under Assumption \ref{ass2} and $\bar{J}_{gg} > 0$, $K^o$ is well defined. If $E^a_{g}$ is not full row rank, the matrix $K^o$ is singular, leading to the failure of DeeNE. This issue can be solved using constraint back-propagation algorithm \cite{ghaemi2008neighboring}.
\end{remark}

Using DeeNE \eqref{law}, one can obtain $g^* = g^o + \delta g$ and then $u^* = U_F g^*$. Under DeeNE, one has $\delta u = K_{ne}^r \delta r + K_{ne}^{ini} \delta w_{ini}$, which shows that the online computation grows linearly with the prediction horizon. DeeNE typically requires lower computation than DeePC and reduced-order DeePC because it does not need to solve an optimization problem at each time step, just applying precomputed matrices. If perturbations become large or the active constraint set changes a lot, we may need to refresh the nominal solution, at which point the advantage narrows. On the other hand, DeeNE requires saving the matrices $K^*_1 \in \mathbb{R}^{L \times qT_{ini}}$ and $K^*_2 \in \mathbb{R}^{L \times pN}$, which depend on the length of collected data $T$; thus, the reduced-order DeePC requires less memory. However, it is easy to develop DeeNE for the reduced-order DeePC to have even less computation burden and also require less memory. Finally, DeeNE is a local approximation of the perturbed DeePC optimum; therefore, statistical bias-variance behavior is the same as DeePC. The only extra bias is algorithmic/approximation error, which is small.

\section{Real-Time Implementation}
While data-driven optimal control policies have been extensively explored in theory and validated through simulations, their true impact is most evident when demonstrated in experimental settings and real-time applications. Deploying such methods on physical systems provides critical insights into robustness, adaptability, and scalability under practical constraints. Experimental studies have already confirmed the effectiveness of DeePC in diverse contexts, including quadcopter flight \cite{elokda2021data}, synchronous motor drives \cite{carlet2022data}, and Li-ion batteries fast charging \cite{vahidi2024online}. Moreover, using efficient computational techniques, we have enabled successful real-time implementations in robotic arms, soft robotic manipulators, and autonomous driving platforms. In what follows, we review representative case studies, with particular emphasis on our own experimental results, to highlight the practical capabilities of efficient DeePC in real-world systems. These studies bridge the gap between theoretical development and practical realization, underscoring efficient DeePC’s potential as a versatile paradigm for data-driven optimal policy.

\begin{figure}[!h]
     \centering
     \includegraphics[width=0.99\linewidth]{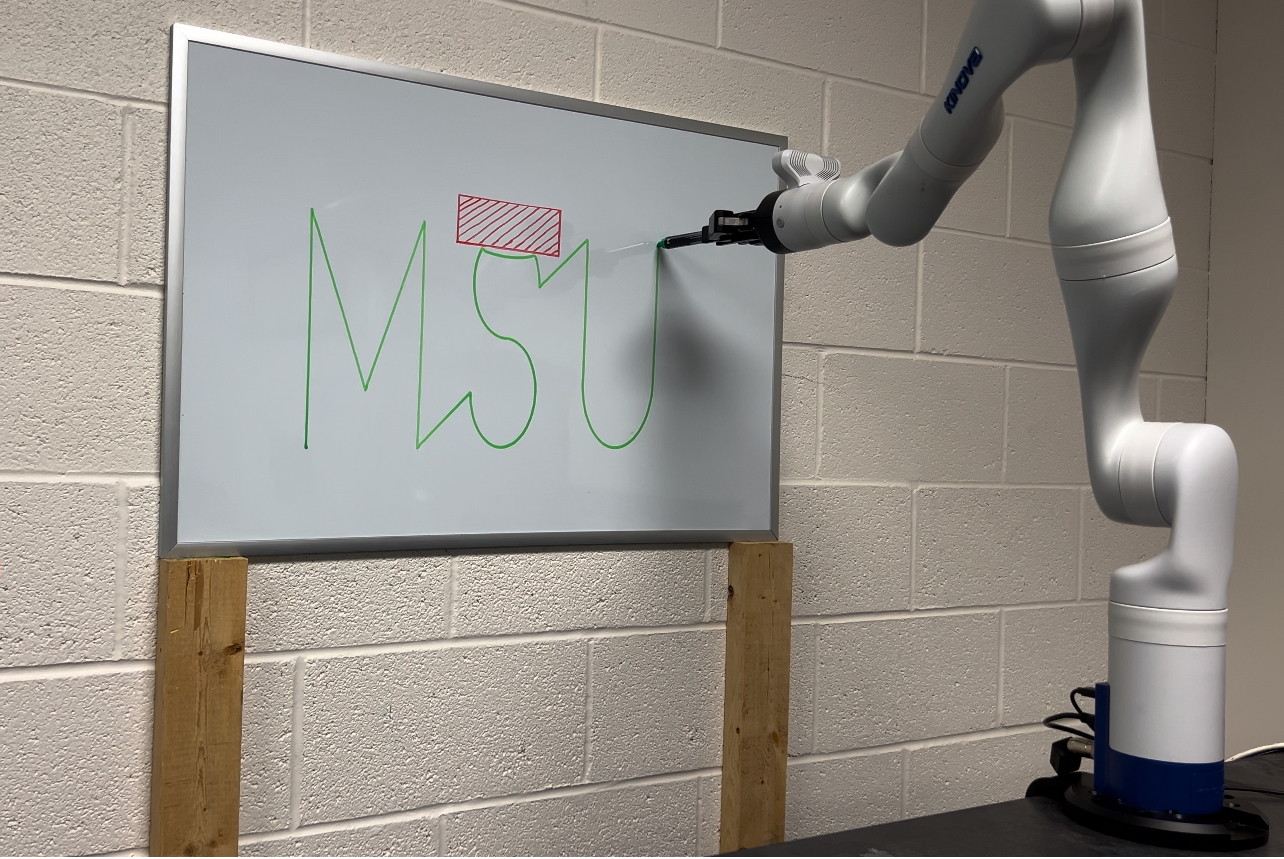}
    \caption{Safe Reference Tracking by 7-DoF Robotic Arm. Experimental results video: \protect\url{https://www.youtube.com/watch?v=BlKTUgkAMVo}}
    \label{Robotic_Arm}
 \end{figure}

\begin{figure}[!h]
     \centering
     \includegraphics[width=0.99\linewidth]{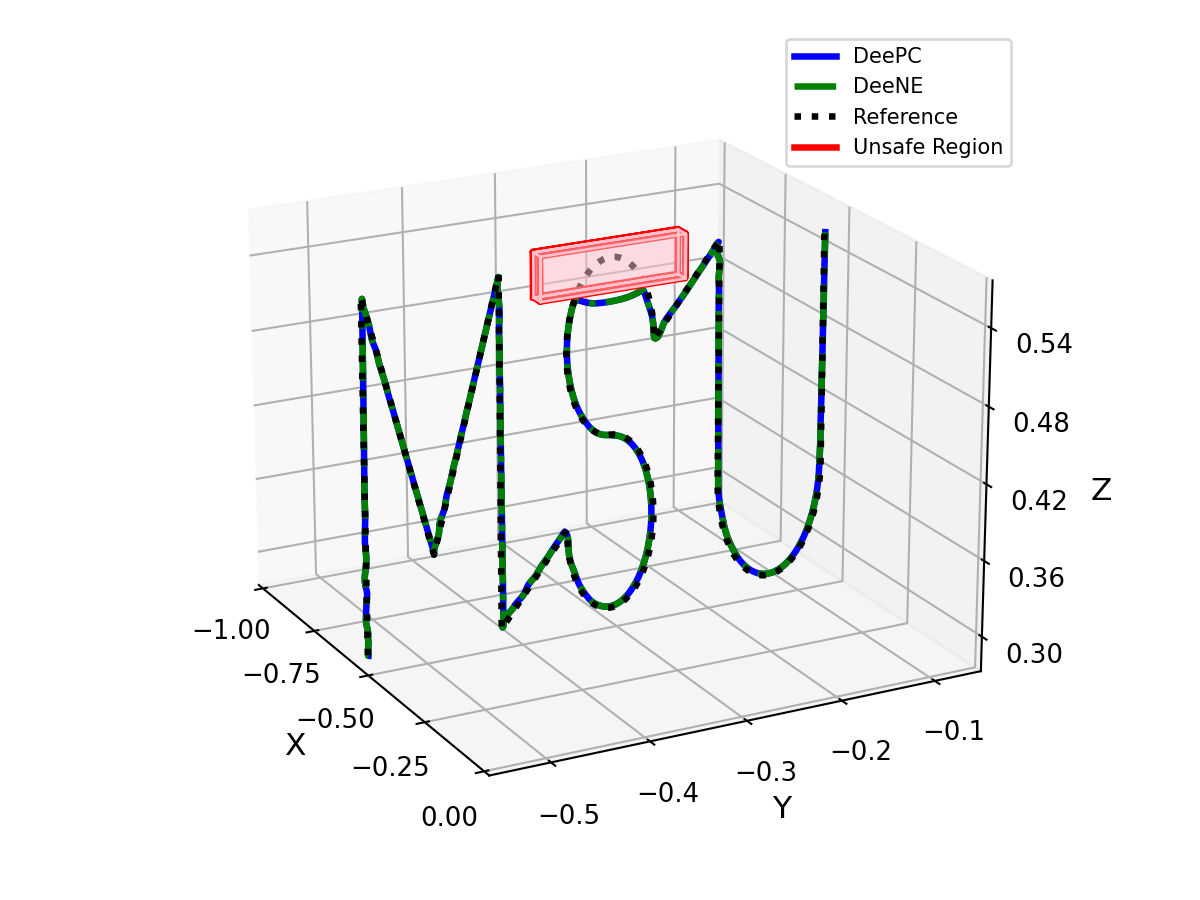}
    \caption{Safe reference tracking for 7-DoF Robotic Arm.}
     \label{Outputs3D_Safe}
 \end{figure}

\vspace{2 mm}
\textbf{7-DoF Robotic Arm Motion Control:}
We conducted an experimental study on a 7-DoF KINOVA Gen3 robotic manipulator to evaluate the performance of DeePC and DeeNE in safe reference tracking \cite{vahidi2025data}, as illustrated in Fig. \ref{Robotic_Arm}. The system was subject to constraints including joint velocity limits, workspace boundaries, and laboratory safety considerations. The experimental task required the robot to draw the pattern ``MSU'' on a board while avoiding a designated unsafe region (represented by the red box) along the reference trajectory. In contrast to model-based approaches, which demand the derivation of forward and inverse kinematics for high-DoF manipulators---a process that is both complex and calibration-intensive---data-driven optimal control policies bypass these challenges by directly exploiting I/O data. For this study, data were collected from 50 randomized trajectories and organized into mosaic Hankel matrices for DeePC. Both DeePC and DeeNE were deployed in closed-loop operation and successfully satisfied the imposed constraints, as demonstrated in Fig. \ref{Outputs3D_Safe}. Quantitative results, Table II, show that DeePC provided accurate trajectory tracking but suffered from a computation time of 0.2\,s, which exceeded the hardware sampling rate of 0.1\,s. This mismatch led to delayed control updates, producing discontinuous and slowed robot motion that limits DeePC’s real-time applicability. In contrast, DeeNE reduced computation time to 0.03\,s while maintaining comparable tracking accuracy, thereby enabling smooth trajectory execution and practical real-time control.  

 \vspace{-5 pt}
\begin{table}[!ht]
\centering
 \caption{Comparison of Performance and Computational Time for DeePC and DeeNE with safety guarantees}
\begin{tabular}{ |p{2.3cm}|p{2.3cm}|p{2.3cm}|  }
\hline
\hline
Controller & RMSE & Time (per loop) \\
\hline
DeePC & $1.48 \hspace{1 mm} cm$ & $200.08 \hspace{1 mm} ms$ \\\hline
DeeNE & $1.49 \hspace{1 mm} cm$ & $30.09 \hspace{1 mm} ms$ \\\hline
\hline
\end{tabular}
\end{table} 

\begin{figure}[!h]
     \centering
     \includegraphics[width=8.8cm, height=5.865cm]{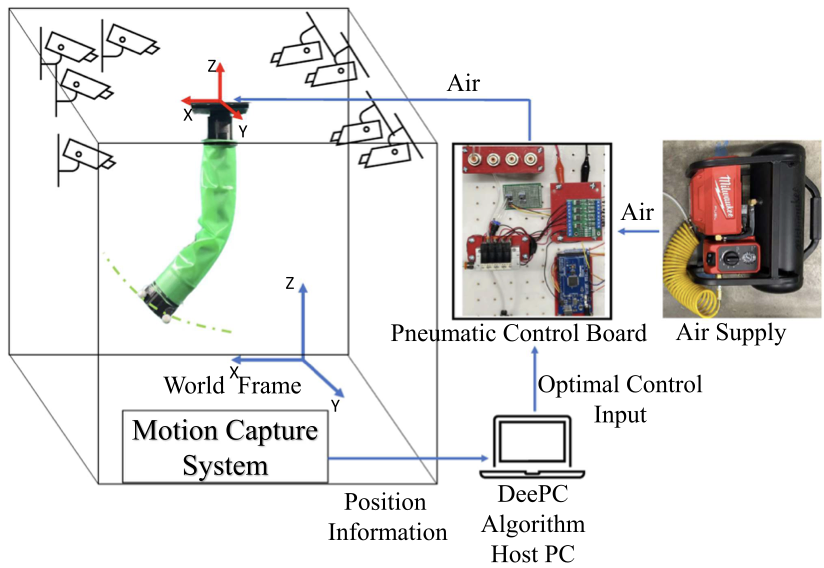}
    \caption{Reference Tracking by Soft Robotic Arm. Experimental results video: \protect\url{https://www.youtube.com/watch?v=yTN6vKkndDg}}
    \label{Soft-Robot}
 \end{figure}

\begin{figure}[!h]
     \centering
     \includegraphics[width=8.8cm, height=6.6cm]{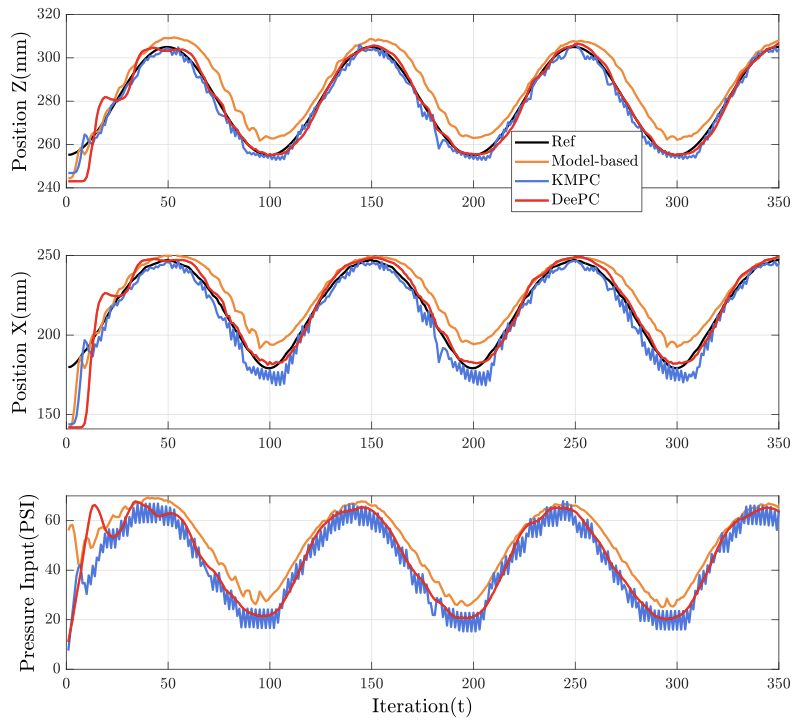}
    \caption{Reference tracking for Soft Robotic Arm.}
     \label{Soft-Output}
\end{figure}

\textbf{Soft Robotic Arm Motion Control:}
Another experimental study was carried out on a pneumatically actuated soft robotic arm to evaluate the performance of DeePC and reduced-order DeePC \cite{wang2024mechanical}, as illustrated in Fig. \ref{Soft-Robot}. The I/O data, defined as the chamber pressure and the end-effector coordinates in task space, were collected under randomized excitation and organized into Hankel matrix. The control objective involved tracking sinusoidal trajectories in both the $z$ and $x$ directions, where an interpolation function $f$ was employed to capture the coupled underactuated dynamics and ensure feasible reference trajectories. The soft robot was equipped with a Qualisys motion capture system, providing accurate real-time measurements of the end-effector position for closed-loop tracking.

\begin{table}[!ht]
\centering
 \caption{Comparison of Performance and Computational Time for DeePC and Reduced-Order DeePC}
\begin{tabular}{ |p{2.3cm}|p{2.3cm}|p{2.3cm}|  }
\hline
\hline
Controller & RMSE & Time (per loop) \\
\hline
DeePC & $4.29 \hspace{1 mm} mm$ & $120.00 \hspace{1 mm} ms$ \\\hline
R-O DeePC & $4.20 \hspace{1 mm} mm$ & $65.00 \hspace{1 mm} ms$ \\\hline
\hline
\end{tabular}
\end{table}

For benchmarking, two methods were considered: (i) model-based controller derived from Euler--Lagrange dynamics with feedback linearization, and (ii) Koopman operator--based MPC (KMPC), where extended dynamic mode decomposition was used to approximate the Koopman operator. Experimental results, Fig. \ref{Soft-Output}, demonstrated that DeePC achieved superior tracking accuracy with smooth pressure inputs, thereby reducing jitter in the pneumatic actuation. While KMPC also performed reasonably well, it generated more oscillatory inputs. However, the model-based controller exhibited larger tracking errors due to modeling inaccuracies. As illustrated in Table III, reduced-order DeePC established a practical balance between tracking accuracy and computational efficiency. 

\begin{figure*}[!t]
     \centering
     \includegraphics[width=0.99\linewidth]{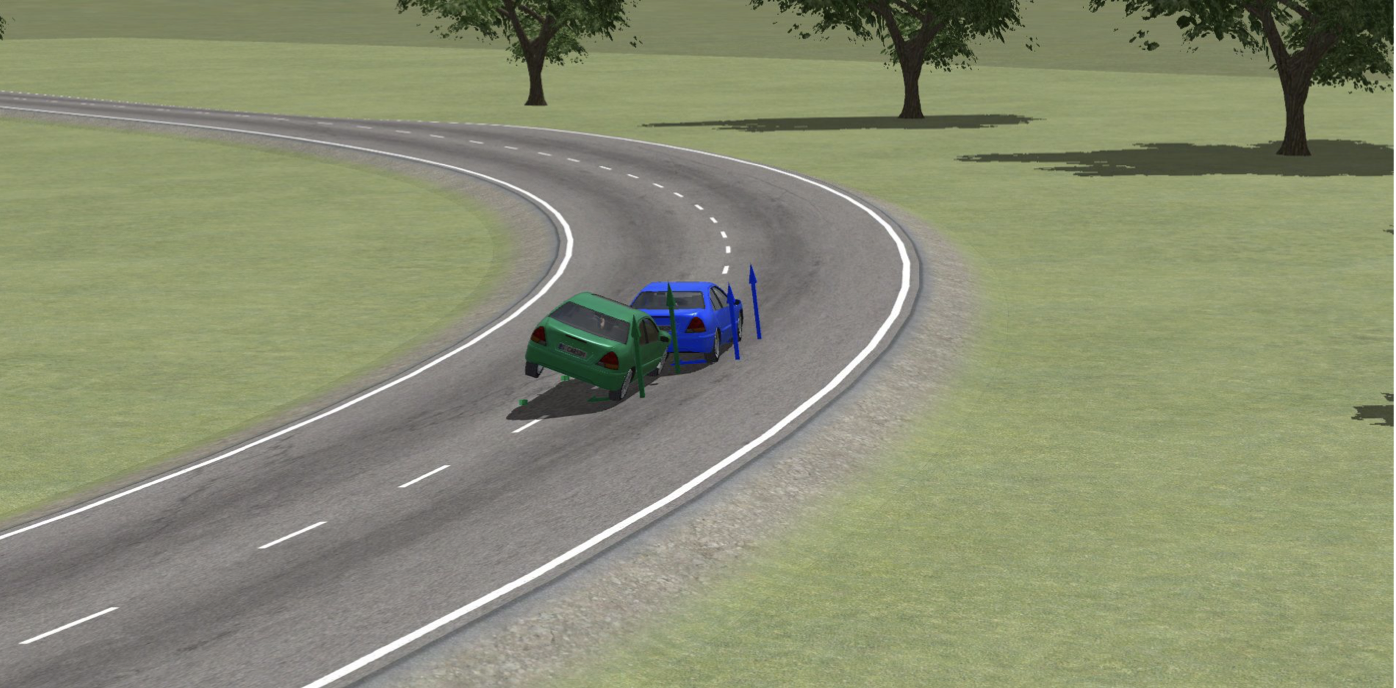}\\
     \caption{A snapshot of the CarSim environment illustrating the implementation of DeePC1 and LMPC1. The image captures the moment at t= 21.7s, where the green vehicle (LMPC1) violates the LTR constraint and begins to lose stability, while the blue vehicle (DeePC1) successfully navigates the sharp turn.}
     \label{CarSim}
\end{figure*}

It is important to note that the difference in RMSE between the 7-DoF robotic arm and the soft robotic arm experiments is primarily attributable to the complexity of the reference trajectory. The ``MSU'' trajectory used in the 7-DoF robotic arm motion control task is significantly more intricate and thus more challenging to track compared to the sinusoidal trajectory employed in the soft robotic arm experiment. This distinction is evident from the wider variation in the $y$- and $z$-directions, as illustrated in Figs. \ref{Outputs3D_Safe} and \ref{Soft-Output}. The use of two different reference trajectories was intentional, aimed at highlighting the efficiency of DeePC under varying levels of trajectory complexity. Another key observation concerns the computational cost: DeePC required 0.2\,s per iteration in the 7-DoF robotic arm case but only 0.12\,s in the soft robotic arm case. This discrepancy arises from the larger Hankel matrix considered for 7-DoF robotic arm because of more complex reference trajectory. Finally, Table III indicates that reduced-order DeePC can even yield improved tracking performance, as the SVD-based dimension reduction scheme effectively mitigates the influence of noise.

\begin{figure}[!h]
     \centering
     \includegraphics[width=8.8cm, height=6.55cm]{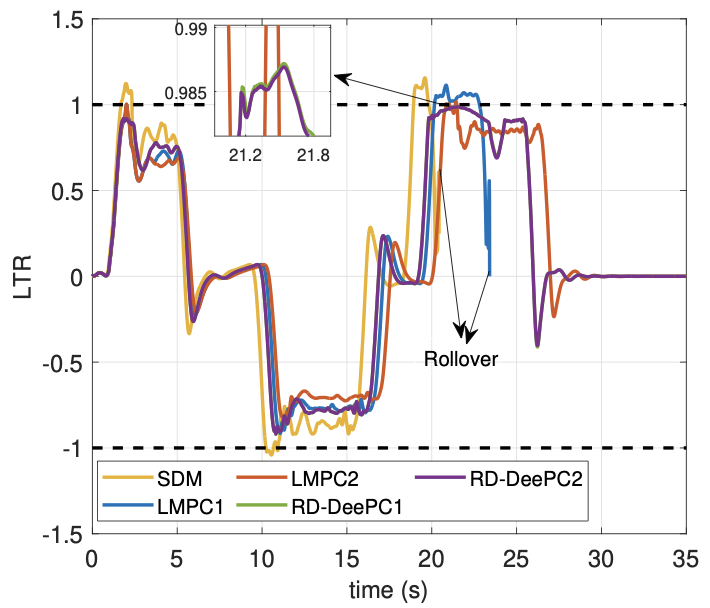}
     \includegraphics[width=8.8cm, height=6.55cm]{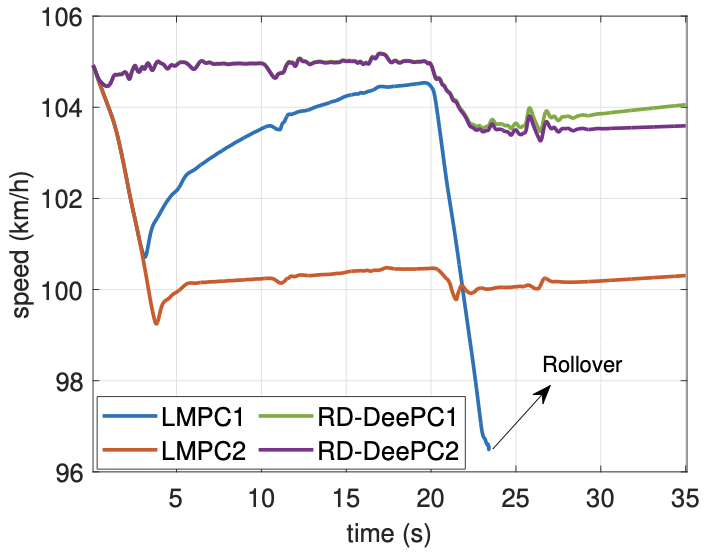}
     \includegraphics[width=8.8cm, height=6.55cm]{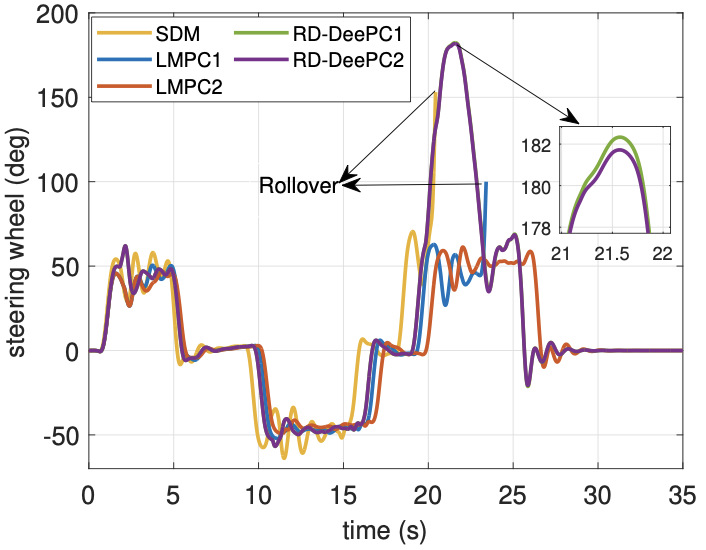}
    \caption{Rollover Avoidance for Sedan Vehicle.}
     \label{CarSim-Output}
\end{figure}

\vspace{2 mm}
\textbf{Vehicle Motion Control:}
Another case study was carried out on a sedan vehicle in the CarSim simulation environment to evaluate the performance of DeePC and reduced-order DeePC \cite{hajidavalloo2025model}, as illustrated in Fig.~\ref{CarSim}. The framework was formulated with steering wheel angle and longitudinal speed as control inputs, and the load transfer ratio (LTR) as the system output. This formulation can be viewed as a data-driven counterpart of active steering control systems. The I/O data were generated using CarSim under high-speed driving conditions with sharp turns, where persistent excitation was ensured by injecting Gaussian noise into the applied inputs. CarSim simulation results demonstrated that reduced-order DeePC achieved comparable tracking performance to the standard DeePC while substantially reducing computational time. When benchmarked against linear MPC (LMPC) and human driver models, reduced-order DeePC consistently outperformed both baselines, particularly in scenarios involving aggressive cornering at $105\,$km/h. As shown in Fig. \ref{CarSim-Output}, while LMPC and the human driver model frequently violated safety constraints on LTR, leading to rollover or near-rollover events, reduced-order DeePC successfully maintained vehicle stability across all turns by dynamically adjusting steering and speed only when necessary. The simple driver model (SDM) is the human-driver baseline embedded in CarSim, employing a PID controller to generate steering commands. LMPC schemes, denoted as LMPC1 and LMPC2, differ only in their cost weighting matrix $R$: LMPC1 emphasizes speed tracking, while LMPC2 reduces this emphasis, improving stability but still permitting safety constraint violations. The reduced-order DeePC schemes, RO-DeePC1 and RO-DeePC2, adopt the same cost weightings as LMPC1 and LMPC2, respectively. Unlike their LMPC counterparts, both RO-DeePC1 and RO-DeePC2 maintain vehicle stability and prevent rollover, while significantly reducing computational complexity through SVD-based Hankel matrix reduction. 

These findings highlight the capability of DeePC---and in particular its efficient variant---to deliver computationally efficient, safety-critical control for robot/vehicle motion control. The results underline the flexibility of DeePC in handling different cost trade-offs without sacrificing safety, a feature not observed in the LMPC counterparts. The integration of SVD-based model reduction not only improves computational efficiency but also enhances robustness by mitigating noise in the collected data, as illustrated in Table IV. 

\vspace{3 mm}
\begin{table}[!ht]
\centering
 \caption{Comparison of Performance and Computational Time for DeePC and Reduced-Order DeePC}
\begin{tabular}{ |p{2.3cm}|p{2.3cm}|p{2.3cm}|  }
\hline
\hline
Controller & Cost & Time (per loop) \\
\hline
DeePC & $9.94 \hspace{1 mm}$ & $144.60 \hspace{1 mm} ms$ \\\hline
R-O DeePC & $9.92 \hspace{1 mm} $ & $89.00 \hspace{1 mm} ms$ \\\hline
\hline
\end{tabular}
\end{table}

\section{Discussion}
The review of efficient data-driven optimal policy approaches highlights that practical deployment of MPC, ML-based MPC, RL, MPC-based RL, and DeePC hinges less on theoretical optimality and more on how computation and memory demands are managed in real time. While MPC has long been considered the benchmark for constrained control, its efficiency depends heavily on the availability of accurate, compact models. ML-based MPC addresses model limitations by incorporating learned surrogates, but this introduces offline training costs and larger memory requirements for storing complex models. RL, by contrast, eliminates online optimization altogether and provides fast execution through policy inference, yet this benefit is offset by the massive data and computation needed for training and by the absence of intrinsic safety guarantees. MPC-based RL mitigates these drawbacks by embedding MPC within the learning loop, balancing safety with adaptability at the expense of higher runtime complexity. DeePC occupies a unique place in this spectrum by dispensing with explicit models and instead relying directly on past trajectory data, but this directness comes with significant computational and memory burdens that grow with data length. The efficiency-enhancing methods reviewed in the final section are therefore crucial for translating DeePC and related approaches into feasible real-time controllers.  

Among these methods, SPC stands out as an elegant compromise between DeePC and MPC. By constructing an autoregressive predictor from data and optimizing directly over future inputs, SPC yields optimization problems of comparable size to MPC while retaining DeePC’s data-driven character. This drastically reduces both computation time and memory storage relative to standard DeePC, though at the cost of introducing modeling bias when nonlinearities are strong. NPC offers a different perspective by reparameterizing the DeePC problem to reduce decision variables, thereby preserving the original feasible set and optimizer. While NPC guarantees identical solutions to DeePC, the computational relief is often modest, and memory requirements remain significant because of the stored factorizations. Reduced-order DeePC provides perhaps the most balanced trade-off by applying singular value decomposition to compress the Hankel matrix. Without truncation, it is equivalent to DeePC but better conditioned and lighter to store; with truncation, it offers further reductions in runtime and memory at the expense of a controlled approximation error. eDDPC achieves similar online dimensions but with improved sample efficiency, making it especially valuable when only limited data are available. Their primary limitation lies in potential numerical fragility during kernel construction, which can compromise robustness in practice. Range-space reformulations also deliver optimization problems of fixed size and modest memory demand, though they may suffer from poor conditioning, which slows convergence unless carefully preconditioned. 

Other approaches focus more explicitly on computational acceleration. DFT-based factorizations eliminate the need to store the Hankel matrix altogether, instead exploiting fast Fourier transforms to implement matrix–vector operations efficiently. This matrix-free formulation preserves DeePC’s statistical behavior while achieving excellent scalability for streaming or online applications where memory is scarce and data windows shift continuously. Learning-based approximations take an even more radical stance by replacing DeePC’s optimization with offline-trained surrogates, such as neural networks that approximate either the cost terms or the entire policy. Once trained, these approximations reduce online execution to fast inference, which is especially valuable in hard real-time settings, though they shift the burden to offline computation and introduce approximation errors that must be carefully managed. Finally, DeeNE leverages local sensitivity information around a nominal DeePC solution to deliver closed-form updates. This drastically reduces online computation to simple matrix–vector multiplications and enables ultra-fast control with only a moderate increase in memory for storing update gains. DeeNE is especially powerful when combined with reduced-order DeePC, which keeps both storage and computation tractable.  

Taken together, these approaches illustrate the diverse strategies available for addressing DeePC’s computational and memory bottlenecks, and by extension the limitations of other data-driven and model-based control schemes. Methods such as SPC and reduced-order DeePC primarily reduce problem size and storage, convex reformulations and DFT-based strategies accelerate the core linear algebra, and amortization techniques like learning-based approximations or DeeNE shift optimization into offline or local computations. When compared to MPC and ML-based MPC, these methods often bring DeePC closer in efficiency while preserving its flexibility in handling unmodeled dynamics. When contrasted with RL and MPC-based RL, they highlight the importance of balancing data efficiency and safety with real-time feasibility. The overall lesson is that no single approach offers a universal solution; rather, hybridization and context-specific tailoring are essential. For instance, SVD-based dimensionality reduction combined with DeeNE updates offers a compelling route to real-time DeePC, while kernel methods may be favored when data are scarce, and learning-based surrogates become attractive when extremely low latency is required. Overall, these methods collectively shift DeePC from a conceptually elegant but computationally demanding tool into a competitive and practical option alongside MPC, ML-based MPC, and RL. Across all paradigms, efficiency-enhancing methods play a decisive role in bridging the gap between theoretical optimality and practical implementation. Finally, efficient online DeePC begins to introduce a new path by combining safety, adaptability, and tractability in a unified framework. Future progress will likely depend on hybrid approaches that integrate the best of model-based structure, data-driven adaptability, and learning-based generalization. In this way, DeePC and its variants are poised to become central enablers of next-generation control systems for robotics, autonomous driving, and energy applications. As these methods mature, they will increasingly enable closed-loop learning and adaptation directly on hardware platforms. This evolution points toward a future in which DeePC-based strategies operate as standard practice in real-time, safety-critical domains.    

\section{Conclusion}
In this work, we explored eight strategies for improving the computational efficiency of DeePC, situating them within the broader landscape of data-driven optimal policy, including MPC, ML-based MPC, RL, MPC-based RL, and LLM agents. Overall, DeePC represents a paradigm shift in predictive control, complementing classical MPC. Although computational demands remain the principal barrier, advances in regularization, dimensionality reduction, and hybridization provide a clear roadmap toward practical deployment. In particular, the combination of SVD-based reductions with DeeNE appears especially promising for developing efficient DeePC implementations. Looking ahead, extending these efficiency-enhancing strategies to conventional MPC paradigms and designing hybrid frameworks that dynamically balance data-driven adaptability, model-based reliability, and computational feasibility represent exciting directions—especially for real-time deployment in robotics, autonomous vehicles, and other complex systems.

\bibliographystyle{ieeetr}
\bibliography{References.bib}
 
\end{document}